%% file: main.tex
\documentclass[10pt,twocolumn,letterpaper]{article}

 \usepackage{cvpr}              

\usepackage{booktabs}

\usepackage{xcolor}
\usepackage{fancyvrb}
\usepackage{fvextra}                     
\usepackage[skins,breakable]{tcolorbox}

\usepackage{colortbl}
\definecolor{gainone}{RGB}{239,248,242}
\definecolor{gaintwo}{RGB}{224,242,230}
\definecolor{gainthree}{RGB}{207,234,216}
\definecolor{gainfour}{RGB}{188,224,200}
\definecolor{lossone}{RGB}{252,239,239}
\definecolor{losstwo}{RGB}{248,224,224}
\definecolor{lossthree}{RGB}{244,207,207}
\definecolor{lossfour}{RGB}{238,190,190}
\definecolor{proftrack}{HTML}{E3EEF9}   
\definecolor{profover}{HTML}{FBEBD3}    
\definecolor{profunder}{HTML}{ECECEC}   

\definecolor{promptbar}{HTML}{3E5573}
\definecolor{promptback}{HTML}{F1F4F8}
\definecolor{prompttext}{HTML}{1C2B3D}
\newtcolorbox{promptbox}[1][]{%
  enhanced, unbreakable,
  colback=promptback, colframe=promptbar, coltext=prompttext,
  boxrule=0.5pt, toprule=4.5pt, arc=1.5mm,
  left=6pt, right=6pt, top=3pt, bottom=3pt,
  before skip=4pt, after skip=4pt,
  #1}

\definecolor{cvprblue}{rgb}{0.21,0.49,0.74}
\usepackage[pagebackref,breaklinks,colorlinks,allcolors=cvprblue]{hyperref}

\def\confName{CVPR}
\def\confYear{2027}

\title{Same Clip, Different Moment: Temporal Twins for Contextual Safety in Multimodal Models}

\author{Mariya I. Vasileva\\
\raisebox{-0.05em}{\includegraphics[width=0.3cm]{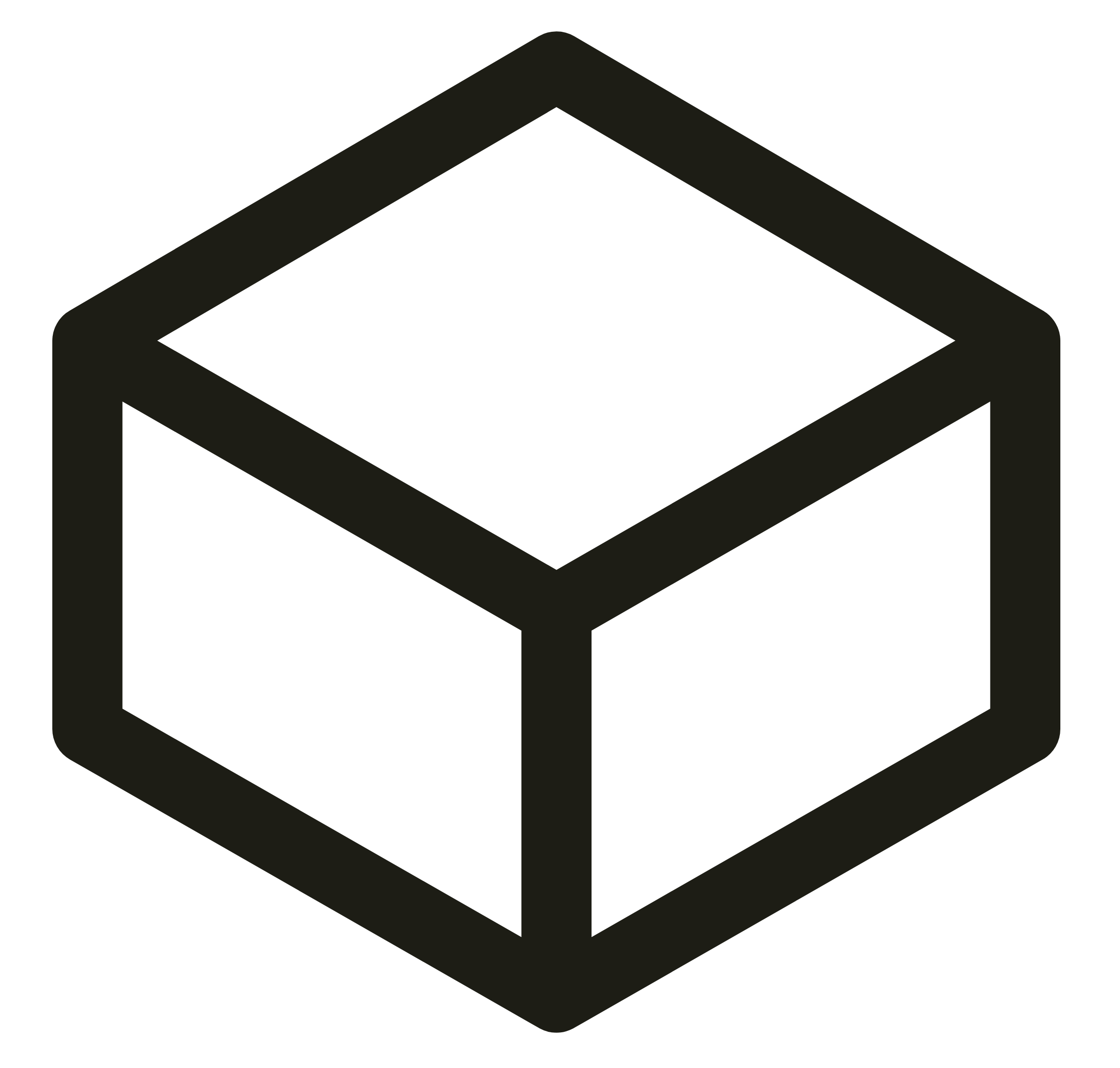}}~Patronus AI\\
{\tt\small mariya.vasileva@patronus.ai}\\
{\small \url{https://mariya.fyi/temporal-twins}}
}
\usepackage{capt-of}
\usepackage{stfloats} 

\begin{document}
\twocolumn[{%
\renewcommand\twocolumn[1][]{#1}%
\maketitle
\begin{center}
\centering
\includegraphics[width=\linewidth]{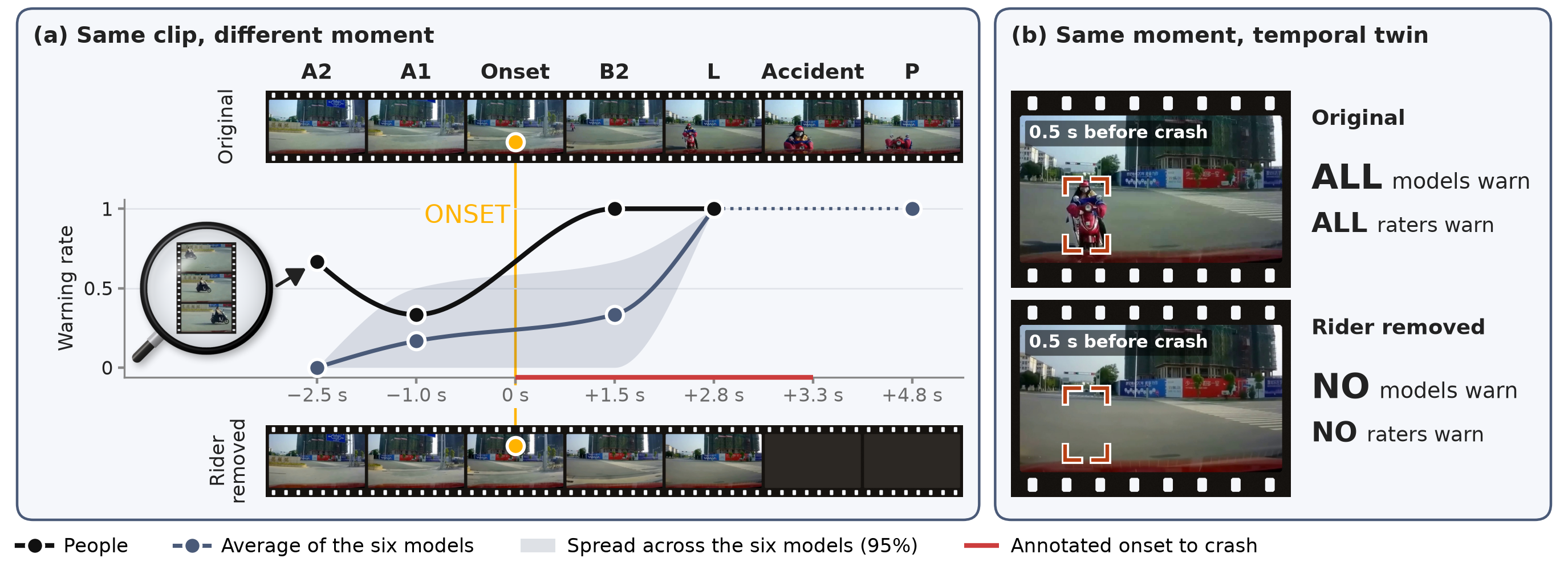}
\captionof{figure}{\textbf{Auditing \emph{when} and \emph{why} video safety monitors warn.}
\textbf{(a)~Temporal trajectory:} share of human raters (black) and six frontier models (slate) issuing warnings across successive 5\,s windows of a real dashcam incident, with its counterfactual twin below.
People begin warning before the annotated onset (as a scooter crosses ahead, inset) and reach unanimous warning soon after onset while most models remain silent, a gap missed by benchmarks that count any warning from onset to impact (shown in red) as timely.
\textbf{(b)~Counterfactual test:} 0.5\,s before impact (red corners mark the motorcyclist's position), erasing the motorcyclist drops warnings from all six models and all raters on the original to zero model warnings on the edit, matching a blinded human observer.}
\label{fig:motivation}
\label{fig:temporal-examples}
\end{center}
\vspace{0.5em}
}]

\input{abstract}
\input{introduction}
\input{related_work}
\input{protocol}
\input{experiments}
\input{conclusion}
\input{llm-usage}

\clearpage
{
    \small
    \bibliographystyle{ieeenat_fullname}
    \bibliography{bibliography}
}

\appendix
\maketitlesupplementary
\input{appendix}

\input{limitations}


\end{document}

%% file: abstract.tex
\begin{abstract}
Frontier multimodal models are increasingly evaluated as real-time video safety monitors, yet standard benchmarks score warnings only against retrospective onset timestamps on unedited clips. Such evaluation cannot distinguish genuine hazard recognition from scene priors or indiscriminate alarming. We address this gap by auditing six frontier models across 925 real dashcam and everyday clips through two lenses: full temporal warning and risk trajectories anchored against human raters viewing identical frames, and 207 counterfactual temporal twins that generatively erase or recolour the implicated road user alongside matched editing controls. Our evaluation reveals that annotated onsets lag human perception, with raters already warning before onset on 29\% of hazard clips, while all six models trail human onset discrimination by 0.06--0.18 AUC and fail in opposite ways: warning chronically during benign lead-up or staying silent as impact approaches. An independent edit audit and blinded human observer confirm that hazard removal cuts pre-impact warnings by up to 77 percentage points where removal succeeds and has no effect where edits fail, yet over-sensitive monitors continue warning without visible cues a person would act on. We release our evaluation code, counterfactual twins, and anonymized annotations for auditing future video safety monitors.
\end{abstract}

%% file: introduction.tex
\section{Introduction}
\label{sec:intro}

Frontier multimodal models are increasingly evaluated as automated safety monitors where a missed warning carries real consequences, including driving, healthcare, daily life, and industry~\cite{zhao2026pasbench}, surveillance~\cite{yao2026multimodal}, and embodied assistants~\cite{jindal2025ai,panpatil2026egosafetybench}. Whereas dashcam collision prediction~\cite{moura2025nexar} and surveillance anomaly detection~\cite{sultani2018real} have long relied on dedicated models, general-purpose multimodal models are now being tested for the same decisions. In these settings, a warning is useful only if it arrives in time to act and for a reason a person would recognize.

Existing work gives substantial reason to question whether current models meet this bar. Multimodal models struggle when identifying risk requires interpreting visual and textual evidence jointly~\cite{palaskar2026vlsu,zhou2025multimodal,wang2025safe}, overreact to benign scenes resembling emergencies~\cite{choi2025better,li2025mossbench}, and recognize hazards without acting to mitigate them~\cite{torresfonseca2026safetyalfred}. Video introduces further difficulties: models miss harmful meaning distributed across individually benign segments~\cite{wang2026distributed}, sparse frame sampling misses brief harmful events~\cite{cao2026failures}, and proactive video-monitoring benchmarks report a persistent trade-off between early detection and false alarms~\cite{zhao2026pasbench,qiu2026sports}.

Auditing whether a monitor responds to developing danger requires tracking its decisions over time rather than scoring a single video-level label or retrospective onset timestamp. Across 37 video safety and anomaly datasets we surveyed (Appendix~\ref{app:datasets}), existing benchmarks either label whole clips without temporal localization~\cite{chen2025safewatch}, rely on synthetic footage whose lead-up reflects a generator rather than real physics~\cite{liu2025video,wang2026distributed,jindal2025ai,pu2026homesafe,panpatil2026egosafetybench}, use coarse or predictable event timing~\cite{du2024uncovering,bao2020uncertainty}, or prohibit controlled edits~\cite{moura2025nexar}.

We therefore focus on real footage of road and everyday accidents where danger develops on camera, requiring five properties: (i)~real footage; (ii)~temporally precise annotations of the hazard onset and the accident; (iii)~sufficient pre-onset footage to observe the lead-up; (iv)~comparable hazard-free footage; and (v)~spatial localization of the hazard with permission to construct controlled edits. No single surveyed dataset meets all five requirements. We combine dashcam footage from MM-AU~\cite{fang2024abductive,fang2019dada}, which provides hazard localization, with real hazard and hazard-free clips from PaSBench-Video~\cite{zhao2026pasbench}, allowing us to study a specific question:

\begin{center}
\begin{minipage}{0.92\linewidth}
\centering\bfseries
Do frontier video safety monitors warn because of an emerging hazard, and as early and reliably as a person watching the same visual evidence~would?
\end{minipage}
\end{center}

We evaluate six frontier models across 8,011 five-second (31-frame) windows from 925 real clips against a three-rater human reference viewing identical frames, and against 207 five-version counterfactual temporal twins from which we generatively erase or recolour the implicated road user. All six models trail human onset discrimination by 0.06--0.18 AUC, with warning rates and mean risk ratings splitting into chronic pre-onset over-warning versus late under-warning. Removing the implicated road user lowers warning rates by 12.1--47.3 percentage points shortly after onset and 16.4--62.3 points just before impact. A two-coder edit audit and blinded human observer on 100 twins confirm that the removal effect reaches 17.3--76.5 percentage points where removal succeeds (and vanishes on failed edits), while separating models whose residual warnings sit largely on failed edits from models that continue to warn on clean removals without human-corroborated cues.

We make the following contributions:
\begin{itemize}
    \item \textbf{Temporal measurement of contextual safety.} We evaluate frontier video safety monitors through their full warning and risk-response curves as visual evidence of danger develops, capturing timing and calibration failures invisible to video-level detection accuracy alone.

    \item \textbf{A human reference for real-time warning decisions.} We collect independent human judgments on the exact 31-frame observations presented to the models, separating prospective warning judgments from retrospectively assigned onset labels.

    \item \textbf{An audited counterfactual test of hazard-dependent warnings.} We construct 207 five-version temporal twins of real accident footage, together with a two-coder edit audit and a blinded human-observer study on 100 twins, separating genuine causal responsiveness to the localized hazard from imperfect edits, remaining evasive motion, and chronic over-warning.
\end{itemize}

%% file: related_work.tex
\section{Related Work}
\label{sec:related}

\paragraph{Safety in multimodal models.}
Most multimodal safety benchmarks ask whether a model's \emph{response} is safe when risk depends on the image--text combination, testing queries whose safety changes with the image~\cite{zhou2025multimodal}, prompts harmful only in combination~\cite{rottger2025msts}, and refusal on benign images~\cite{li2025mossbench}. Models fail when joint understanding is required~\cite{palaskar2026vlsu,lee2026holisafe}, while benchmark design can introduce shortcuts: text queries can leak the image's risk~\cite{hu2025vlsbench}, and label audits show that some benchmarks can be solved from cues other than the hazard~\cite{barrios2026safety}. Video work extends these tests to generated harmful videos, temporal jailbreaks, and moderation filters~\cite{liu2025video,zhou2026tempjail,cao2026failures,cao2026temporal}. We ask a different question: whether a model that \emph{monitors} real footage notices danger as it becomes visible. Unlike static or clip-level safety checks, this requires tracking decisions across successive moments as visual evidence unfolds, and anchoring them to when people viewing the same frames judge the risk as actionable, separating timely warnings from premature alarms and late misses.

\paragraph{Video safety monitoring and accident anticipation.}
Recent work has begun evaluating multimodal models directly as video safety monitors: PaSBench-Video~\cite{zhao2026pasbench} scores streaming safety warnings on clips with annotated risk onsets, RiskCueBench~\cite{luo2026riskcuebench} tests anticipatory reasoning from early risk cues and reports a large gap to people, and related benchmarks and safeguards probe egocentric, household, and same-scene contextual risk~\cite{panpatil2026egosafetybench,pu2026homesafe,lu2026homeguard,zhang2026guardianbench} or zero-shot surveillance anomaly detection~\cite{yao2026multimodal}. In parallel, computer vision methods for dashcam accident anticipation and detection train dedicated models and report AUC, average precision, and mean time-to-accident~\cite{chan2016anticipating,yao2019unsupervised,bao2020uncertainty,bao2021drive}, using onset-annotated datasets such as DADA-2000~\cite{fang2019dada} and MM-AU~\cite{fang2024abductive}. Recent work also compares human and model attention during anticipation~\cite{ryu2026mind}. We build on this footage and its onset labels, but track general-purpose monitors across the full lead-up rather than at a single timestamp, anchoring their warning decisions against people viewing the same frames.

\paragraph{Counterfactual and minimal pair evaluation.}
Minimal pair benchmarks test what a model relies on in image--text~\cite{hsieh2023sugarcrepe,le2023coco,li2024naturalbench} and video~\cite{zhang2024vinoground,krojer2025shortcut}, while counterfactual image edits isolate safety-critical features~\cite{helbling2026safetypairs}. In driving, removing an agent serves the same diagnostic purpose for identifying risk objects~\cite{li2020who} and stress-testing motion forecasting~\cite{roelofs2022causalagents}. Recent diffusion and instruction-driven video editors make it possible to perform such removals directly on real clips by erasing objects and their side effects~\cite{zhou2023propainter,zi2025minimax,miao2025rose,bfl2026fluxvideoedit}. However, because edits leave detectable traces and imperfect fills~\cite{montibeller2026ainpaint,huang2026vor}, comparing an edited or generated clip against an unedited one (as in CCTVBench~\cite{zhou2026cctvbench}, which pairs a real traffic clip with a generated counterfactual continuation) confounds the target factor with generation artifacts. To avoid this confound, we follow image work that uses reconstruction or no-op edits as controls~\cite{prabhu2023lance,menzat2026counterfactual}, editing real footage in both arms with the same editor~\cite{bfl2026fluxvideoedit}.

%% file: protocol.tex
\section{Temporal Monitoring Protocol and Counterfactual Twins}
\label{sec:protocol}

\subsection{Data and Observation Protocol}
\label{sec:proto-obs}

\paragraph{Observation windows and decision points.}
Each decision concerns the preceding 5.0\,s of footage, represented by 31 frames spaced 1/6\,s apart. Models receive the frames in chronological order, with no information about subsequent events or responses to earlier queries. Human raters view the same frames and answer the same two questions: whether to issue a warning now, and how high the risk is on a five-level scale from none to critical. We use the neutral safety-monitoring prompt of PaSBench-Video~\cite{zhao2026pasbench}, adapted to independent observation windows (Appendix~\ref{app:prompt}). The detailed frame-sampling rationale appears in Appendix~\ref{app:frame-sampling}.

Let $r$ denote the annotated hazard onset and $a$ the annotated accident time. We use fixed decision points to compare responses at different stages of an event: A2 ends 2.5\,s before $r$, A1 ends 1.0\,s before $r$, B ends 0.5\,s after $r$, and B2 ends 1.5\,s after $r$, when sufficient footage remains before the accident. L ends 0.5\,s before $a$, and P ends 1.5\,s after $a$, where footage permits. All times refer to the final frame of the 31-frame observation window. We distinguish the annotated onset from the first visible sign of danger and from the moment at which a warning becomes warranted.

\paragraph{Evaluation set and evaluated models.}
We evaluate six frontier models on 8,011 observation windows per model from 925 clips: 490 MM-AU DADA accident clips, 176 real PaSBench hazard clips, and 259 hazard-free clips. The models are Claude Opus~5.5, GPT-6 Astra, Gemini~4 Argon, Kimi-K3.1-Pro, GLM-5.3-Flash, and Nemotron-3.5-Super-VL. Every model receives the same 31-frame observation at each decision point, using independent queries with no accumulated history. We also repeat a fixed sample of 801 windows to measure response stability; repeat calls do not enter the primary response curves. For each hazard clip, we sample decision times every 0.5\,s, beginning before the annotated onset and continuing through the development of the event, subject to the available footage, together with the accident-aligned windows L and P.

\paragraph{Metrics and temporal response measures.}
Our primary measure of temporal discrimination is within-clip onset AUC: the risk rating at B receives a score of 1 if it exceeds the rating at A1, 0.5 if the ratings are equal, and 0 otherwise. We average these scores across clips, so 0.5 indicates no directional separation between the two moments. We also compare A1 with A2 to measure changes during the lead-up, and B2 and L with A1 to examine how separation develops as the accident approaches. To compare each model against people, we compute paired clip-level differences in onset AUC against the mean of individual-rater scores (primary reference, with a planned equivalence margin of $\pm0.10$ AUC) and against the risk ratings averaged across raters (secondary reference). All confidence intervals are obtained by resampling clips.

Because two models may have similar onset AUCs while differing substantially in when and how often they warn, we also report warning rates and mean risk levels at each named decision point and on hazard-free footage. We additionally summarize each model's peak warning rate ($\ge 100$ eligible clips), its largest increase between consecutive 0.5\,s decision times on matched clips ($\ge 100$ matched clips), its median first-warning lead time before the accident, and the proportion of clips without a pre-accident warning (Appendix~\ref{app:model-timing}).

\subsection{Human Reference Study}
\label{sec:proto-human}

We collect independent human assessments on 820 windows from 180 real clips, including 150 hazard clips and 30 hazard-free clips. The hazard clips comprise 120 MM-AU DADA examples~\cite{fang2024abductive,fang2019dada} and 30 PaSBench-Video examples~\cite{zhao2026pasbench}. Three human raters independently assess the same 31-frame inputs presented to the models. They can pause, replay, and inspect individual frames, but cannot revise an earlier assessment after viewing a later window from the same video. We ask for both a binary warning decision and a risk rating. The human reference therefore records the decisions people make from the available visual evidence, without disclosing how the event ends.

\subsection{Counterfactual Temporal Twins}
\label{sec:proto-twins}

\begin{figure*}[!t]
\centering
\includegraphics[width=\textwidth]{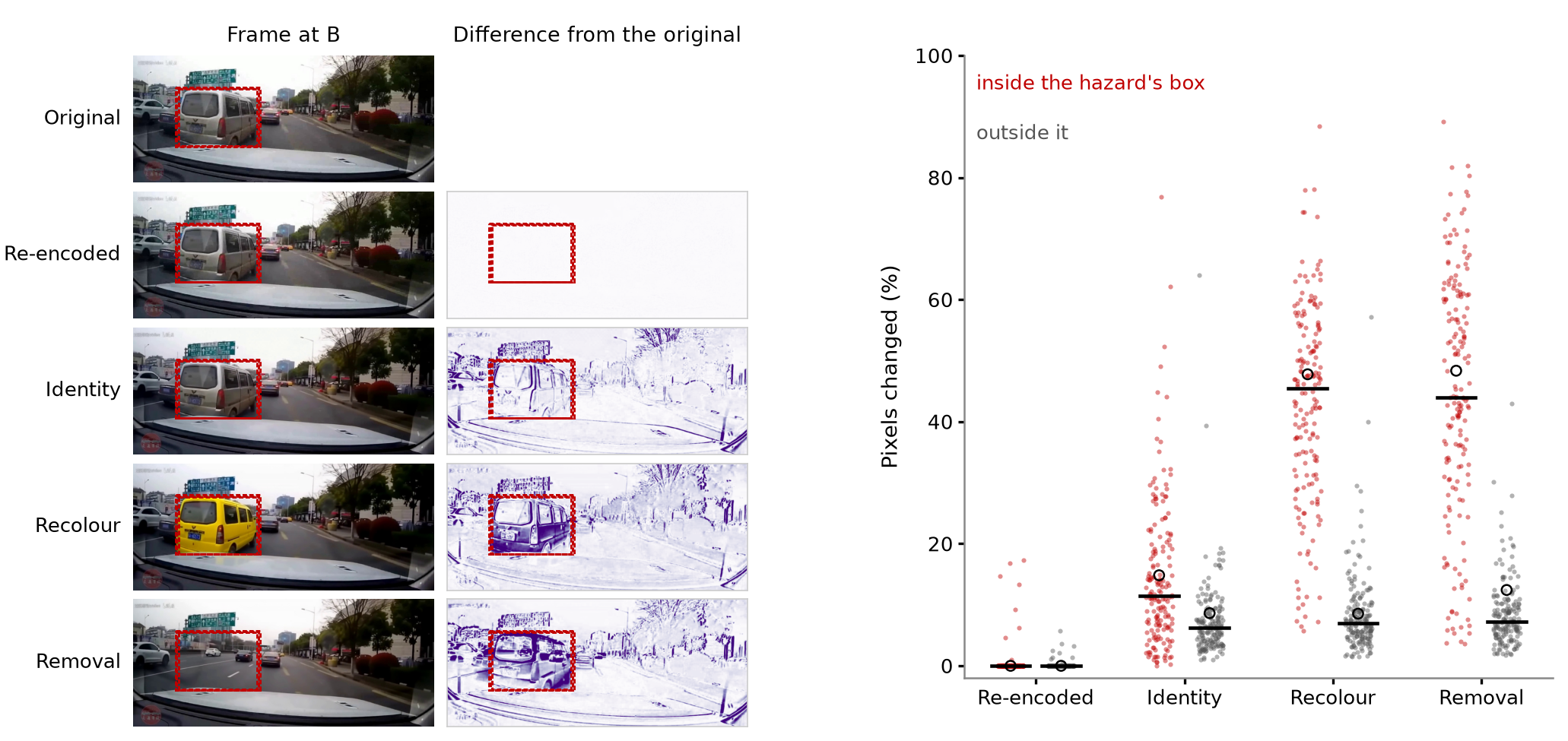}
\caption{\textbf{Where the edits change the image.} Left: the last frame of the window at B for one twin (MM-AU DADA, accident category 11, video 31) in each version, next to its absolute difference from the original frame; the dashed box is the target road user's annotated box. Right: the share of pixels that change inside and outside that box, for the 176 of the 207 twins with a box at B; bars mark medians and circles the example on the left. A pixel counts as changed if its grey level differs by more than 25 of 255 after a light blur. The median change inside the box is 11\% for the identity edit, 45\% for recolour and 44\% for removal; outside the box it is 6\%, 7\% and 7\%.}
\label{fig:app-twin-construction}
\end{figure*}

A warning at the annotated onset can reflect either an implicated road user or broader visual context (e.g., traffic density, intersection geometry, or camera motion). To test whether warnings depend on the hazard itself, we construct edited versions of real accident footage that remove that road user and inpaint the occluded background. We call these matched variants \emph{counterfactual temporal twins}, and evaluate each of them at identical timestamps using the same 31-frame protocol. Because each comparison holds the source clip and timing fixed, it does not rely on the annotated onset being the true warning time.

\paragraph{Counterfactual and control arms.}
We construct five versions of each clip: 
(a)~the original footage;
(b)~a re-encoded control (E0) matching the editor's format, resolution, and frame rate;
(c)~an identity edit (prompted to change nothing) to measure pass-through effects;
(d)~a recolour edit that changes the target road user's colour while keeping its position and motion; and
(e)~a removal edit that deletes that road user and fills in the background.
Our primary comparison is recolour versus removal: both apply generative editing to the target region, isolating sensitivity to the road user's presence from other editing effects.

\paragraph{Identifying and editing the hazard.}
We draw from MM-AU DADA and PaSBench's DADA subset, for which MM-AU provides hand-corrected bounding boxes around road users~\cite{fang2024abductive,fang2019dada}, excluding single-vehicle accidents. An automated rule selects the candidate road user from the accident description and bounding boxes, and an independent vision-language judge outside our evaluated panel (Meta Muse-Spark-Open) identifies the causal road user from six numbered frames around the accident without seeing our rule's choice. Retaining clips where the two selections agree yields 207 twins: 193 from MM-AU and 14 from PaSBench (Appendix~\ref{app:twins}).

We use FLUX Video Edit~\cite{bfl2026fluxvideoedit} to produce the identity, recolour, and removal versions over the segment needed up to L ($a - 0.5$\,s), stopping before the accident itself. Across the 207 twins, 116 provide A2, A1, B, and L; 68 provide A1, B, and L; and 23 provide B and L. Because FLUX returns 24\,fps video ($1440\times608$) versus 30\,fps original footage and generative editing can introduce slight temporal drift, we align each nominal sample time to the edited frame whose background outside the target's box best matches the original. Across all 41,337 sampled frames, 97.7\% match within one original source frame (931 frames, or 2.3\%, differ by more than one frame), and pixel changes concentrate inside the target's box (median 44\% for removal and 45\% for recolour versus 7\% outside, closely matching the identity edit's 6\% outside; Figure~\ref{fig:app-twin-construction}). We repeat primary comparisons on the 115 twins within two source frames and 52 within one (Appendix~\ref{app:twins}).

%% file: experiments.tex
\section{Experiments and Results}
\label{sec:results}
\label{sec:res-temporal}

\subsection{When do people judge that a warning is warranted?}
\label{sec:res-human}

\paragraph{Temporal discrimination and pre-onset warnings.}
People distinguish the visual evidence on either side of the annotated onset. The mean individual-rater onset AUC is 0.73 (95\% CI 0.70 to 0.77; individual raters achieve 0.76, 0.68, and 0.77), while the AUC obtained from the raters' mean risk is 0.82 (95\% CI 0.77 to 0.87). The distinction becomes clearer closer to the accident. On the 75 clips with a B2 window, individual-rater AUC rises from 0.72 at B to 0.84 at B2 (paired difference 0.12, 95\% CI 0.08 to 0.16), and on all 150 clips it reaches 0.89 at L (95\% CI 0.86 to 0.92). The warning rate increases from 25\% at A1 to 56\% at B and 84\% at L, and mean risk rises from 0.65 at A1 to 1.50 at B and 2.64 at L (Figure~\ref{fig:curves}). On hazard-free footage, the corresponding onset comparison on the raters' mean risk has an AUC of 0.50 (95\% CI 0.40 to 0.58), and people warn on 13\% of windows.

The annotated onset does not always coincide with the first moment people consider dangerous. Across the MM-AU DADA subset, a majority of raters first warn before the annotated onset on 29\% of clips (35 of 120), while on 9\% no evaluated moment receives a majority warning. By comparison, on the 30 rated hazard-free clips, a majority warns at one of the two earliest windows on 10\% of clips, putting MM-AU DADA's pre-onset majority rate 19 percentage points above the hazard-free baseline (95\% CI 5 to 32; Appendix~\ref{app:human-reference}). Human risk also changes during the lead-up: the A1-versus-A2 AUC is 0.56 for individual raters (95\% CI 0.53 to 0.59), and 0.63 for the raters' mean risk (95\% CI 0.57 to 0.68). The onset annotation is therefore informative about changes in visible risk, but cannot be treated as the unique time at which a warning becomes justified.

\paragraph{Agreement between human raters.}
Human judgments show moderate agreement on binary warning decisions (81.9\% pairwise exact agreement; Fleiss' $\kappa=0.63$) and reasonably strong ordinal agreement on risk severity (Krippendorff's $\alpha=0.75$; Table~\ref{tab:human-reference}). Together, these measurements support a human reference that captures both shared judgments and variation in warning thresholds. Further details of the annotation protocol, consensus thresholds, and interface appear in Appendix~\ref{app:human-reference}.

\begin{table}[!t]
\centering
\small
\setlength{\tabcolsep}{3pt}
\begin{tabular}{@{}p{0.47\columnwidth}l@{}}
\toprule
Measure & Estimate (95\% CI) \\
\midrule
Warning agreement (\%) & 81.9 (79.2--84.4) \\
Warning Fleiss' $\kappa$ & 0.63 (0.58--0.68) \\
Risk agreement (\%) & 58.7 (55.3--62.2) \\
Ordinal Krippendorff's $\alpha$ & 0.75 (0.71--0.79) \\
Human onset AUC & 0.73 (0.70--0.77) \\
\bottomrule
\end{tabular}
\caption{\textbf{Agreement and temporal discrimination in human judgments.} Warning and risk agreement refer to matched observation windows; onset AUC compares B with A1 within clips, averaged across individual raters. Intervals are 95\% clip-bootstrap confidence intervals.}
\label{tab:human-reference}
\end{table}

\subsection{Do frontier monitors recognize emerging hazards as reliably as people?}
\label{sec:res-models}

\begin{figure*}[!t]
\centering
\includegraphics[width=\linewidth]{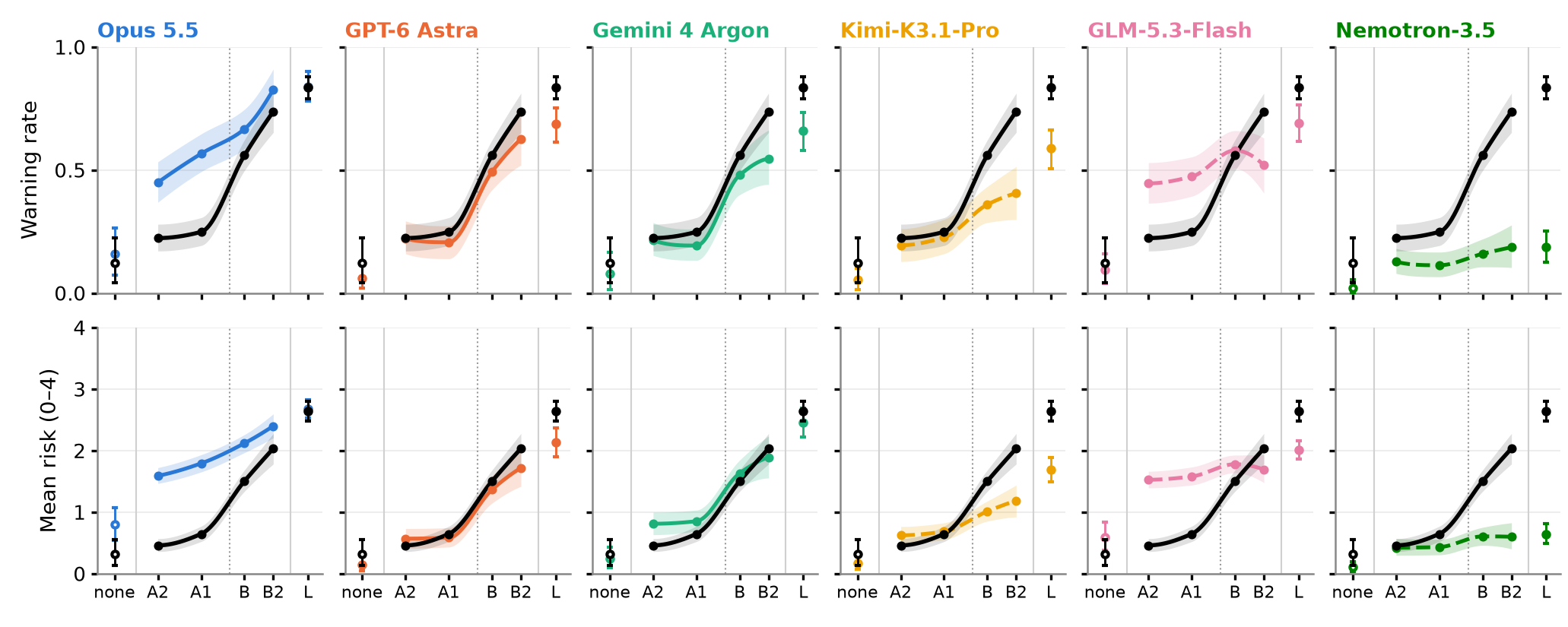}
\caption{\textbf{Same evidence, different alarms: warning rates and mean risk levels show the same response profiles.} Warning rate (top row) and mean risk level (bottom row; 0 is none and 4 is critical) of each model (colour) and of people (black, repeated in every panel) on the 150 hazard clips that people rated (120 MM-AU DADA, 30 PaSBench). Dots mark decision times A2, A1, B, and B2 (A2 on 149 clips, B2 on 75); curves interpolate between them, and the dotted line marks the annotated onset. L, 0.5\,s before the accident, is shown separately because its onset-relative time differs between clips; ``none'' gives the value on 30 hazard-free clips. Columns group models by response profile. Each title gives the model's within-clip onset AUC next to the human reference (0.73, averaged across individual raters). Bands and bars show 95\% confidence intervals from resampling clips.}
\label{fig:curves}
\label{fig:risk-curves}
\end{figure*}

\begin{figure*}[!tb]
\centering
\begin{minipage}[t]{0.49\textwidth}
\centering
\includegraphics[width=\linewidth]{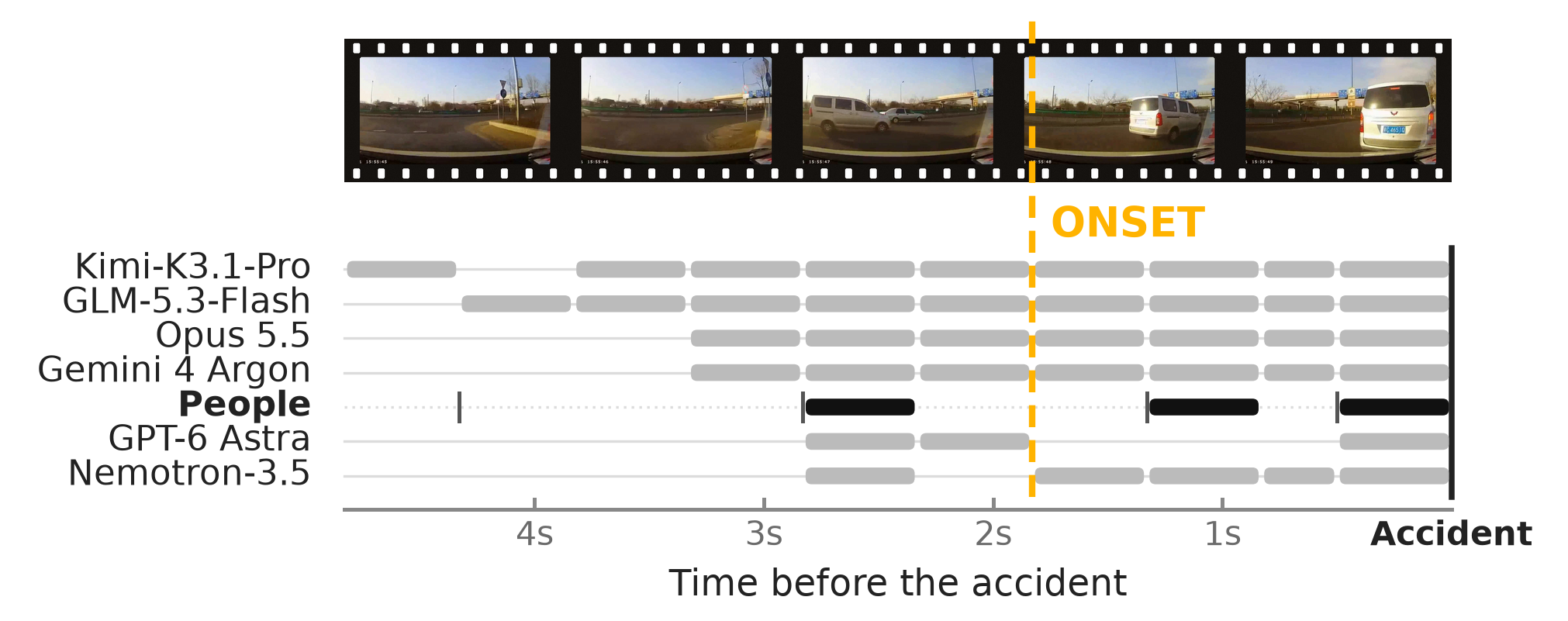}
\end{minipage}\hfill
\begin{minipage}[t]{0.49\textwidth}
\centering
\includegraphics[width=\linewidth]{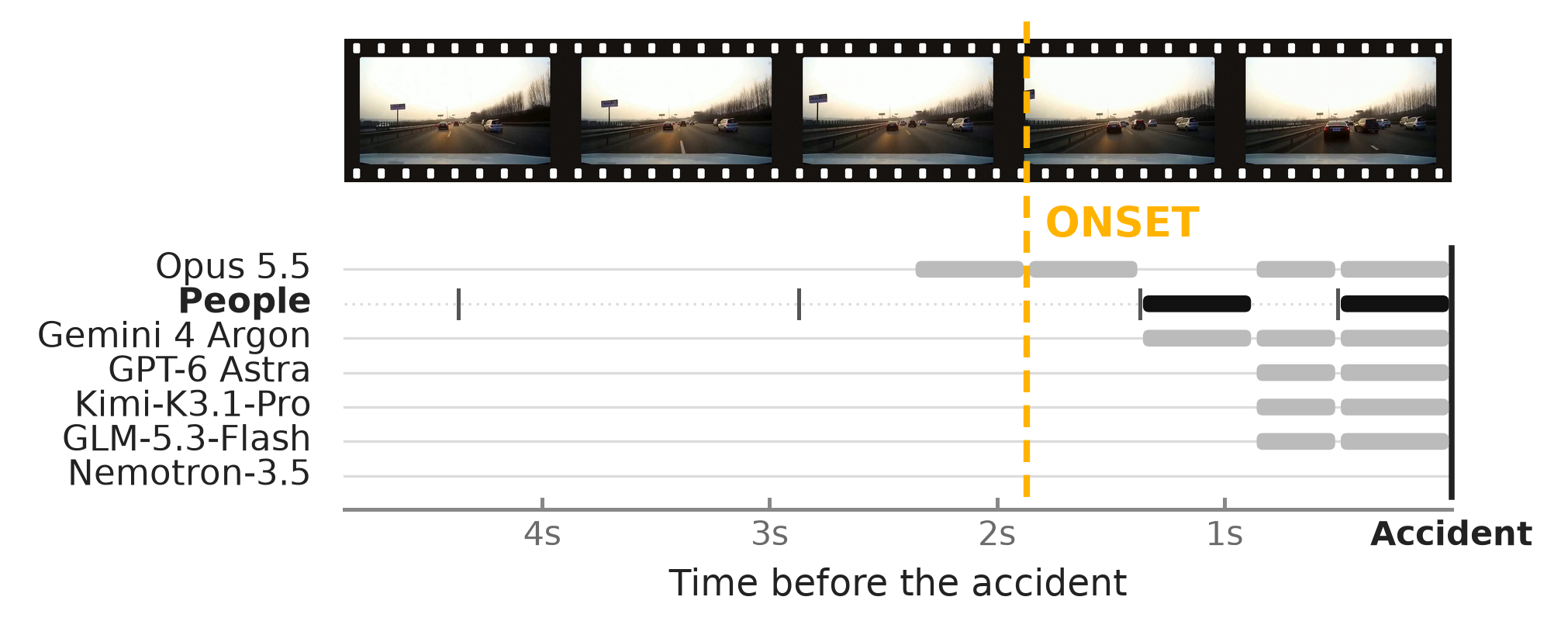}
\end{minipage}

\vspace{4pt}
\begin{minipage}[t]{0.49\textwidth}
\centering
\includegraphics[width=\linewidth]{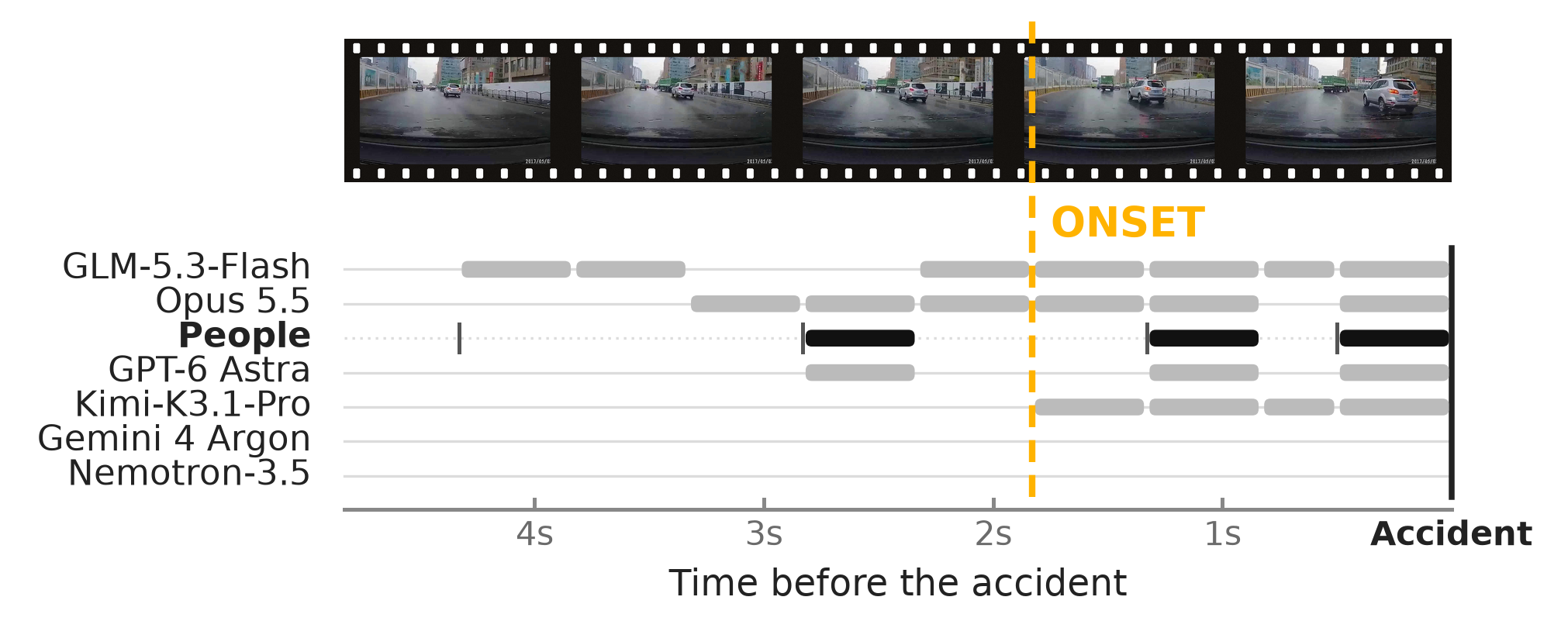}
\end{minipage}\hfill
\begin{minipage}[t]{0.49\textwidth}
\centering
\includegraphics[width=\linewidth]{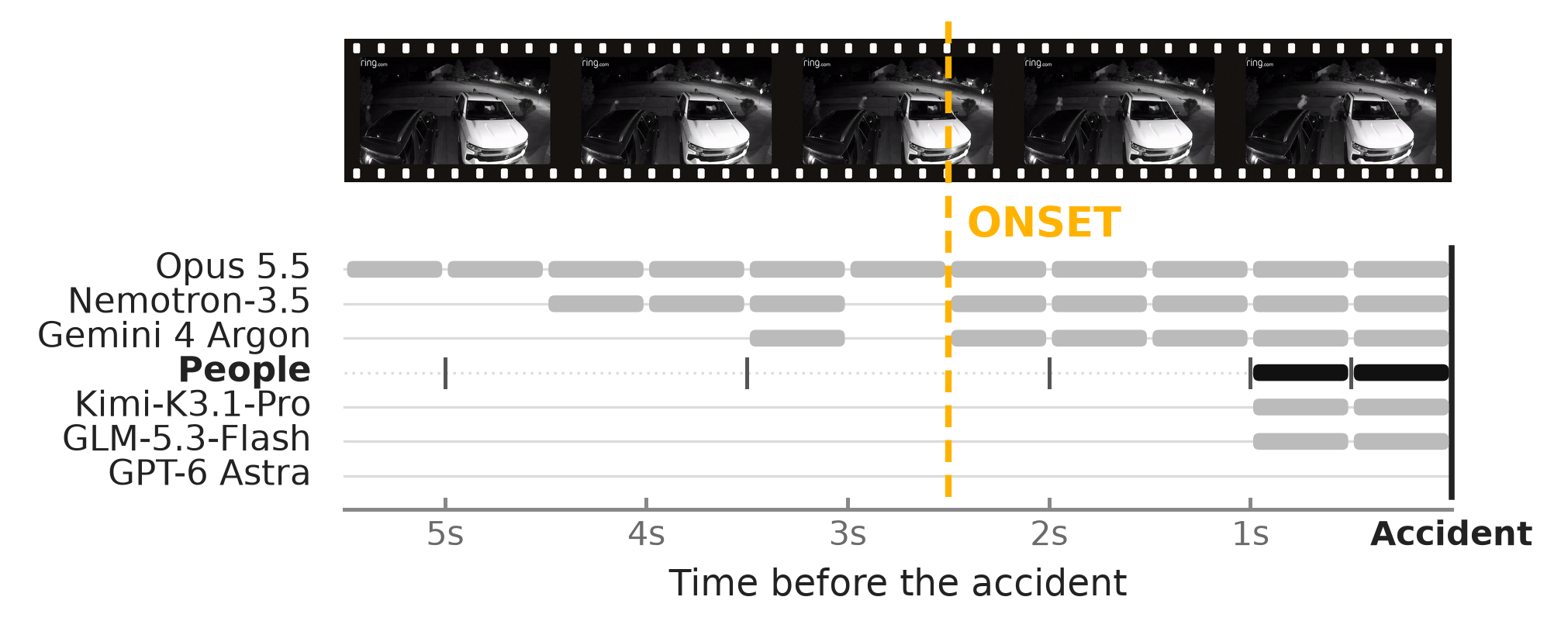}
\end{minipage}
\caption{\textbf{When people and models warn, clip by clip.} Four hazard clips illustrating contrasting response profiles: five frames evenly spaced over the window shown and, below them, one row per responder on an axis of time before the accident. Grey blocks mark model warnings at 0.5\,s decision points (rows ordered by first warning); black blocks on the dotted row mark human majority warnings at the four or five rated moments; the dashed yellow line marks the annotated hazard onset and the solid black line marks the accident. People warn before the annotated onset (top left), immediately after it (top right), throughout the clip (bottom left), and only later (bottom right, PaSBench). Additional clip timelines appear in Appendix~\ref{app:model-timing}.}
\label{fig:clip-timelines-main}
\end{figure*}

\paragraph{Onset discrimination and response profiles.}
The models differ substantially in how their warning responses develop around the onset (Figure~\ref{fig:curves}). All six have lower within-clip onset AUC than the mean individual-rater reference. The difference in favor of people is 0.14 for Claude Opus~5.5 (95\% CI 0.08 to 0.20), 0.14 for Kimi-K3.1-Pro (95\% CI 0.09 to 0.20), and 0.18 for GLM-5.3-Flash and Nemotron-3.5-Super-VL (95\% CIs 0.12 to 0.24 and 0.13 to 0.23, respectively). All four of these models fall reliably below every individual rater (3 of 3) and every leave-one-rater-out reference (Appendix~\ref{app:model-results}). The difference is smallest for GPT-6 Astra, at 0.06 (95\% CI 0.01 to 0.11), and Gemini~4 Argon, at 0.06 (95\% CI 0.01 to 0.12), which match the lowest individual rater (Rater~2 at 0.68) while trailing Raters~1 and~3 (0.76 and 0.77). Astra's difference also lies within the planned equivalence margin of $\pm0.10$ (90\% CI 0.018 to 0.099), whereas Argon's does not (90\% CI 0.017 to 0.113). Against the raters' mean risk, all six models have lower onset AUC, with differences ranging from 0.15 to 0.27.

A lower onset AUC does not imply that a model always responds later, nor is it simply a binary warning-threshold offset. Opus~5.5 and GLM-5.3-Flash warn on 57\% and 47\% of clips at A1 (compared with 25\% for people), and also rate mean risk roughly one level above people there (1.79 and 1.58 versus 0.65; Figure~\ref{fig:curves}, bottom row), leaving less separation across the onset. Kimi-K3.1-Pro and Nemotron-3.5-Super-VL show the opposite pattern: at L, they warn on 59\% and 19\% of clips (against 84\% for people), with mean risk 0.94 and 1.99 levels below people (1.70 and 0.65 versus 2.64). Astra and Argon warn less often than people at L (69\% and 66\%), and Astra's mean risk is also lower there (2.13 versus 2.64), whereas Argon is the only model whose lower warning rate at B2 (55\% versus 74\%) and L is not accompanied by a reliably lower mean risk (Appendix~\ref{app:model-results}).

\paragraph{Full evaluation set and timing.}
Across the full evaluation set (Table~\ref{tab:all-clips}), onset AUC ranges from 0.54 for Nemotron-3.5-Super-VL to 0.72 for Astra and Argon. Five models issue a warning on 69--81\% of available post-event windows P, while Nemotron does so on 40\%. Warning rates on hazard-free windows range from 4\% to 16\%, and repeat agreement ranges from Cohen's $\kappa=0.51$ to 0.93. Over the 666 hazard clips, Opus~5.5 and GLM-5.3-Flash give their first warnings earliest (median 3.4 and 3.5\,s before the accident), whereas Nemotron-3.5-Super-VL issues no pre-accident warning on 52\% of hazard clips (Figure~\ref{fig:clip-timelines-main}; full timing summaries and additional timelines appear in Appendix~\ref{app:model-timing}). These measurements do not establish which visual cues drive model warnings; we therefore test whether their warnings depend on the implicated road user in the same footage.

\begin{table}[!t]
\centering
\small
\setlength{\tabcolsep}{3pt}
\resizebox{\columnwidth}{!}{%
\begin{tabular}{@{}lcccc@{}}
\toprule
Monitor &
\begin{tabular}[b]{@{}c@{}}Onset\\AUC $\uparrow$\end{tabular} &
\begin{tabular}[b]{@{}c@{}}Warns\\at P\end{tabular} &
\begin{tabular}[b]{@{}c@{}}Hazard-free\\warnings $\downarrow$\end{tabular} &
\begin{tabular}[b]{@{}c@{}}Repeat\\$\kappa$ $\uparrow$\end{tabular} \\
\midrule
Claude Opus 5.5 & 0.67 & 0.81 & 0.16 & 0.82 \\
GPT-6 Astra & \textbf{0.72} & 0.69 & 0.09 & \textbf{0.93} \\
Gemini 4 Argon & \textbf{0.72} & 0.77 & 0.07 & 0.87 \\
Kimi-K3.1-Pro & 0.62 & 0.69 & 0.09 & 0.68 \\
GLM-5.3-Flash & 0.56 & 0.81 & 0.13 & 0.54 \\
Nemotron-3.5-Super-VL & 0.54 & 0.40 & 0.04 & 0.51 \\
\bottomrule
\end{tabular}%
}
\caption{\textbf{Model responses across the full evaluation set.} Onset AUC compares the risk level at B with that at A1 within each of 601 eligible hazard clips; 0.5 indicates no directional separation. Warns at P is the warning rate 1.5\,s after the accident, on 618 eligible clips. Hazard-free warnings are measured on 1,233 windows. Counts precede model-specific exclusions of failed or invalid calls, which lower each count by at most five. Repeat $\kappa$ is Cohen's $\kappa$ on the warning decision between two calls on the same window, over those of the 801 repeated windows where both answers are valid. Arrows indicate the preferred direction. Bold values indicate the highest onset AUC and repeat agreement.}
\label{tab:all-clips}
\end{table}

\subsection{Do monitors warn because of the hazard?}
\label{sec:exp-twins}

\begin{table}[!t]
\centering
\scriptsize
\newcommand{\tabci}[1]{{\fontsize{5}{6}\selectfont [#1]}}
\setlength{\tabcolsep}{3pt}
\resizebox{\columnwidth}{!}{%
\begin{tabular}{@{}lccc@{}}
\toprule
\multicolumn{4}{@{}l}{\textbf{(a) Removal effect at A2, A1, and B (recolour minus removal, pp)}} \\
\midrule
Model & A2 (116) & A1 (184) & B (207) \\
\midrule
Claude Opus 5.5       & $+14.7$ \tabci{4.3, 24.1}   & $+18.5$ \tabci{10.3, 25.5}  & $+37.2$ \tabci{29.5, 44.4} \\
GPT-6 Astra           & $+8.6$ \tabci{1.7, 15.5}    & $+12.0$ \tabci{6.5, 17.9}   & $+47.3$ \tabci{40.6, 54.1} \\
Gemini 4 Argon        & $+6.0$ \tabci{$-$1.7, 13.8} & $+9.8$ \tabci{3.3, 16.3}    & $+38.6$ \tabci{30.9, 46.4} \\
Kimi-K3.1-Pro         & $+9.5$ \tabci{1.7, 17.2}    & $+6.0$ \tabci{$-$1.1, 13.6} & $+30.9$ \tabci{23.7, 38.6} \\
GLM-5.3-Flash         & $+18.1$ \tabci{6.9, 29.3}   & $+15.2$ \tabci{7.1, 23.9}   & $+25.1$ \tabci{17.4, 33.3} \\
Nemotron-3.5-Super-VL & $+3.4$ \tabci{$-$1.7, 9.5}  & $+3.8$ \tabci{$-$1.6, 9.2}  & $+12.1$ \tabci{6.3, 17.9}  \\
\addlinespace[4pt]
\midrule
\multicolumn{4}{@{}l}{\textbf{(b) Pre-impact window L ($n = 207$)}} \\
\midrule
Model & Removal (pp) & Risk drop & \begin{tabular}[b]{@{}c@{}}Disappears /\\appears (\%)\end{tabular} \\
\midrule
Claude Opus 5.5       & $+48.8$ \tabci{41.5, 56.0} & 1.13 & 50.7 / 1.9 \\
GPT-6 Astra           & $+62.3$ \tabci{55.1, 69.6} & 1.96 & 63.8 / 1.4 \\
Gemini 4 Argon        & $+55.1$ \tabci{47.3, 62.3} & 1.86 & 58.9 / 3.9 \\
Kimi-K3.1-Pro         & $+43.0$ \tabci{35.3, 50.7} & 1.22 & 47.3 / 4.3 \\
GLM-5.3-Flash         & $+32.4$ \tabci{24.2, 40.1} & 0.71 & 39.1 / 6.8 \\
Nemotron-3.5-Super-VL & $+16.4$ \tabci{10.6, 22.7} & 0.47 & 19.3 / 2.9 \\
\bottomrule
\end{tabular}%
}
\caption{\textbf{Removal lowers every model's warnings at B and L, and much less before the onset.} (a)~Removal effect (warning rate on the recoloured version minus that on the removal version, in pp) at A2, A1, and B across all twins with each decision time (clip count in header), with 95\% CIs from 2,000 resamples of twins. (b)~At L ($n = 207$): removal effect (pp), difference in mean risk (0--4 scale), and share of twins where a recolour warning disappears after removal versus appears.}
\label{tab:twin-results}
\label{tab:app-twin-effects}
\end{table}

\paragraph{Removal effect across decision times.} Removing the hazard lowers every model's warning rate at B and L, with all twelve 95\% confidence intervals excluding zero (Table~\ref{tab:twin-results}). At L, the removal effect ranges from $+16.4$ percentage points for Nemotron-3.5-Super-VL to $+62.3$ for GPT-6 Astra, and mean risk falls by 0.47 to 1.96 levels on the 0--4 scale. Almost every change runs in one direction: at L, a recolour warning disappears after removal on 19.3\% to 63.8\% of twins (Nemotron-3.5-Super-VL to Astra), while a new warning appears on at most 6.8\% (GLM-5.3-Flash). Across models, removal suppresses 49\% to 77\% of recolour warnings, from 81 of 164 for GLM-5.3-Flash to 132 of 172 for Astra. Before the annotated onset, removal effects are smaller and include zero for Kimi-K3.1-Pro and Nemotron-3.5-Super-VL at A1 ($n=184$) and for Gemini~4 Argon and Nemotron-3.5-Super-VL at A2 ($n=116$), consistent with the human finding that the hazard already drives some warnings before its annotated onset.

\paragraph{Residual warnings after removal.} On the 184 twins with an A1 window, residual warning rates after removal at L return to within four points of each model's pre-onset warning rate on the original clip at A1 for five models (for example, 46.7\% versus 46.7\% for GLM-5.3-Flash and 20.7\% versus 24.5\% for Astra), while Opus~5.5 warns on 42.9\% of removal windows at L against 55.4\% at A1. The editing controls change warnings little at L (Table~\ref{tab:app-twin-controls}): E0 minus original shifts warning rates by $-1.9$ to $+5.8$ points, identity minus E0 by $-3.9$ to $+1.9$ points, and recolour minus E0 by $-2.9$ to $+1.0$ points. The removal effect at L persists under tighter frame alignment: $+15.7$ to $+66.1$ points on the 115 twins within two source frames and $+15.4$ to $+59.6$ points on the 52 within one.

\begin{table}[!tb]
\centering
\scriptsize
\newcommand{\tabci}[1]{{\fontsize{5}{6}\selectfont [#1]}}
\setlength{\tabcolsep}{3pt}
\resizebox{\columnwidth}{!}{%
\begin{tabular}{@{}lccc@{}}
\toprule
Model & E0 $-$ original & Identity $-$ E0 & Recolour $-$ E0 \\
\midrule
Claude Opus 5.5       & $-1.9$ \tabci{$-$4.3, 0.5}  & $+1.9$ \tabci{$-$1.4, 5.3}  & $+1.0$ \tabci{$-$2.9, 4.8} \\
GPT-6 Astra           & $+0.5$ \tabci{$-$1.9, 2.9}  & $+1.9$ \tabci{$-$1.4, 5.3}  & $-1.9$ \tabci{$-$5.3, 1.4} \\
Gemini 4 Argon        & $+4.3$ \tabci{$-$0.5, 9.2}  & $+0.0$ \tabci{$-$4.8, 4.8}  & $-1.4$ \tabci{$-$6.8, 3.9} \\
Kimi-K3.1-Pro         & $+2.9$ \tabci{$-$2.4, 8.3}  & $+1.0$ \tabci{$-$4.8, 6.8}  & $-2.9$ \tabci{$-$8.7, 2.9} \\
GLM-5.3-Flash         & $+3.4$ \tabci{$-$2.4, 8.7}  & $-3.9$ \tabci{$-$10.6, 2.4} & $+0.0$ \tabci{$-$6.3, 6.3} \\
Nemotron-3.5-Super-VL & $+5.8$ \tabci{$-$0.5, 12.6} & $-2.4$ \tabci{$-$7.7, 3.4}  & $+0.5$ \tabci{$-$5.3, 6.3} \\
\bottomrule
\end{tabular}%
}
\caption{\textbf{The control versions change warnings little.} Differences in warning rate at L between versions of the same 207 twins, in percentage points, with 95\% CIs from 2,000 resamples of twins. E0 is the original re-encoded to the editor's format; identity is the editor's output with an instruction to change nothing.}
\label{tab:app-twin-controls}
\end{table}

\subsection{What explains the warnings that remain after removal?}
\label{sec:exp-audit}

\begin{table}[!t]
\centering
\scriptsize
\newcommand{\tabci}[1]{{\fontsize{5}{6}\selectfont [#1]}}
\setlength{\tabcolsep}{3pt}
\resizebox{\columnwidth}{!}{%
\begin{tabular}{@{}lcccc@{}}
\toprule
\multicolumn{5}{@{}l}{\textbf{(a) Removal effect at L: recolour minus removal (pp)}} \\
\midrule
Model & All (100) & Removed (81) & Clean (48) & Failed (19) \\
\midrule
Claude Opus 5.5       & $+50.0$          & $+61.7$          & $+68.8$          & $+0.0$ \tabci{$-15.8$, $+15.8$}  \\
GPT-6 Astra           & $\mathbf{+61.0}$ & $\mathbf{+76.5}$ & $\mathbf{+77.1}$ & $-5.3$ \tabci{$-15.8$, $+0.0$}   \\
Gemini 4 Argon        & $+54.0$          & $+69.1$          & $+70.8$          & $-10.5$ \tabci{$-31.6$, $+10.5$} \\
Kimi-K3.1-Pro         & $+39.0$          & $+51.9$          & $+52.1$          & $-15.8$ \tabci{$-31.6$, $+0.0$}  \\
GLM-5.3-Flash         & $+30.0$          & $+40.7$          & $+47.9$          & $-15.8$ \tabci{$-42.1$, $+10.5$} \\
Nemotron-3.5-Super-VL & $+16.0$          & $+17.3$          & $+20.8$          & $+10.5$ \tabci{$-10.5$, $+31.6$} \\
\addlinespace[4pt]
\midrule
\multicolumn{5}{@{}l}{\textbf{(b) Blinded observer agreement and shared / model warnings}} \\
\midrule
Model & $\kappa$ $\uparrow$ & Clean (48) & Altered (33) & Failed (19) \\
\midrule
Claude Opus 5.5       & 0.37          & 3 / 12 & 0 / 13 & 13 / 16 \\
GPT-6 Astra           & \textbf{0.58} & 2 / 7  & 0 / 3  & 12 / 13 \\
Gemini 4 Argon        & \textbf{0.58} & 3 / 5  & 0 / 5  & 11 / 13 \\
Kimi-K3.1-Pro         & 0.50          & 2 / 5  & 0 / 5  & 10 / 11 \\
GLM-5.3-Flash         & 0.21          & 1 / 18 & 1 / 13 & 12 / 16 \\
Nemotron-3.5-Super-VL & 0.07          & 0 / 5  & 0 / 3  & 3 / 3   \\
\bottomrule
\end{tabular}%
}
\caption{\textbf{The removal effect is larger on successful removals and absent on failed ones, where the observer's shared warnings concentrate.} (a)~Difference in warning rate between the recolour and removal versions at L, in percentage points (pp), across the audited strata of 100 random twins (brackets in Failed give 95\% CIs from 2,000 resamples of twins). (b)~Cohen's $\kappa$ between each model and the blinded observer across all 100 removal windows at L, and the number of model removal warnings that the observer shares, out of the model's removal warnings in each stratum. Bold marks the largest removal effect in each column and the highest observer $\kappa$.}
\label{tab:app-audit-combined}
\end{table}

\paragraph{Edit audit and blinded observer.} To determine how often removal succeeds and whether models' remaining warnings reflect visible evidence that a person would also act on, we evaluate a random sample of 100 twins at L through two independent studies (full protocols, descriptions, and frame strips appear in Appendix~\ref{app:twin-controls}). First, two independent coders, blind to all model and observer answers, compare all 31 frames of each removal window at L against the original; they agree on whether the target road user is gone in 96 of 100 twins (Cohen's $\kappa=0.87$) and on whether the rest of the scene is preserved in 71\% ($\kappa=0.35$), with an author adjudicating the 34 disagreements on a page hiding all model and observer outputs. The target is removed in 81 twins (48 \emph{clean} with the rest of the scene preserved, and 33 where other road users or scene elements were also altered) and remains visible or replaced in 19 \emph{failed} removals. Second, a fresh blinded observer rates the 100 removal windows at L interleaved with 20 unedited calibration windows in the same $1440\times608$ format, locking each warning and risk decision before describing what prompted every warning. The observer agrees with the three-rater panel on 19 of the 20 calibration windows (9 of 10 unanimous-warning L windows and 0 of 10 A1 windows) and warns on 19 of the 100 removal windows, never citing an editing artefact.

\begin{figure*}[!tb]
\centering
\includegraphics[width=\textwidth]{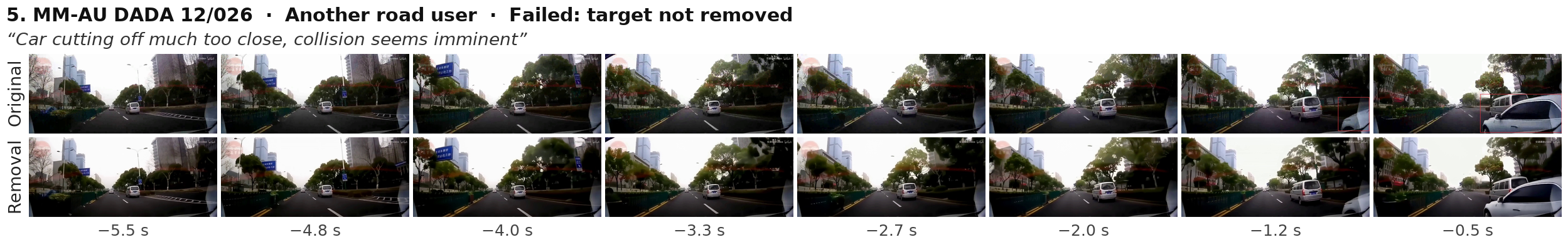}
\vspace{3pt}
\includegraphics[width=\textwidth]{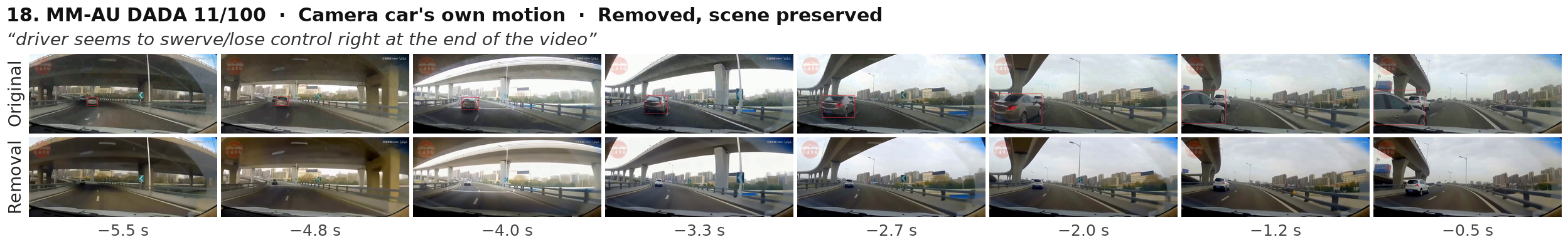}
\caption{\textbf{The two main reasons a blinded human observer still warns after removal (17 of 19 warnings).} Each strip shows eight of the 31 frames at L (upper row: original with target boxed in red; lower row: removal version shown to models and the observer). \textbf{Top (failed removal; 13 of 19):} the cutting-in car remains visible (observer: ``\emph{Car cutting off much too close, collision seems imminent}''). \textbf{Bottom (clean removal, camera car's own path; 4 of 6 warnings on successful removals):} the car ahead is cleanly erased, leaving the camera car's recorded evasive swerve unexplained (observer: ``\emph{driver seems to swerve/lose control right at the end of the video}''). The remaining two warnings on successful removals involve another road user; all 19 strips and verbatim descriptions appear in Appendix~\ref{app:twin-controls}.}
\label{fig:observer-archetypes}
\end{figure*}

\paragraph{Results by edit outcome.} Stratifying by edit outcome sharpens both the causal effect and the diagnosis of residual warnings (Table~\ref{tab:app-audit-combined}). On the 81 twins where the target is removed, the removal effect at L rises above the all-100 estimate for every model, reaching $+76.5$ percentage points for GPT-6 Astra ($+77.1$ on the 48 clean removals), $+69.1$ for Gemini~4 Argon ($+70.8$ clean), and $+61.7$ for Claude Opus~5.5 ($+68.8$ clean). Conversely, on the 19 failed removals, the removal effect disappears for every model ($-15.8$ to $+10.5$ points, with every 95\% CI spanning zero), providing an internal negative control. The blinded observer warns on 13 of the 19 failed removals (68\%, as in Figure~\ref{fig:observer-archetypes}, top), compared with only 6 of the 81 successful removals (7\%). Those six warnings fall on 5 of the 48 clean removals and 1 of the 33 scene-altered twins. Crucially, the models separate into two groups on these audited twins: for Astra, Argon, and Kimi-K3.1-Pro, more than half of all post-removal warnings (11 to 13 of 21 to 23) occur on the 19 failed edits, where the observer also warns on 10 to 12; on the 81 successful removals, each warns on only 10 twins (sharing 2 to 3 with the observer). By contrast, Opus~5.5 and GLM-5.3-Flash still warn on 25 and 31 of the 81 successful removals, of which the observer shares only 3 and 2. The same pattern holds on the 96 twins where both coders agree on the target without author adjudication.

\paragraph{Interpretation.}
Together, the 207-twin comparison and the 100-twin audit show that every model responds strongly to the targeted road user where removal succeeds, while imperfect edits in the full pool attenuate the unstratified removal estimates. Moreover, the observer's follow-up descriptions explain why a small share of warnings remains warranted even after a clean removal: erasing the road user removes the cause of the incident, not the camera car's recorded evasive manoeuvre. Four of the observer's six warnings on successful removals cite the camera car's own path (such as an unexplained late swerve toward a sign, exit, or curb; Figure~\ref{fig:observer-archetypes}, bottom, and Appendix~\ref{app:twin-controls}). Beyond those rare residual cues and failed edits, however, the elevated post-removal warning rates of Opus~5.5 and GLM-5.3-Flash reflect the same background over-sensitivity seen prior to the hazard onset.

%% file: conclusion.tex
\section{Conclusion}
\label{sec:conclusion}

We study when and why frontier multimodal models warn of danger in real video, comparing their judgments with those of people who see the same frames and with their own judgments on counterfactual footage from which the implicated road user is removed. We find that people often warn before the annotated hazard onset, while all six monitors distinguish the onset from the lead-up less reliably than the mean individual-rater reference. Their warning behavior and underlying risk ratings split into two failure modes: some warn frequently during benign lead-up, while others continue to withhold warnings even half a second before impact. Removing the implicated road user suppresses 49--77\% of pre-impact warnings across all 207 twins, and a two-coder edit audit paired with a blinded human observer on 100 twins confirms that the removal effect grows to 17--77 percentage points where removal succeeds and disappears on failed edits, while showing that over-sensitive models continue to warn on successful removals without human-corroborated cues. Since a warning helps only when it comes in time and for the right reason, video safety monitors should be evaluated against people viewing identical frames and against audited counterfactual footage testing whether warnings depend on the hazard itself.

%% file: llm-usage.tex
\section*{LLM Usage Statement}

We used Gemini 4 Argon to assist with literature discovery, surveying relevant prior work, and identifying related research. GPT-6 was used for final-stage language polishing and proofreading, including checks for grammatical and typographical errors. The empirical study evaluates six vision-language models: Claude Opus 5.5, GPT-6 Astra, Gemini 4 Argon, Kimi-K3.1-Pro, GLM-5.3-Flash, and Nemotron-3.5-Super-VL. We used Meta Muse Spark-Open as an independent vision-language judge to assess agreement with our rule-based procedure for identifying the road user responsible for each accident. We used Black Forest Labs FLUX Video-Edit to construct the counterfactual video conditions described in the experimental methodology. We take full responsibility for the study design, analysis, interpretation, and accuracy of the manuscript.

%% file: appendix.tex
\section{Counterfactual Temporal Twin Construction}
\label{app:twins}
\label{app:counterfactual-quality}

\begin{figure*}[t]
\centering
\includegraphics[width=\textwidth]{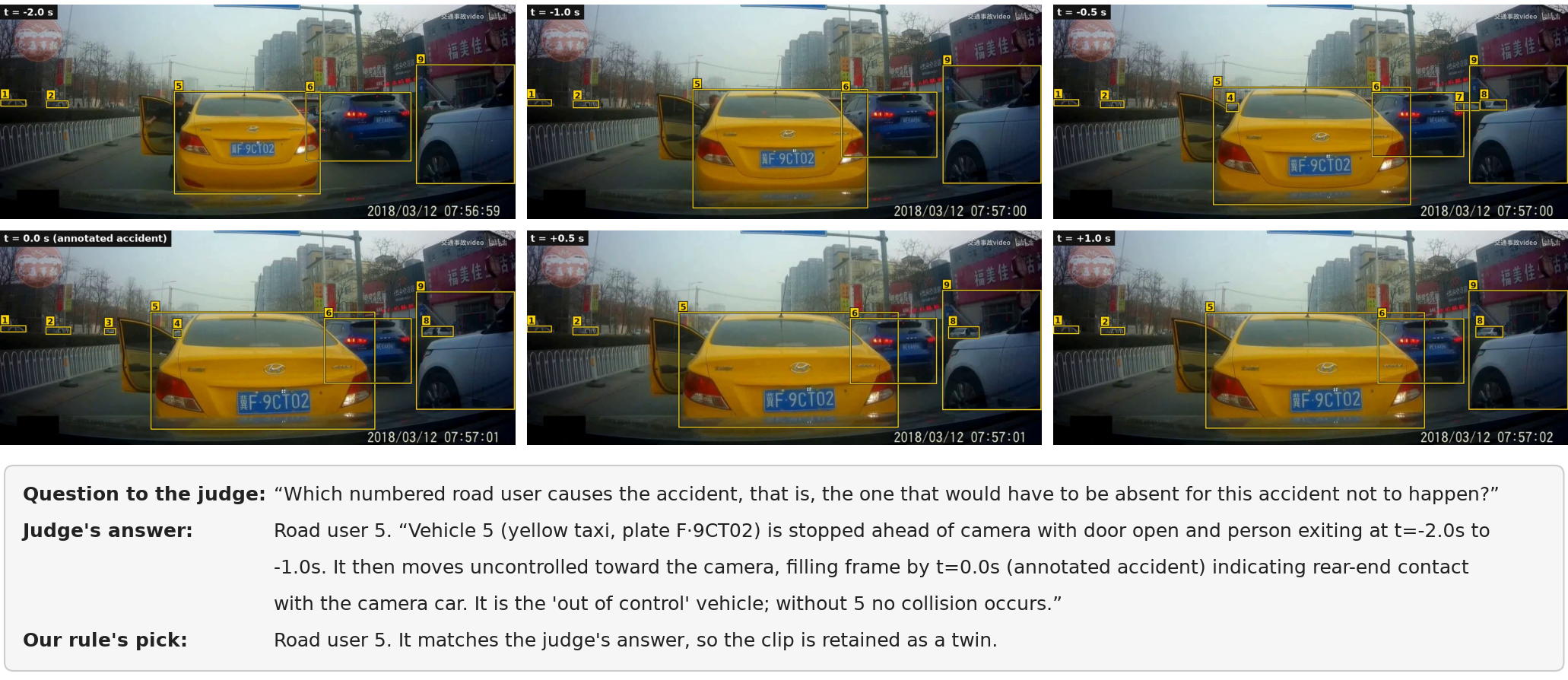}
\caption{\textbf{What the judge sees.} The six frames sent to the judge for one MM-AU DADA clip (accident category 11, video 231): 2.0, 1.0 and 0.5\,s before the annotated accident, the accident, and 0.5 and 1.0\,s after it. Road users with a box within 0.5\,s of the accident are numbered, with the same number in every frame. The judge also receives the dataset's accident description and cause, and is not told which road user our rule selected. In this clip, the judge and our rule agree on road user 5, the yellow car, so the clip is retained as a twin. This twin is the first panel of Figure~\ref{fig:twin-examples}.}
\label{fig:app-judge-sheet}
\end{figure*}

\paragraph{Hazard identification and judge verification.}
To construct counterfactual temporal twins (Section~\ref{sec:proto-twins}), we draw from DADA-2000 accident clips in MM-AU and in PaSBench-Video's DADA subset, for which MM-AU provides hand-corrected road-user bounding boxes every fifth frame~\cite{fang2024abductive,fang2019dada}. We exclude single-vehicle accidents that lack an external road user to remove. For each candidate clip, an automated rule first parses the dataset's accident description to identify the implicated road-user class and selects the track of that class with the largest box in the last 0.5\,s before the accident, widening the window to 1\,s and then 2\,s, and then to related classes, if no such track is found. Because multiple road users of the same class can appear near the camera car, we verify each candidate independently using Meta Muse-Spark-Open, which is outside our evaluated model panel. The judge receives six frames around the accident with all annotated road-user boxes numbered (Figure~\ref{fig:app-judge-sheet}), together with the dataset's accident description and cause, without being told which box our rule selected, and is asked which road user would have to be absent for the accident not to occur. We retain only clips on which the automatic rule and the independent judge select the same road user; disagreements and ambiguous cases are set aside rather than forced into the sample. This procedure yields 207 verified twins (193 from our MM-AU split and 14 from PaSBench-Video's DADA subset).

\begin{figure*}[tp]
\centering
\begin{minipage}[t]{0.45\textwidth}
\centering
\includegraphics[width=\linewidth]{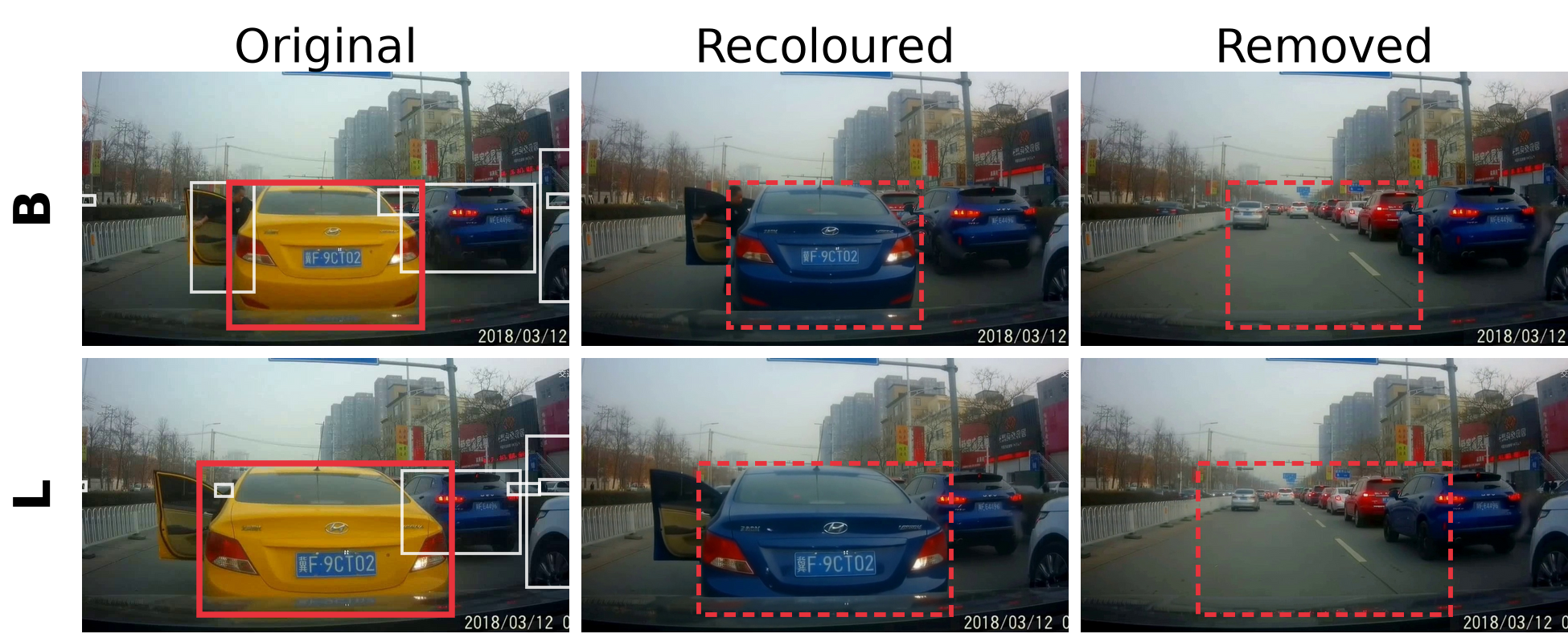}
\end{minipage}\hfill
\begin{minipage}[t]{0.45\textwidth}
\centering
\includegraphics[width=\linewidth]{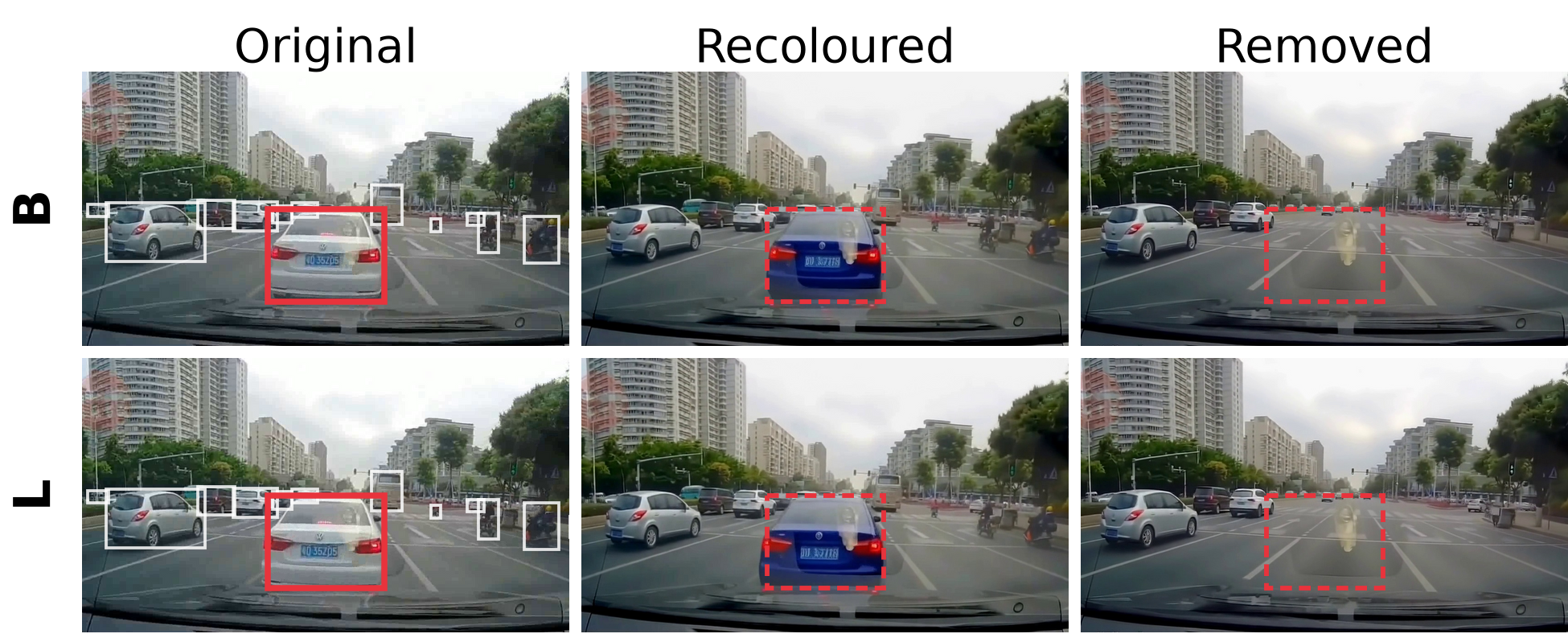}
\end{minipage}

\vspace{5pt}
\begin{minipage}[t]{0.45\textwidth}
\centering
\includegraphics[width=\linewidth]{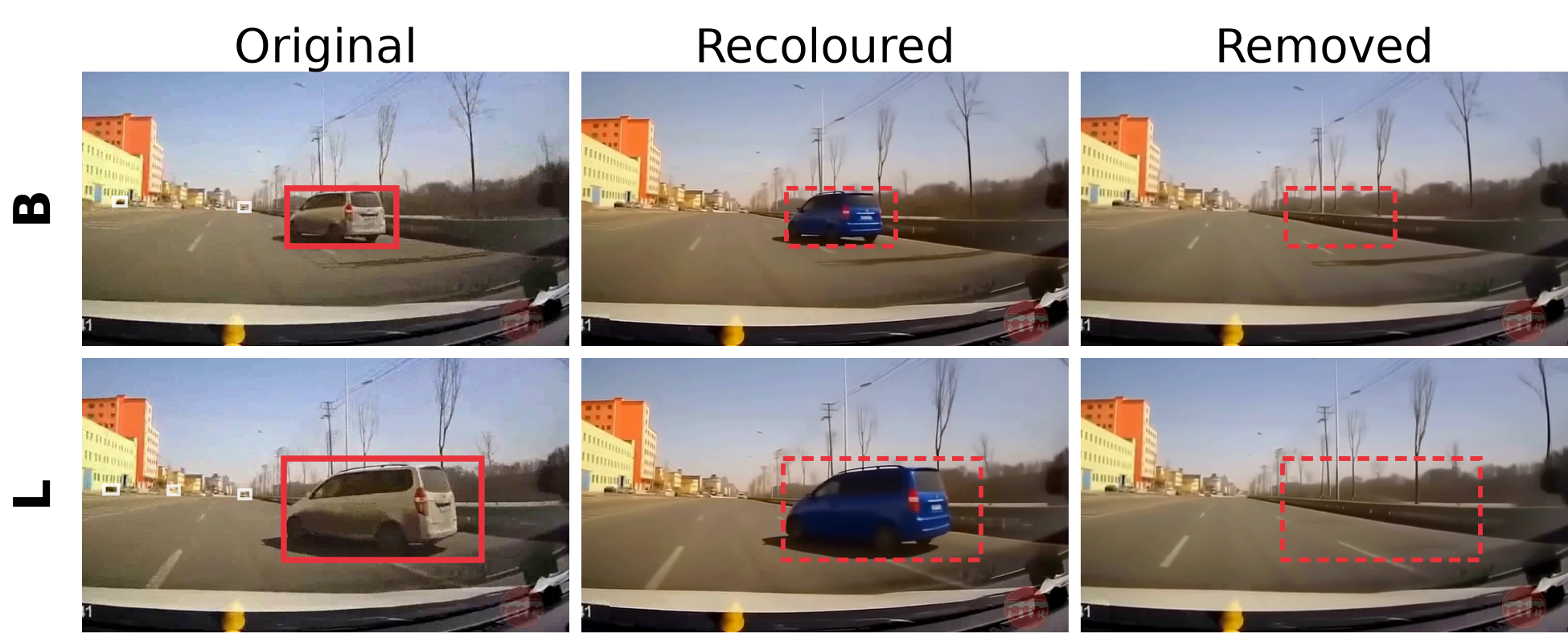}
\end{minipage}\hfill
\begin{minipage}[t]{0.45\textwidth}
\centering
\includegraphics[width=\linewidth]{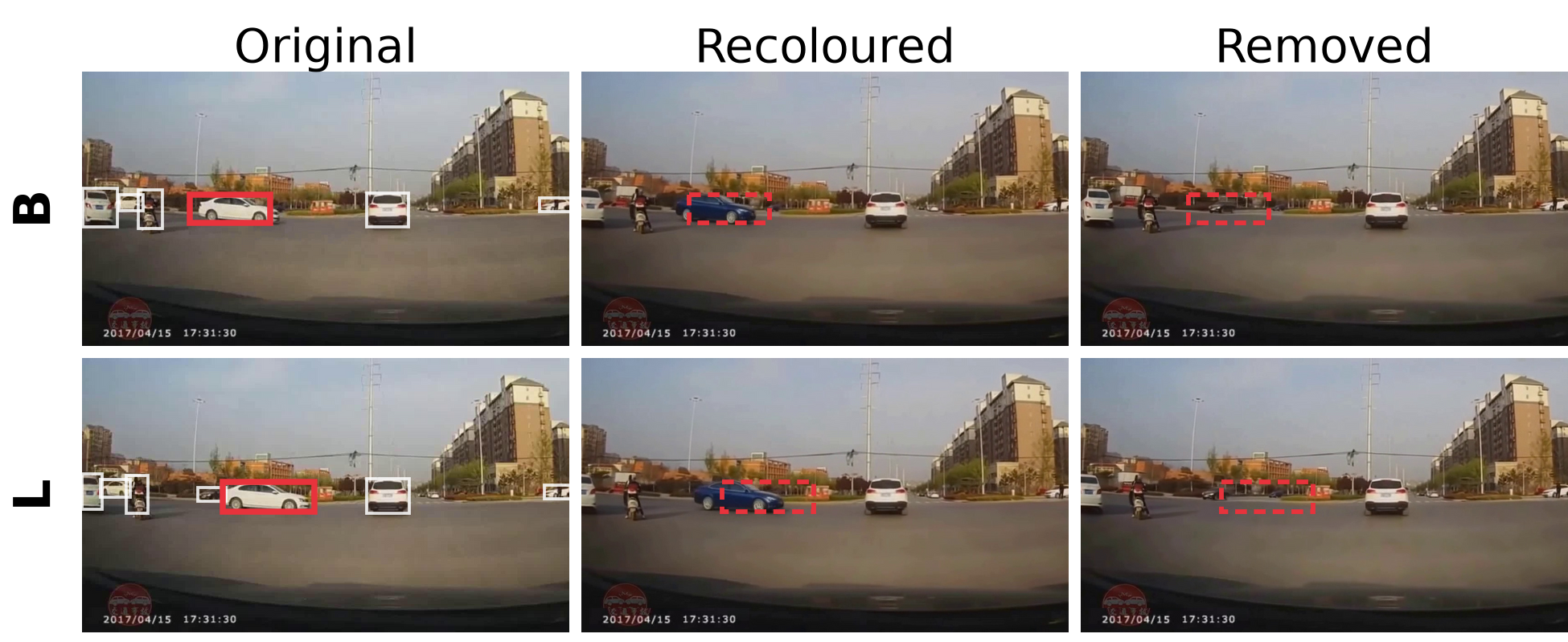}
\end{minipage}

\vspace{5pt}
\begin{minipage}[t]{0.45\textwidth}
\centering
\includegraphics[width=\linewidth]{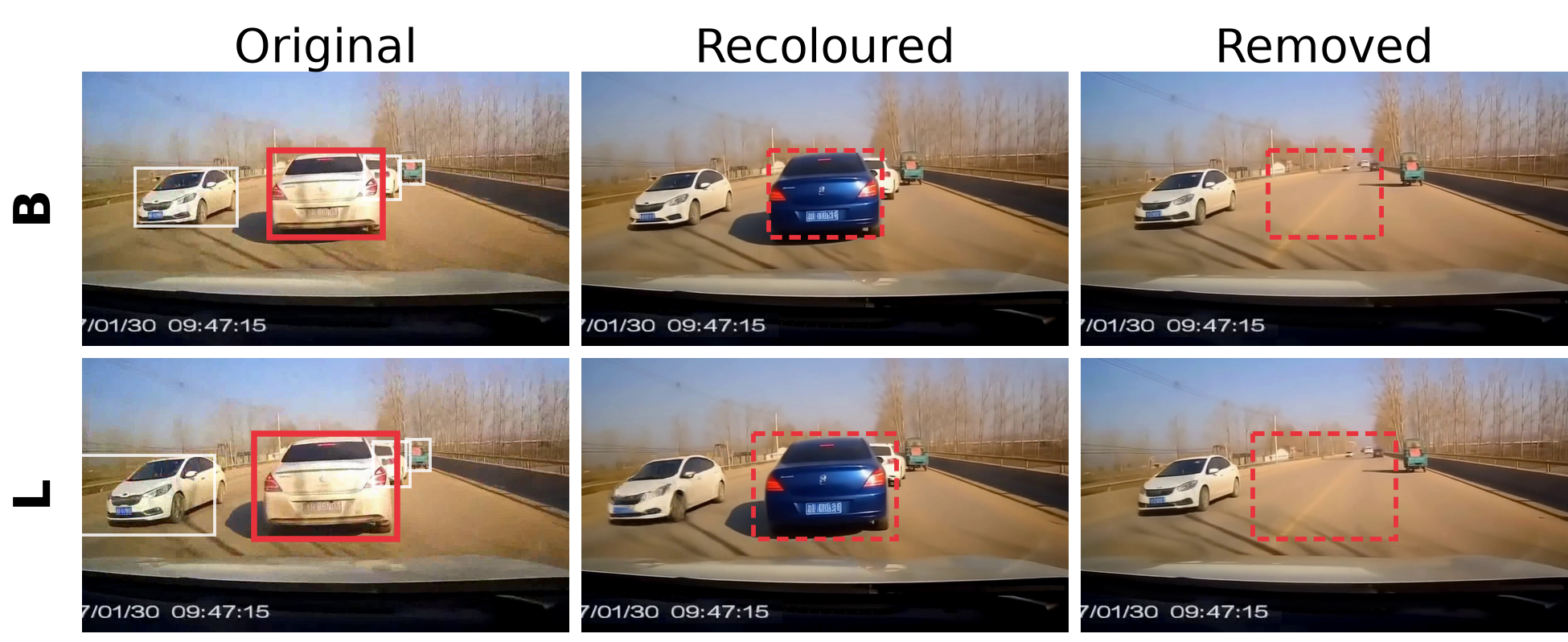}
\end{minipage}\hfill
\begin{minipage}[t]{0.45\textwidth}
\centering
\includegraphics[width=\linewidth]{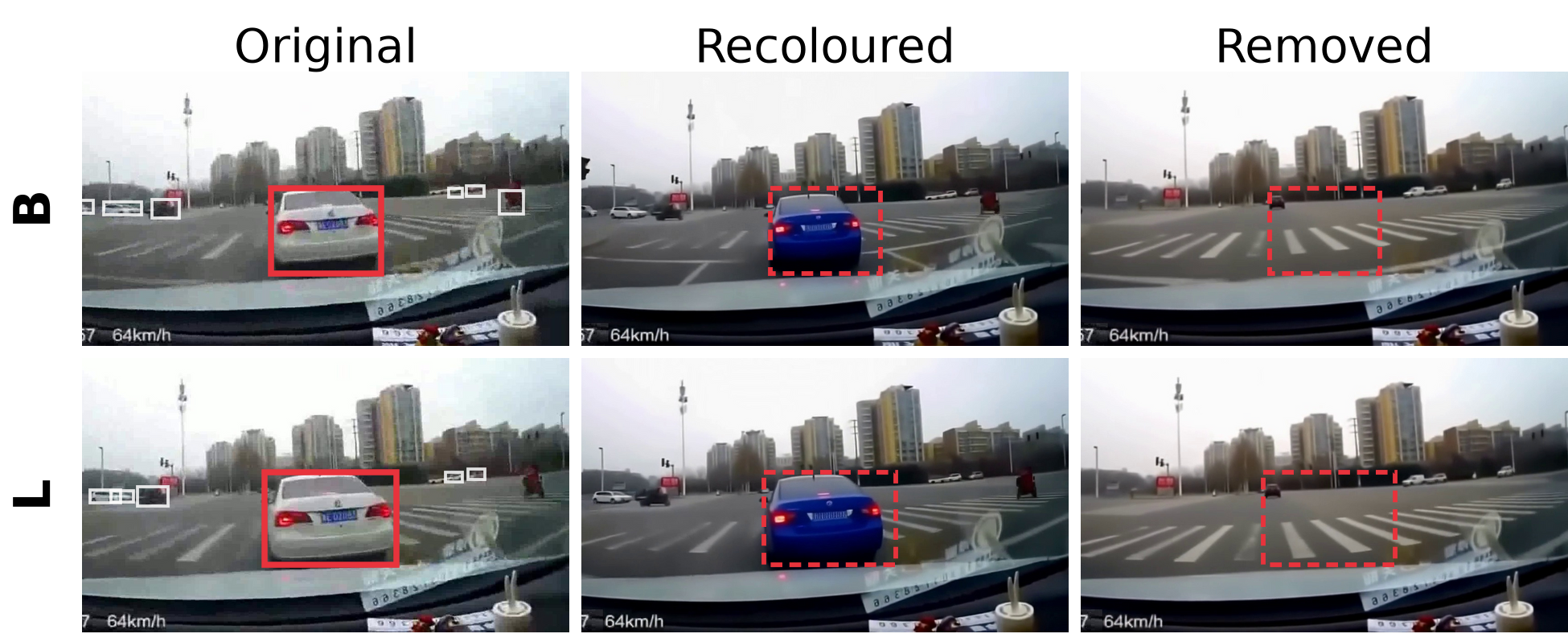}
\end{minipage}

\vspace{5pt}
\begin{minipage}[t]{0.45\textwidth}
\centering
\includegraphics[width=\linewidth]{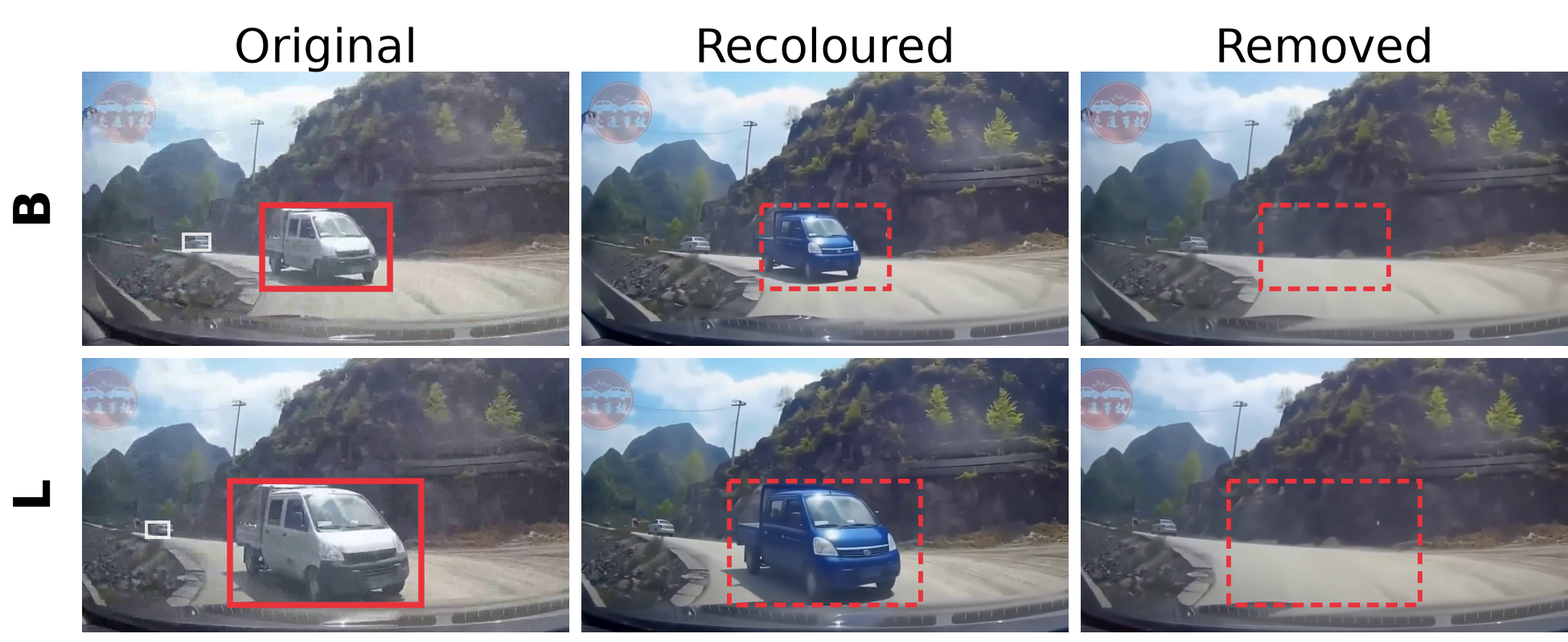}
\end{minipage}\hfill
\begin{minipage}[t]{0.45\textwidth}
\centering
\includegraphics[width=\linewidth]{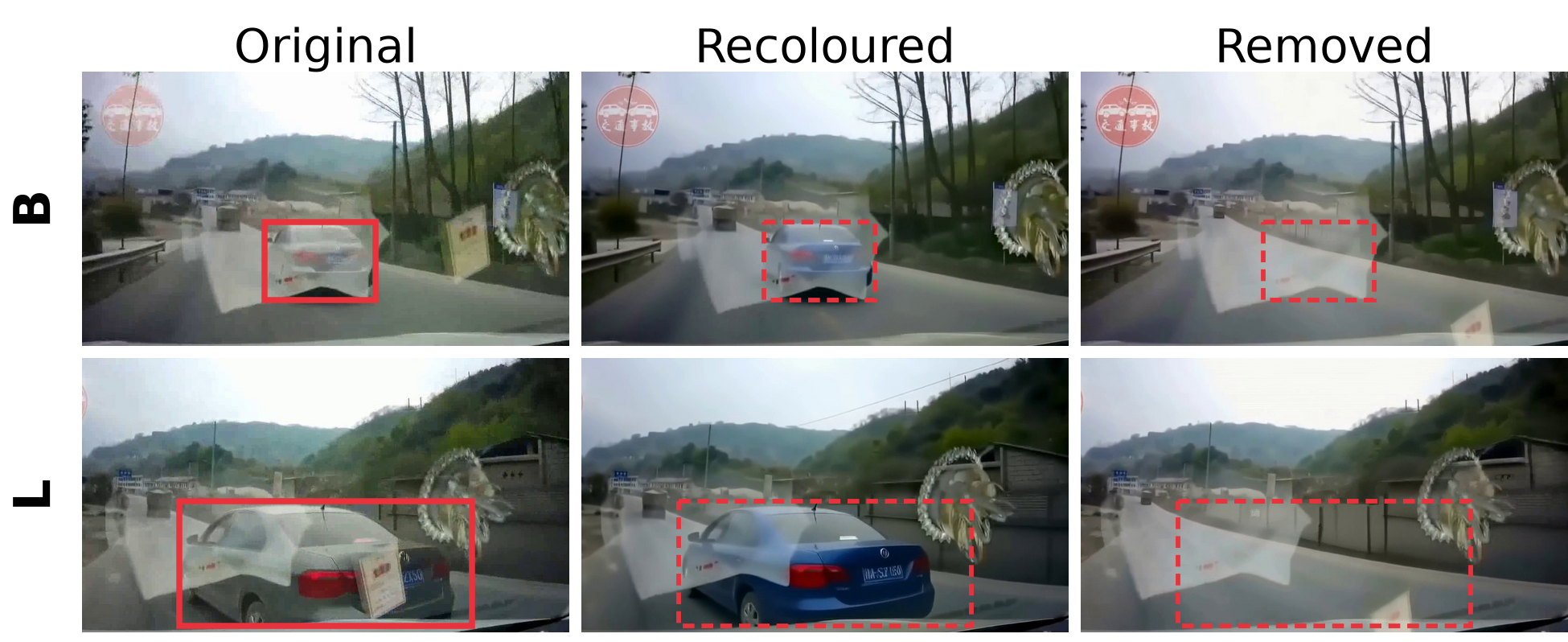}
\end{minipage}

\vspace{5pt}
\begin{minipage}[t]{0.45\textwidth}
\centering
\includegraphics[width=\linewidth]{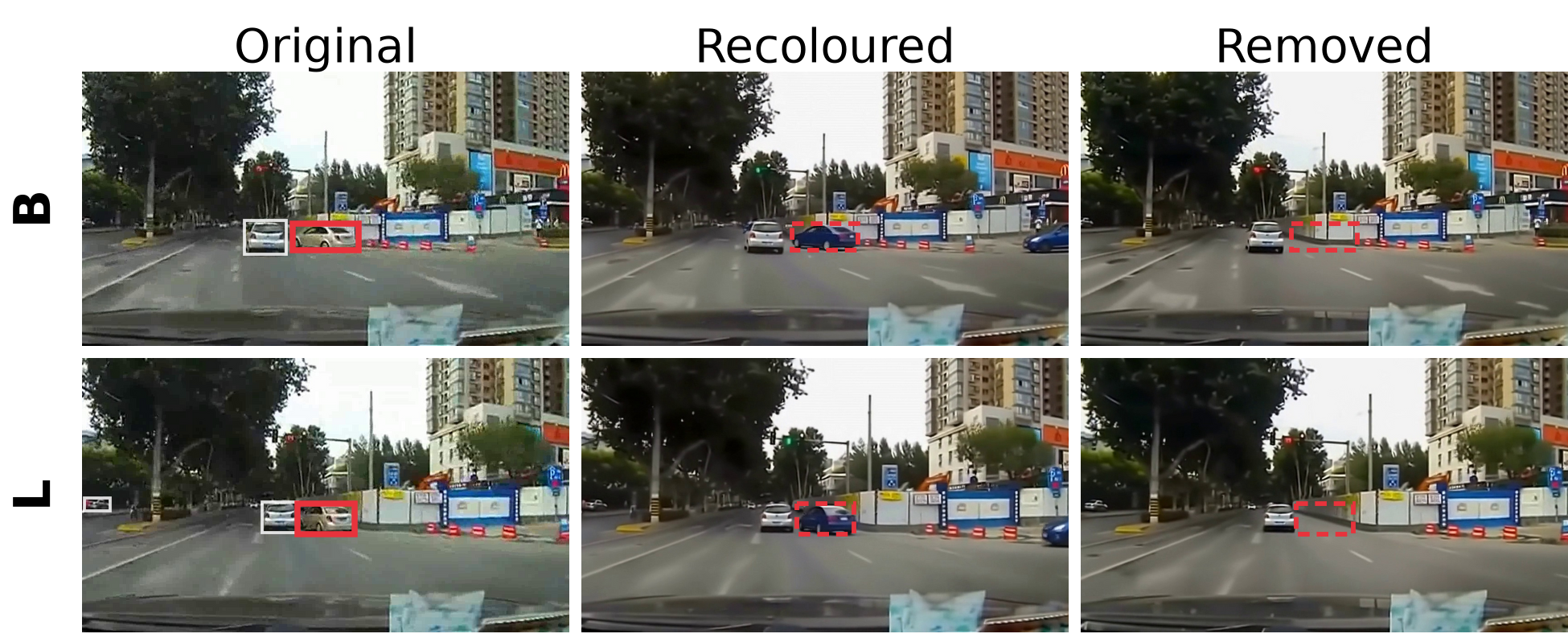}
\end{minipage}\hfill
\begin{minipage}[t]{0.45\textwidth}
\centering
\includegraphics[width=\linewidth]{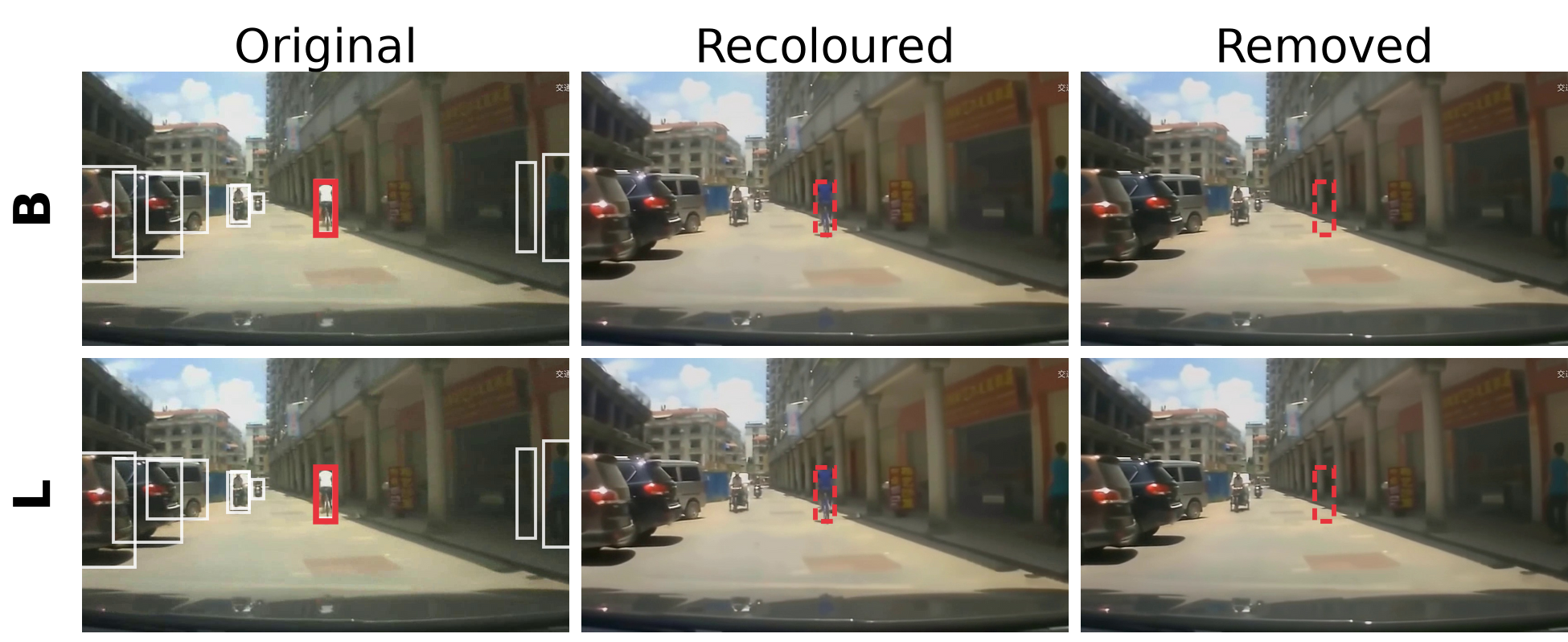}
\end{minipage}

\vspace{5pt}
\begin{minipage}[t]{0.45\textwidth}
\centering
\includegraphics[width=\linewidth]{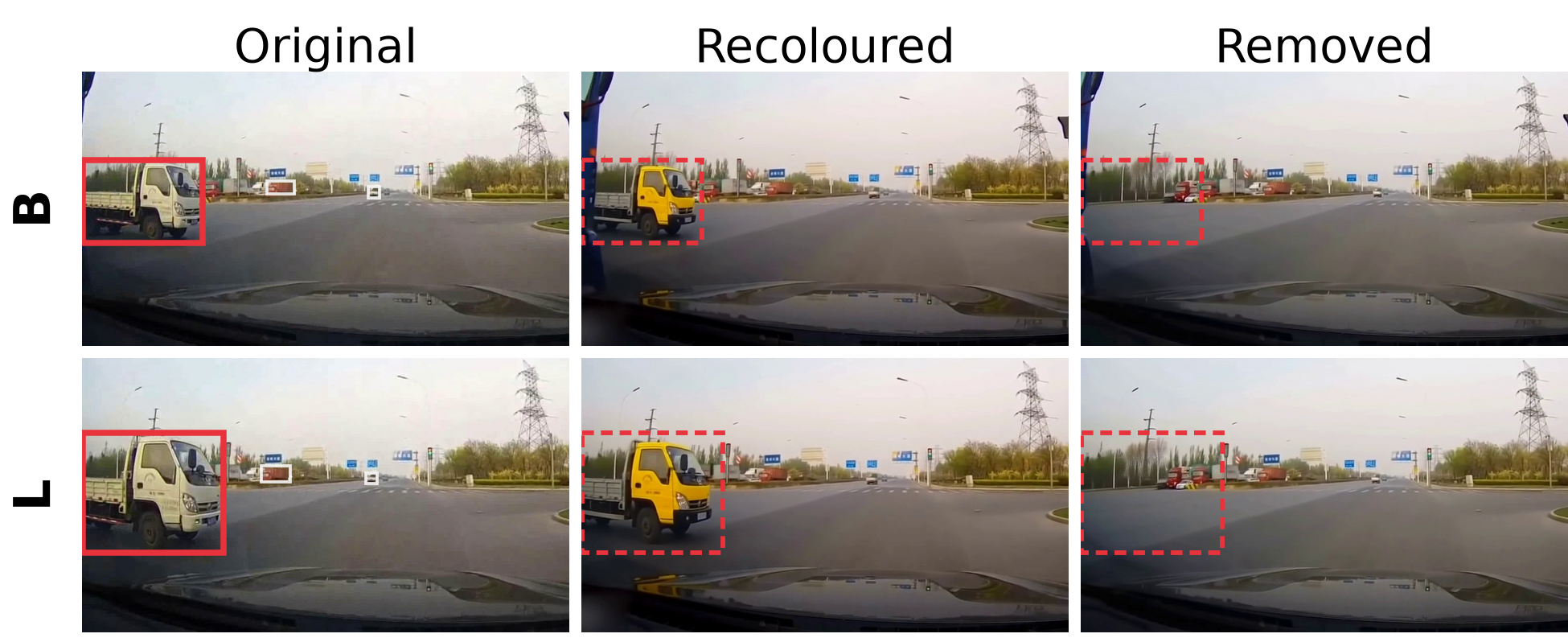}
\end{minipage}\hfill
\begin{minipage}[t]{0.45\textwidth}
\centering
\includegraphics[width=\linewidth]{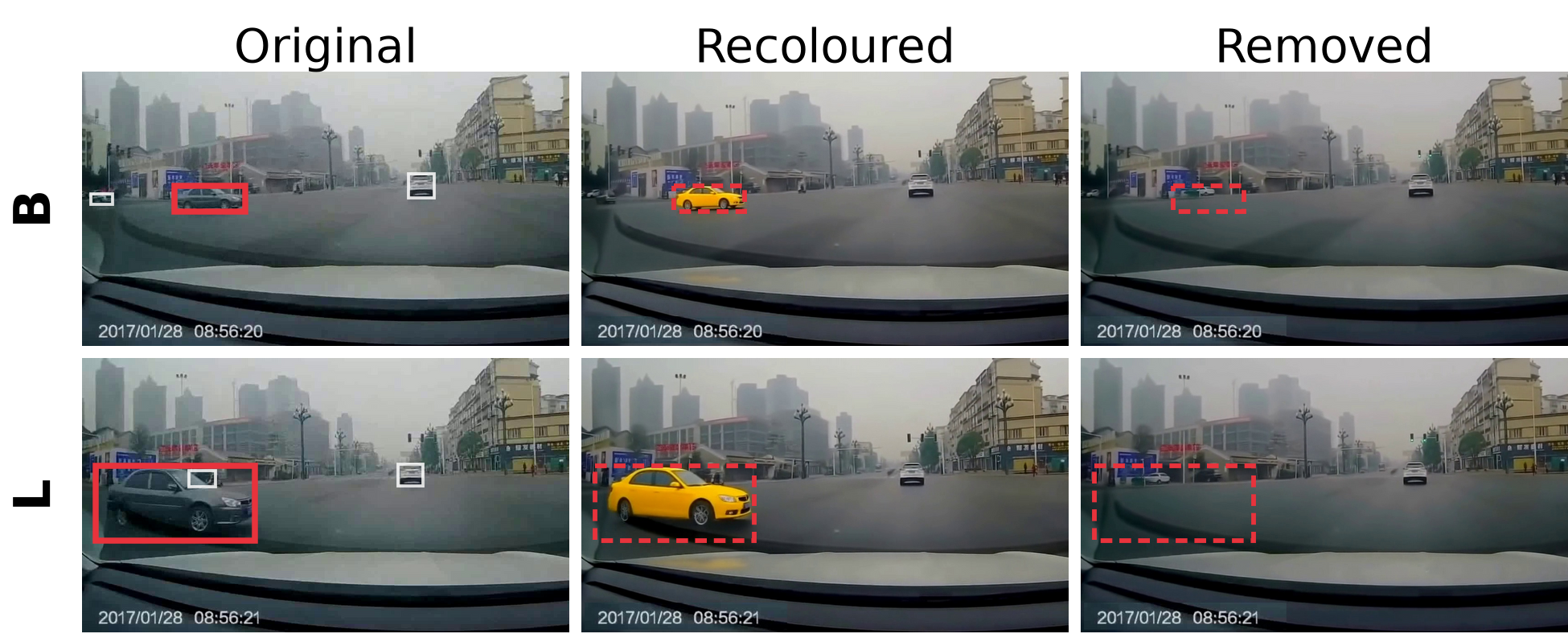}
\end{minipage}
\caption{\textbf{Examples of recoloured and removal twins.} Each panel shows one source clip at B (0.5\,s after the annotated onset) and at L (0.5\,s before the accident). The three images in a row are the last frame of the 31-frame window given to the models, for the original clip, the recoloured version and the removal version, cropped identically around the implicated road user. The solid red box marks the implicated road user's annotated box in the original, and the dashed box marks the same location in the edited versions. White boxes mark the other annotated road users. The examples were chosen for illustration and are not a random sample: the first five rows use the dark-blue recolour, which applies to 25 of the 207 twins, and the last row shows two of the other 182, which were recoloured bright yellow.}
\label{fig:twin-examples}
\end{figure*}

\paragraph{Segment boundaries and editing prompts.}
Let $r$ denote the annotated hazard onset and $a$ the accident time; throughout the appendices, each decision point marks the final frame of a 5.0\,s (31-frame) observation window at A2 (2.5\,s before the onset, $r - 2.5$\,s), A1 (1.0\,s before the onset, $r - 1.0$\,s), B (0.5\,s after the onset, $r + 0.5$\,s), B2 (1.5\,s after the onset, $r + 1.5$\,s), L (0.5\,s before the accident, $a - 0.5$\,s), or P (1.5\,s after the accident, $a + 1.5$\,s; full sampling details appear in Appendix~\ref{app:frame-sampling}). We edit only the temporal segment required to form the evaluated 5.0\,s observation windows up to L ($a - 0.5$\,s), so that no edited window contains the accident itself (Section~\ref{sec:proto-twins}). Across the 207 twins, 116 have sufficient pre-onset footage to provide all four decision points A2, A1, B, and L; 68 provide A1, B, and L; and 23 provide B and L (707 clip-time windows in total, as B and L coincide on 7 twins with $a - r = 1.0$\,s). Each segment is processed with FLUX Video Edit~\cite{bfl2026fluxvideoedit} under three prompt conditions: (1)~an \emph{identity} edit instructing the model to preserve the video without changes; (2)~a \emph{recolour} edit instructing the editor to change the target road user to a contrasting shade (dark blue or bright yellow, whichever is farther in CIELAB colour space from the target's median colour; Figure~\ref{fig:twin-examples}) while preserving its trajectory and all other road users; and (3)~a \emph{removal} edit instructing the editor to delete the target road user throughout the segment and reconstruct the background behind it while preserving all other road users. Both the recolour and removal prompts specify the target's class, the dataset's accident description, its timestamped MM-AU boxes and derived positions over the segment (subsampling boxes at random when a prompt exceeds FLUX's 4,096-character limit), and any nearby road user to keep, treating a rider and their bicycle or motorcycle as one target. We also create a re-encoded control version (E0) by transcoding the original segment to match the editor's $1440\times608$, 24\,fps output format without generative editing. In the 96 twins with a burned-in clock, we paste the original clock region into the frames of every version.

\paragraph{Frame alignment and edit localization.}
The original DADA-2000 clips run at 30\,fps, whereas FLUX Video Edit outputs video at 24\,fps and $1440\times608$ resolution. Our $1/6$\,s (6\,fps) window sampling corresponds to every fifth frame of the 30\,fps original and every fourth frame of the 24\,fps edited output. Because generative video editing can introduce slight temporal drift, matching frames by nominal index alone can misalign ego-motion between arms. For each sampled timestamp, we compare the nominal edited frame and the four frames on either side against the original source frame, selecting the one whose background best matches when the target's boxes and the clock region are masked. Across all 41,337 sampled frames in the edited versions, 97.7\% match within one original source frame ($\le 33$\,ms), and only 931 frames (2.3\%) are displaced by more than one original frame. Figure~\ref{fig:app-twin-construction} (Section~\ref{sec:proto-twins}) shows the resulting pixel-change localization inside and outside the target road user's bounding box across the 176 twins with a box at B.

\section{Counterfactual Controls, Observer Study, and Edit Audit Details}
\label{app:twin-controls}

\paragraph{Control versions and preliminary rationale check.}
Per-model control differences at L across all 207 twins appear in Table~\ref{tab:app-twin-controls} (Section~\ref{sec:exp-twins}); every 95\% confidence interval includes zero. Before running the 100-twin observer and edit-audit studies summarized in Section~\ref{sec:exp-audit}, we also inspected residual warnings by eye. On the 78 twins whose target is not a car, the stated reason names the removed road user's class in 91 answers, on 48 removal windows. We inspected 26 of these windows frame by frame. Only 2 refer to a road user absent from the frames; both come from Claude Opus~5.5 at L, and both are hedged. In 22 of the other 24, a road user of that class was visible in the frames: the editor left or replaced the target (8), another road user of the same class remained in view (7), or the model's description could refer to either (7). One further warning referred to the camera car, and the remaining case was ambiguous.

\paragraph{Blinded observer protocol and overall agreement.}
For the 100-twin observer study reported in Section~\ref{sec:exp-audit} (Table~\ref{tab:app-audit-combined}), we drew 100 of the 207 twins uniformly without replacement with a fixed seed, and interleaved 20 unedited calibration windows from 20 other clips rated in Appendix~\ref{app:human-reference}: 10 at L, where all three raters warned, and 10 at A1, where none did. These 20 windows calibrate the observer against the panel and are excluded from every removal estimate. All 120 items were shown in the editor's output format ($1440\times608$, using the re-encoded E0 version for the unedited windows) in one random order after three practice items, so that no clip appeared twice and image format did not reveal which windows were edited. The rating page and rules followed Appendix~\ref{app:rating-page}, with no box, original clip, version label, or accident timing. Both answers became final when the observer clicked Next. Only then, and only after a warning, did the page ask ``What did you see that made you warn? Name the road user or scene feature, and what was happening.'' (with ``I can't identify a specific cue'' allowed). No question followed a decision not to warn. On the 20 calibration windows, the observer agreed with the three-rater panel on 19 of 20 items, warning on 9 of the 10 unanimous-warning L windows (mean risk 2.4, against 3.0 for the panel) and on none of the 10 A1 windows. Across all 100 removal windows at L, the six models warn on 11 (Nemotron-3.5-Super-VL) to 47 (GLM-5.3-Flash) windows, and most of these warnings already appear on the recolour version (from 7 of 11 for Nemotron-3.5-Super-VL to 38 of 41 for Claude Opus~5.5). Overall, of the removal windows on which each model warns, the observer also warns on 61\% for GPT-6 Astra (14 of 23; 95\% CI 41 to 78\%) and Gemini~4 Argon (14 of 23; 41 to 78\%), 57\% for Kimi-K3.1-Pro (12 of 21; 37 to 76\%), 39\% for Claude Opus~5.5 (16 of 41; 26 to 54\%), 30\% for GLM-5.3-Flash (14 of 47; 19 to 44\%), and 27\% for Nemotron-3.5-Super-VL (3 of 11; 10 to 57\%). Table~\ref{tab:app-edit-audit} gives the full two-way breakdown of the adjudicated edit-audit codes across the 100 twins; in 29 of the 33 twins where the target is removed but the scene is altered, the editor also removed other road users. The observer comparison measures agreement with this observer, not a false-alarm rate.

\begin{table}[t]
\centering
\small
\setlength{\tabcolsep}{5pt}
\begin{tabular}{@{}lccc@{}}
\toprule
Target road user & Scene preserved & Scene altered & All \\
\midrule
Removed     & 48 (clean) & 33 & 81 \\
Not removed & 11         & 6  & 17 \\
Replaced    & 1          & 1  & 2  \\
\midrule
All         & 60         & 40 & 100 \\
\bottomrule
\end{tabular}
\caption{\textbf{The editor removes the target in 81 of 100 twins and leaves the rest of the scene intact in 48.} Final codes after two independent coders audited all 31 frames of each removal window at L against the original and an author decided the 34 twins with any disagreement.}
\label{tab:app-edit-audit}
\end{table}

\begin{figure*}[!t]
\centering
\includegraphics[width=\textwidth]{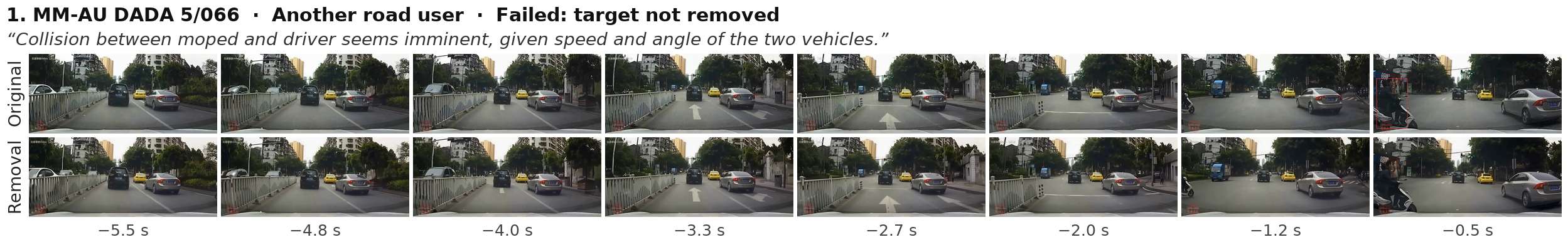}
\vspace{3pt}
\includegraphics[width=\textwidth]{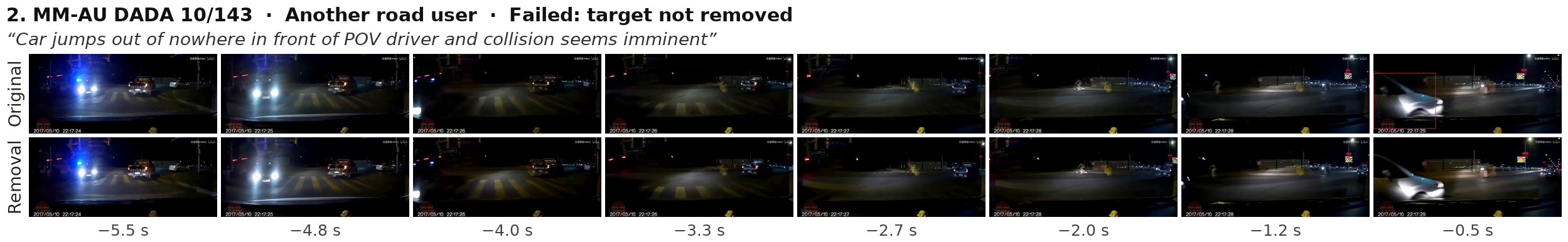}
\vspace{3pt}
\includegraphics[width=\textwidth]{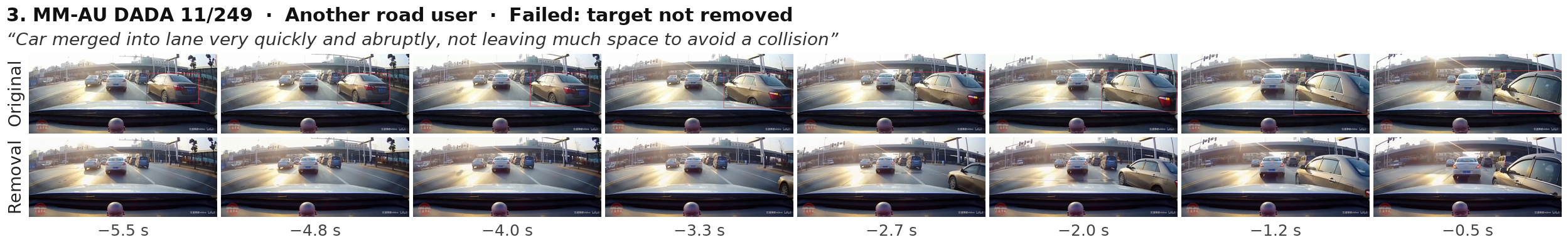}
\vspace{3pt}
\includegraphics[width=\textwidth]{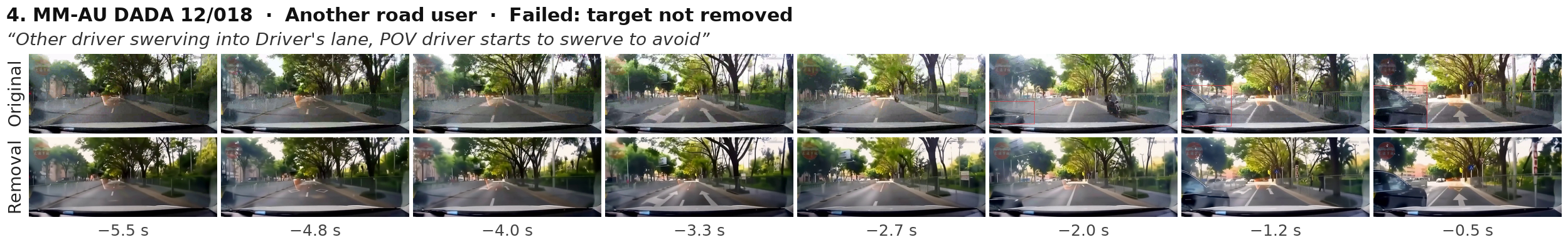}
\vspace{3pt}
\includegraphics[width=\textwidth]{figures/observer_warnings/05_mmau_dada_12_026.png}
\vspace{3pt}
\includegraphics[width=\textwidth]{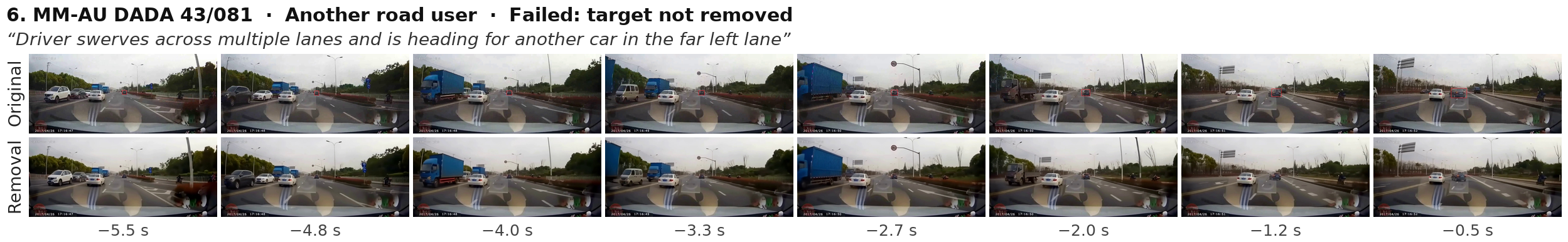}
\vspace{3pt}
\includegraphics[width=\textwidth]{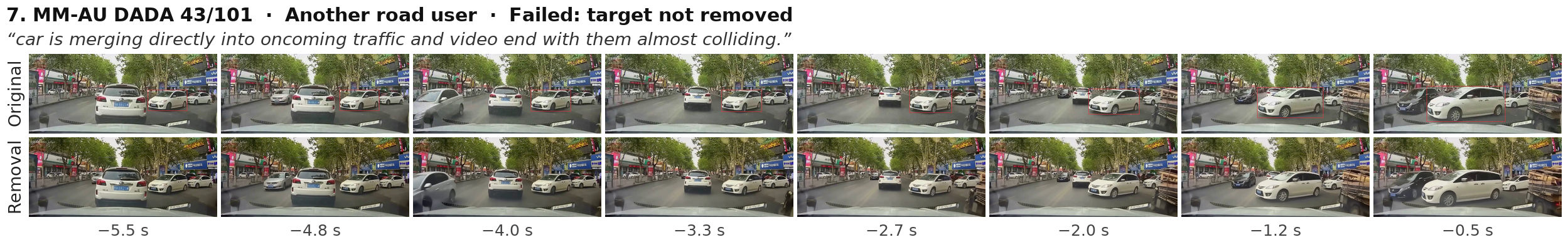}
\caption{\textbf{The 19 removal windows on which the observer warned (rows 1--7 of Table~\ref{tab:app-observer-descriptions}).} Each strip shows eight of the 31 frames of the window at L, with the time before the accident below. Top: the original footage, with the target road user boxed in red. Bottom: the removal version that the observer and the models saw.}
\label{fig:app-observer-strips-a}
\end{figure*}

\begin{figure*}[!t]
\centering
\includegraphics[width=\textwidth]{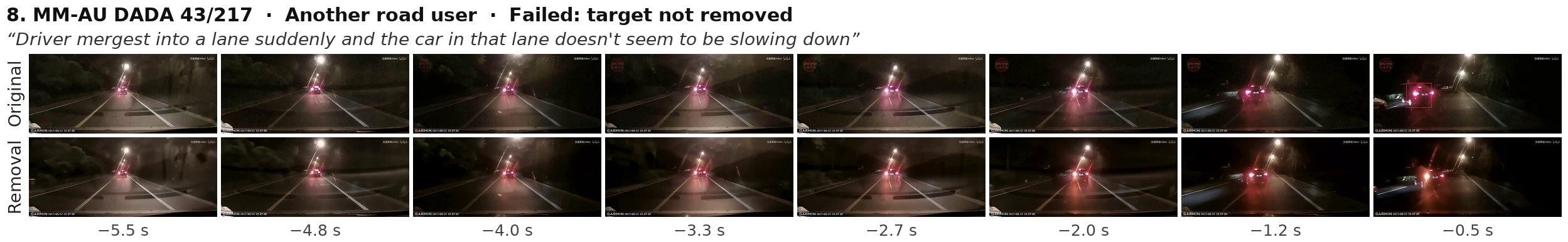}
\vspace{3pt}
\includegraphics[width=\textwidth]{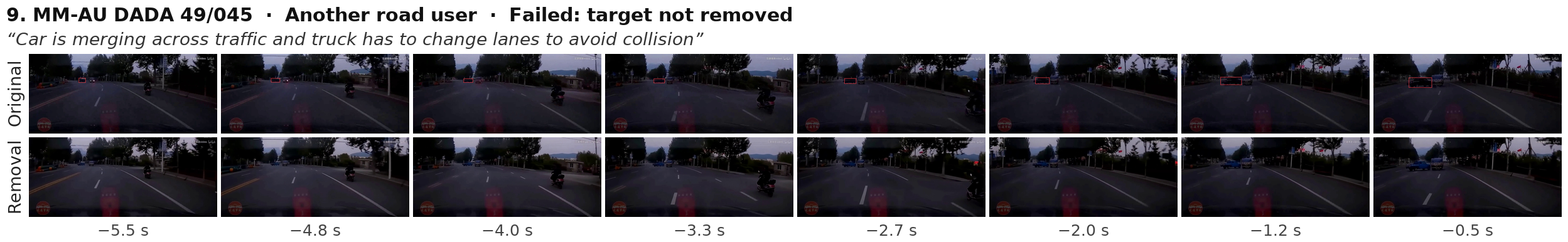}
\vspace{3pt}
\includegraphics[width=\textwidth]{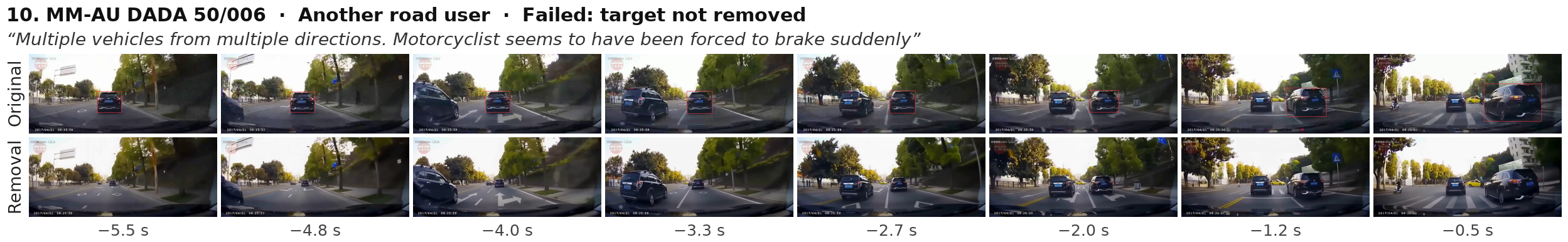}
\vspace{3pt}
\includegraphics[width=\textwidth]{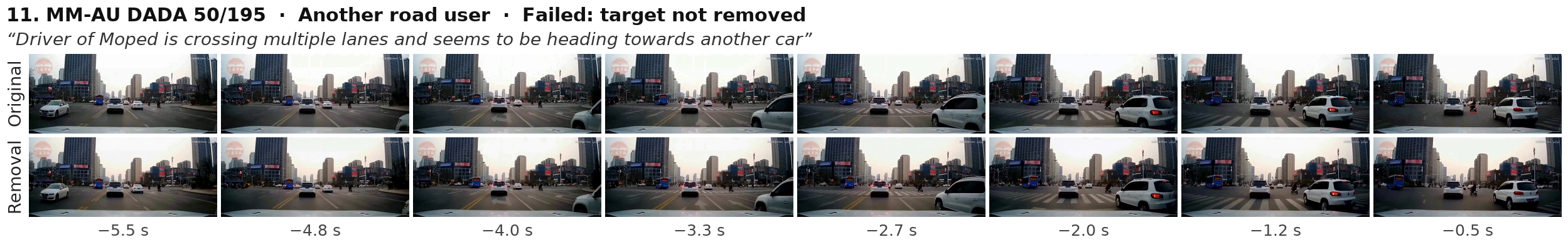}
\vspace{3pt}
\includegraphics[width=\textwidth]{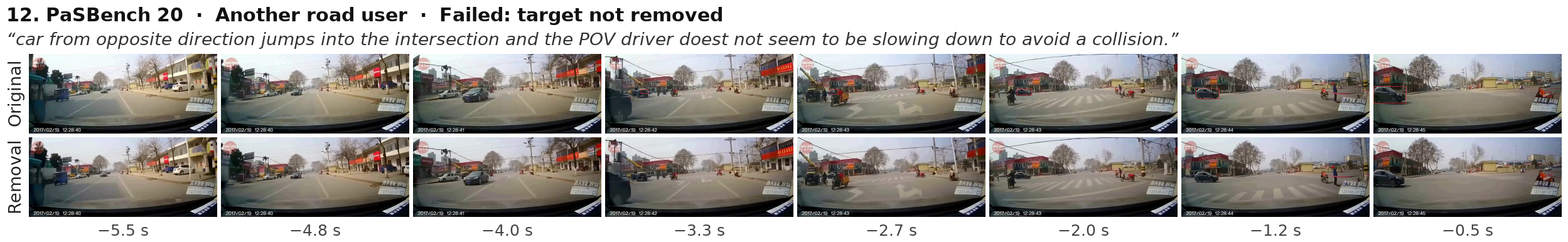}
\vspace{3pt}
\includegraphics[width=\textwidth]{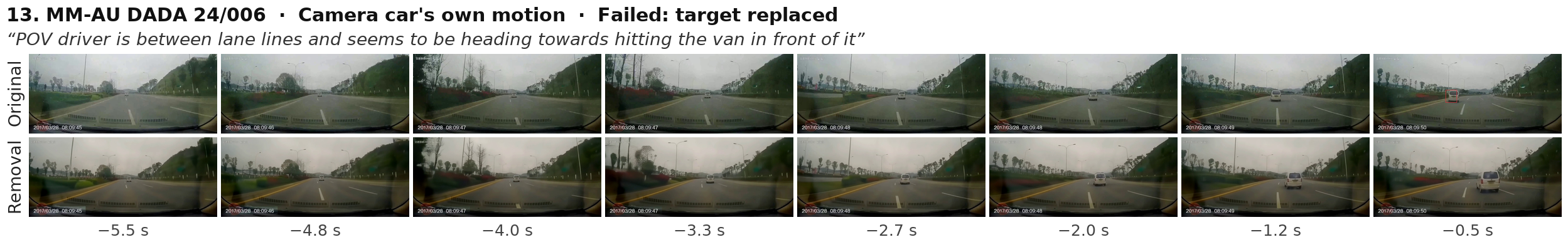}
\caption{\textbf{The 19 removal windows on which the observer warned, continued (rows 8--13 of Table~\ref{tab:app-observer-descriptions}).} Each strip shows eight of the 31 frames of the window at L, with the time before the accident below. Top: the original footage, with the target road user boxed in red. Bottom: the removal version that the observer and the models saw.}
\label{fig:app-observer-strips-b}
\end{figure*}

\begin{figure*}[!t]
\centering
\includegraphics[width=\textwidth]{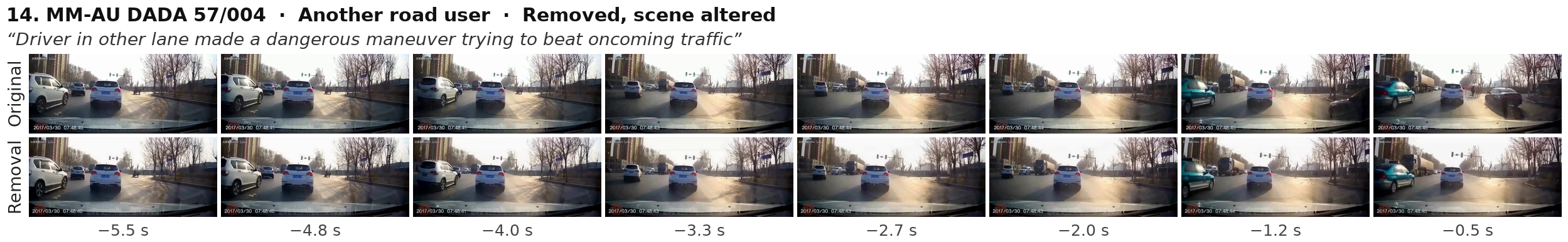}
\vspace{3pt}
\includegraphics[width=\textwidth]{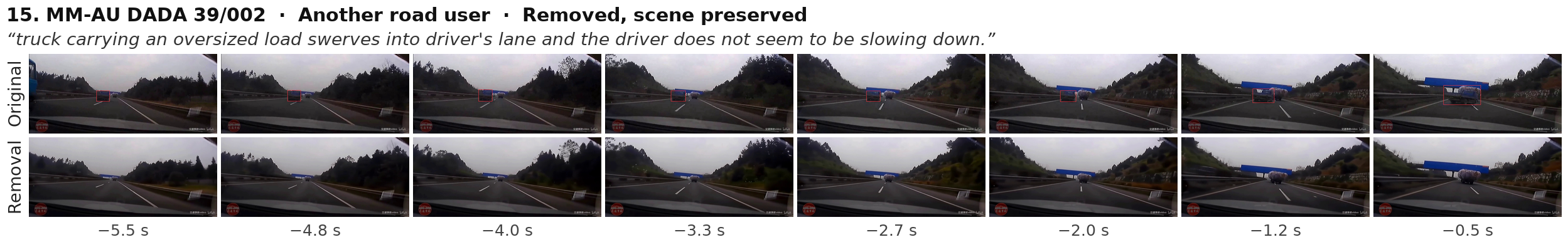}
\vspace{3pt}
\includegraphics[width=\textwidth]{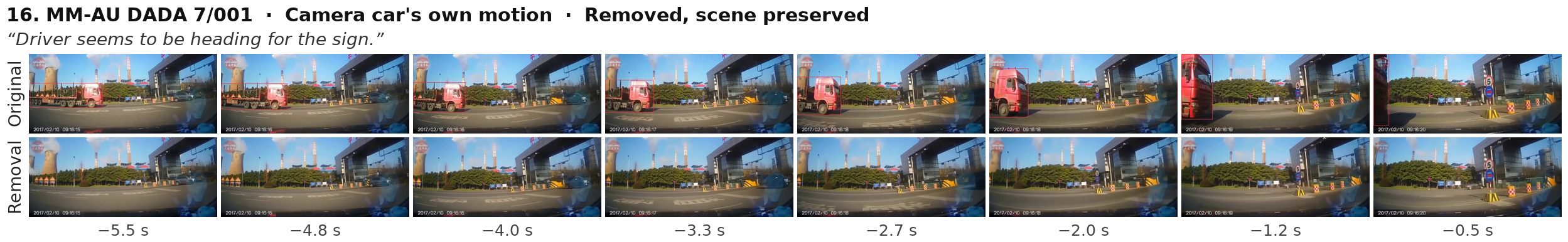}
\vspace{3pt}
\includegraphics[width=\textwidth]{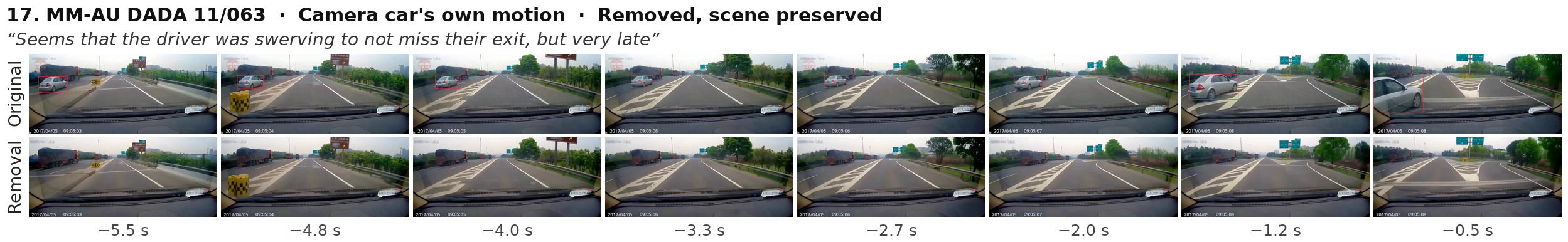}
\vspace{3pt}
\includegraphics[width=\textwidth]{figures/observer_warnings/18_mmau_dada_11_100.png}
\vspace{3pt}
\includegraphics[width=\textwidth]{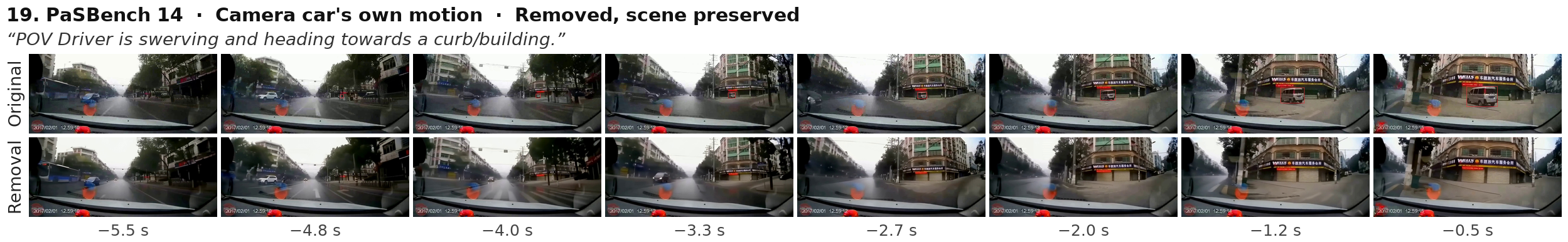}
\caption{\textbf{The 19 removal windows on which the observer warned, continued (rows 14--19 of Table~\ref{tab:app-observer-descriptions}).} Each strip shows eight of the 31 frames of the window at L, with the time before the accident below. Top: the original footage, with the target road user boxed in red. Bottom: the removal version that the observer and the models saw.}
\label{fig:app-observer-strips-c}
\end{figure*}

\paragraph{What the observer cited and why a warning can remain warranted after a clean removal.}
The observer warned on 19 of the 100 removal windows and gave a reason each time. The option ``I can't identify a specific cue'' was never used, and no description mentions an editing artefact (Table~\ref{tab:app-observer-descriptions}; Figures~\ref{fig:app-observer-strips-a}--\ref{fig:app-observer-strips-c} show the frames). We code 14 descriptions as another road user and 5 as the camera car's own motion. Set against the final codes of the edit audit, the two groups separate. Of the 14 warnings about another road user, 12 fall on failed removals, where the target is still in the footage, and two on successful ones. Of the five about the camera car's own motion, four fall on clean removals, where the target is gone and the rest of the scene is preserved; the fifth falls on a twin in which the target is still present but moves differently. On the 81 successful removals, the observer thus warns six times, four of them about the camera car's own motion.

A removal deletes the road user that caused the accident, but not what that road user did to the rest of the scene. Everything else stays as recorded, including the reactions to it: the camera car brakes or swerves, and other road users change course. The observer never saw the original footage, so these reactions appear without a cause, and an unexplained swerve is itself a reason to warn. Row~18 is an example (also shown at the bottom of Figure~\ref{fig:observer-archetypes}). In the original, the camera car closes on a car ahead on a curving ramp and swerves at the end of the window. In the twin, that car is removed cleanly and the swerve remains, which the observer describes as ``driver seems to swerve/lose control right at the end of the video''. In all four clean removals of this group, the camera car follows the same path as in the original footage, in which the removed road user was present. A warning on a clean removal is therefore not by itself a false alarm. The removal tests whether a warning depends on the target's presence; it does not turn the remaining footage into a hazard-free scene.

\begin{table*}[t]
\centering
\footnotesize
\setlength{\tabcolsep}{4pt}
\begin{tabular}{@{}rlp{0.45\textwidth}ll@{}}
\toprule
\# & Clip & Observer's description (verbatim) & Cited cue & Edit audit (final code) \\
\midrule
1 & 5/066 & Collision between moped and driver seems imminent, given speed and angle of the two vehicles. & Another road user & Failed: target not removed \\
2 & 10/143 & Car jumps out of nowhere in front of POV driver and collision seems imminent & Another road user & Failed: target not removed \\
3 & 11/249 & Car merged into lane very quickly and abruptly, not leaving much space to avoid a collision & Another road user & Failed: target not removed \\
4 & 12/018 & Other driver swerving into Driver's lane, POV driver starts to swerve to avoid & Another road user & Failed: target not removed \\
5 & 12/026 & Car cutting off much too close, collision seems imminent & Another road user & Failed: target not removed \\
6 & 43/081 & Driver swerves across multiple lanes and is heading for another car in the far left lane & Another road user & Failed: target not removed \\
7 & 43/101 & car is merging directly into oncoming traffic and video end with them almost colliding. & Another road user & Failed: target not removed \\
8 & 43/217 & Driver mergest into a lane suddenly and the car in that lane doesn't seem to be slowing down & Another road user & Failed: target not removed \\
9 & 49/045 & Car is merging across traffic and truck has to change lanes to avoid collision & Another road user & Failed: target not removed \\
10 & 50/006 & Multiple vehicles from multiple directions. Motorcyclist seems to have been forced to brake suddenly & Another road user & Failed: target not removed \\
11 & 50/195 & Driver of Moped is crossing multiple lanes and seems to be heading towards another car & Another road user & Failed: target not removed \\
12 & PaSBench 20 & car from opposite direction jumps into the intersection and the POV driver doest not seem to be slowing down to avoid a collision. & Another road user & Failed: target not removed \\
13 & 24/006 & POV driver is between lane lines and seems to be heading towards hitting the van in front of it & Camera car's own motion & Failed: target replaced \\
14 & 57/004 & Driver in other lane made a dangerous maneuver trying to beat oncoming traffic & Another road user & Removed, scene altered \\
15 & 39/002 & truck carrying an oversized load swerves into driver's lane and the driver does not seem to be slowing down. & Another road user & Removed, scene preserved \\
16 & 7/001 & Driver seems to be heading for the sign. & Camera car's own motion & Removed, scene preserved \\
17 & 11/063 & Seems that the driver was swerving to not miss their exit, but very late & Camera car's own motion & Removed, scene preserved \\
18 & 11/100 & driver seems to swerve/lose control right at the end of the video & Camera car's own motion & Removed, scene preserved \\
19 & PaSBench 14 & POV Driver is swerving and heading towards a curb/building. & Camera car's own motion & Removed, scene preserved \\
\bottomrule
\end{tabular}
\caption{\textbf{The observer's reasons for warning after removal.} All 19 warnings on the 100 removal windows at L, with the observer's description verbatim, the coded cue category, and the final edit-audit code of the twin.}
\label{tab:app-observer-descriptions}
\end{table*}

\section{Extended Model Evaluation}
\label{app:model-results}

\paragraph{Sensitivity to individual raters and leave-one-rater-out references.}
Tables~\ref{tab:app-per-rater} and~\ref{tab:app-loro} examine how the onset-discrimination gaps reported in Section~\ref{sec:res-models} vary across individual raters and two-rater subsets. Four models, Claude Opus~5.5 (onset AUC 0.60), Kimi-K3.1-Pro (0.59), GLM-5.3-Flash (0.56), and Nemotron-3.5-Super-VL (0.56), have lower onset AUC than each of the three raters individually, with all 95\% confidence intervals excluding zero (Opus~5.5 against Rater~2 is borderline at a lower bound of 0.003, across 18 model-by-rater comparisons), and remain below every two-rater leave-one-out reference.

The two highest-scoring models on this measure, GPT-6 Astra (0.68) and Gemini~4 Argon (0.67), match Rater~2 (0.68; paired differences of 0.00 and 0.01) while trailing Rater~1 (0.76) and Rater~3 (0.77) by 0.08 to 0.10 AUC. The inter-rater spread between Rater~2 and Raters~1 and~3 (0.08 to 0.09) is at least as large as Astra's and Argon's 0.06 gap to the three-rater mean (0.73). In the leave-one-out analysis (Table~\ref{tab:app-loro}), Astra and Argon remain below the reference in point estimate across all subsets (0.04 to 0.09), but their 95\% intervals include zero whenever Rater~2 is retained in a two-rater reference. Astra's equivalence within the planned $\pm0.10$ margin holds in two of the three leave-one-out subsets, but not when Rater~2 is excluded (90\% CI 0.045 to 0.130); Argon lies within $\pm0.10$ only when Rater~3 is excluded. Against the secondary reference formed by the raters' mean risk (0.82 for all three raters; 0.76 to 0.84 across the two-rater subsets), all six models remain below people in every subset, with paired differences of 0.15 to 0.27 against all three raters and 0.09 to 0.28 across the two-rater subsets (every 95\% interval excluding zero). All comparisons in Tables~\ref{tab:app-per-rater} and~\ref{tab:app-loro} are conditional on this panel of three raters.

\begin{table*}[t]
\centering
\small
\setlength{\tabcolsep}{5pt}
\begin{tabular}{@{}lccccc@{}}
\toprule
Model & Onset AUC & vs.\ Rater 1 (0.76) & vs.\ Rater 2 (0.68) & vs.\ Rater 3 (0.77) & Raters reliably above model \\
\midrule
Claude Opus 5.5       & 0.60 & 0.16 [0.09, 0.23] & 0.08 [0.00, 0.16]    & 0.17 [0.10, 0.24] & 3 of 3 \\
GPT-6 Astra           & 0.68 & 0.08 [0.02, 0.15] & 0.00 [$-0.06$, 0.06] & 0.09 [0.04, 0.15] & 2 of 3 \\
Gemini 4 Argon        & 0.67 & 0.09 [0.02, 0.16] & 0.01 [$-0.06$, 0.08] & 0.10 [0.04, 0.16] & 2 of 3 \\
Kimi-K3.1-Pro         & 0.59 & 0.17 [0.10, 0.24] & 0.09 [0.02, 0.15]    & 0.18 [0.12, 0.24] & 3 of 3 \\
GLM-5.3-Flash         & 0.56 & 0.20 [0.13, 0.27] & 0.12 [0.05, 0.19]    & 0.21 [0.15, 0.28] & 3 of 3 \\
Nemotron-3.5-Super-VL & 0.56 & 0.20 [0.15, 0.26] & 0.12 [0.05, 0.19]    & 0.21 [0.15, 0.28] & 3 of 3 \\
\bottomrule
\end{tabular}
\caption{\textbf{Pairwise comparison of each model against each individual human rater on the 150 rated hazard clips.} Entries report the rater's within-clip onset AUC (B vs.\ A1, paired by clip) minus the model's, with 95\% bootstrap confidence intervals from 5,000 clip resamples. For Claude Opus~5.5 vs.\ Rater~2, the unrounded lower bound is $+0.003$.}
\label{tab:app-per-rater}
\end{table*}

\begin{table*}[t]
\centering
\small
\resizebox{\textwidth}{!}{%
\begin{tabular}{@{}lccccccc@{}}
\toprule
Reference panel & Human AUC & Opus 5.5 & Astra & Argon & Kimi-K3.1-Pro & GLM-5.3-Flash & Nemotron-3.5 \\
\midrule
All three (main text) & 0.73 [0.70, 0.77] & 0.14 [0.08, 0.20] & 0.06 [0.01, 0.11]    & 0.06 [0.01, 0.12]    & 0.14 [0.09, 0.20] & 0.18 [0.12, 0.24] & 0.18 [0.13, 0.23] \\
Without Rater 1       & 0.72 [0.68, 0.77] & 0.13 [0.06, 0.20] & 0.05 [$-0.00$, 0.10] & 0.05 [$-0.01$, 0.12] & 0.13 [0.07, 0.19] & 0.17 [0.10, 0.23] & 0.17 [0.11, 0.23] \\
Without Rater 2       & 0.76 [0.73, 0.80] & 0.17 [0.10, 0.23] & 0.09 [0.04, 0.14]    & 0.09 [0.04, 0.15]    & 0.17 [0.12, 0.23] & 0.21 [0.14, 0.27] & 0.21 [0.16, 0.26] \\
Without Rater 3       & 0.72 [0.68, 0.75] & 0.12 [0.06, 0.19] & 0.04 [$-0.01$, 0.09] & 0.05 [$-0.01$, 0.11] & 0.13 [0.07, 0.18] & 0.16 [0.10, 0.22] & 0.16 [0.11, 0.21] \\
\bottomrule
\end{tabular}%
}
\caption{\textbf{Leave-one-rater-out sensitivity of within-clip onset AUC differences on the 150 rated hazard clips.} Each model column reports the mean individual-rater onset AUC for the indicated rater subset minus the model's onset AUC, paired by clip, with 95\% bootstrap confidence intervals from 5,000 clip resamples.}
\label{tab:app-loro}
\end{table*}

\paragraph{Detailed mean risk comparisons at each decision time.}
Figure~\ref{fig:curves} (Section~\ref{sec:res-models}) plots both warning rates and mean risk levels (coded 0 for none to 4 for critical) across A2, A1, B, B2, L, and hazard-free windows. For people, we average the three raters' scores within each clip before averaging across clips; on hazard-free clips, we average each clip over its rated windows first. Comparisons are paired by clip, with 95\% CIs from 5,000 clip resamples. Within a clip, people's mean risk rises by 0.86 levels from A1 to B (95\% CI 0.70 to 1.02). Astra and Argon both rise by 0.78 levels over the same interval (differences from people: 0.08, 95\% CIs $-0.12$ to 0.27 and $-0.15$ to 0.30), whereas the other four models rise by only 0.17 to 0.33 levels (each reliably below people by 0.53 to 0.68 levels). At A1, Opus~5.5 (1.79) and GLM-5.3-Flash (1.58) exceed people (0.65) by 1.15 (95\% CI 0.98 to 1.30) and 0.93 (0.76 to 1.10) levels, and both also rate hazard-free clips higher than people (0.80 and 0.60 vs.\ 0.32; differences 0.48 [0.12 to 0.79] and 0.28 [0.08 to 0.48]). At L, Opus~5.5 matches people (2.67 vs.\ 2.64; difference 0.03 [$-0.14$ to 0.21]), whereas GLM-5.3-Flash (2.01; $-0.63$ [$-0.82$ to $-0.45$]), Astra (2.13; $-0.51$ [$-0.74$ to $-0.28$]), Kimi-K3.1-Pro (1.70; $-0.94$ [$-1.13$ to $-0.75$]), and Nemotron-3.5-Super-VL (0.65; $-1.99$ [$-2.20$ to $-1.79$]) sit reliably below people. Argon is the sole exception whose lower warning rates at B2 and L do not correspond to reliably lower mean risk (differences $-0.14$ [$-0.45$ to 0.18] and $-0.19$ [$-0.40$ to 0.03]).

\subsection{Temporal Warning Measures}
\label{app:model-timing}

\begin{figure*}[tp]
\centering
\begin{minipage}[t]{0.5\textwidth}
\centering
\includegraphics[width=\linewidth]{figures/clip_timelines/01_mmau_dada_11_180.png}
\end{minipage}\hfill
\begin{minipage}[t]{0.5\textwidth}
\centering
\includegraphics[width=\linewidth]{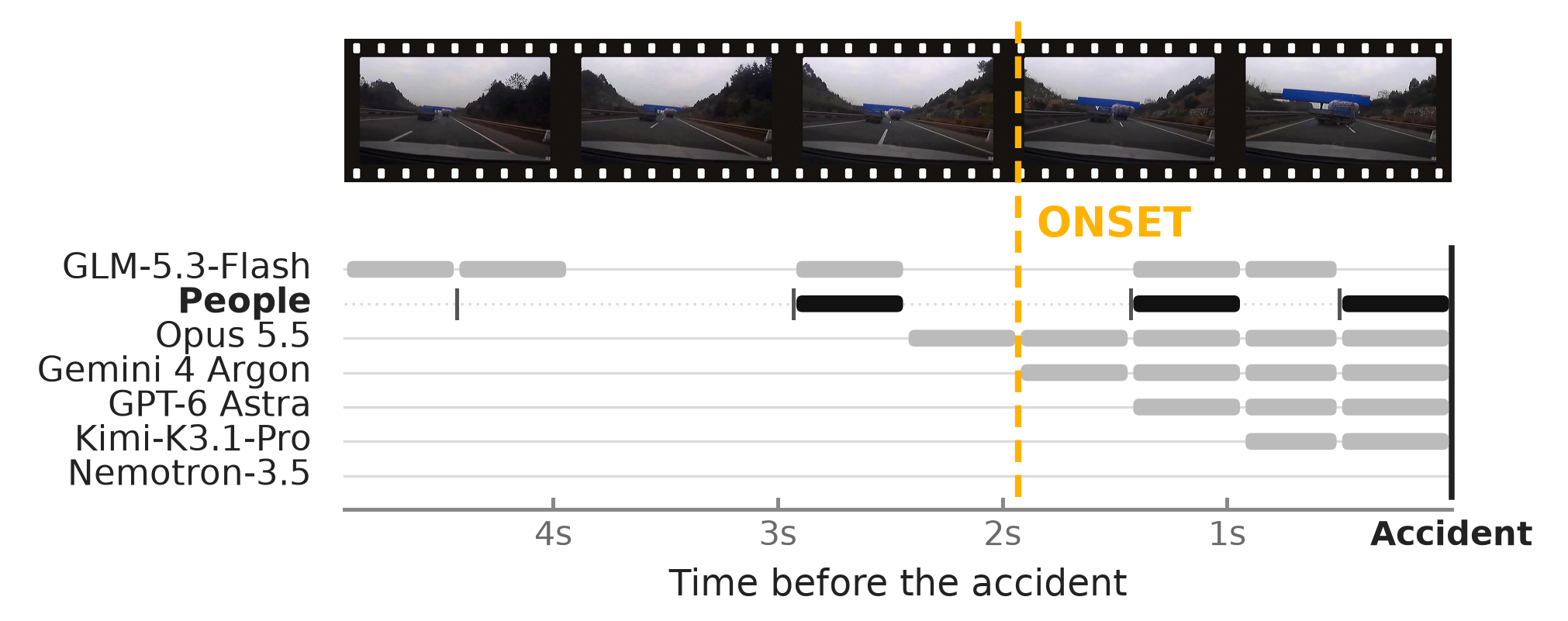}
\end{minipage}

\vspace{5pt}
\begin{minipage}[t]{0.5\textwidth}
\centering
\includegraphics[width=\linewidth]{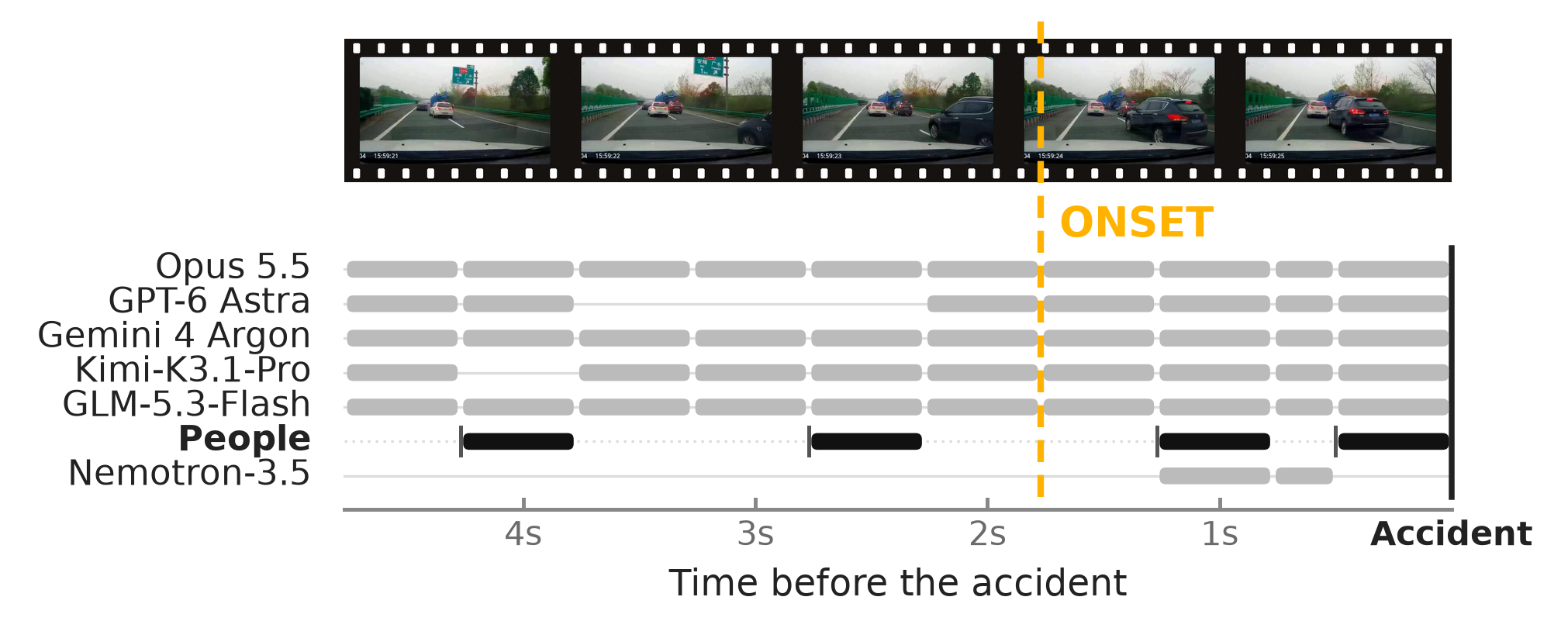}
\end{minipage}\hfill
\begin{minipage}[t]{0.5\textwidth}
\centering
\includegraphics[width=\linewidth]{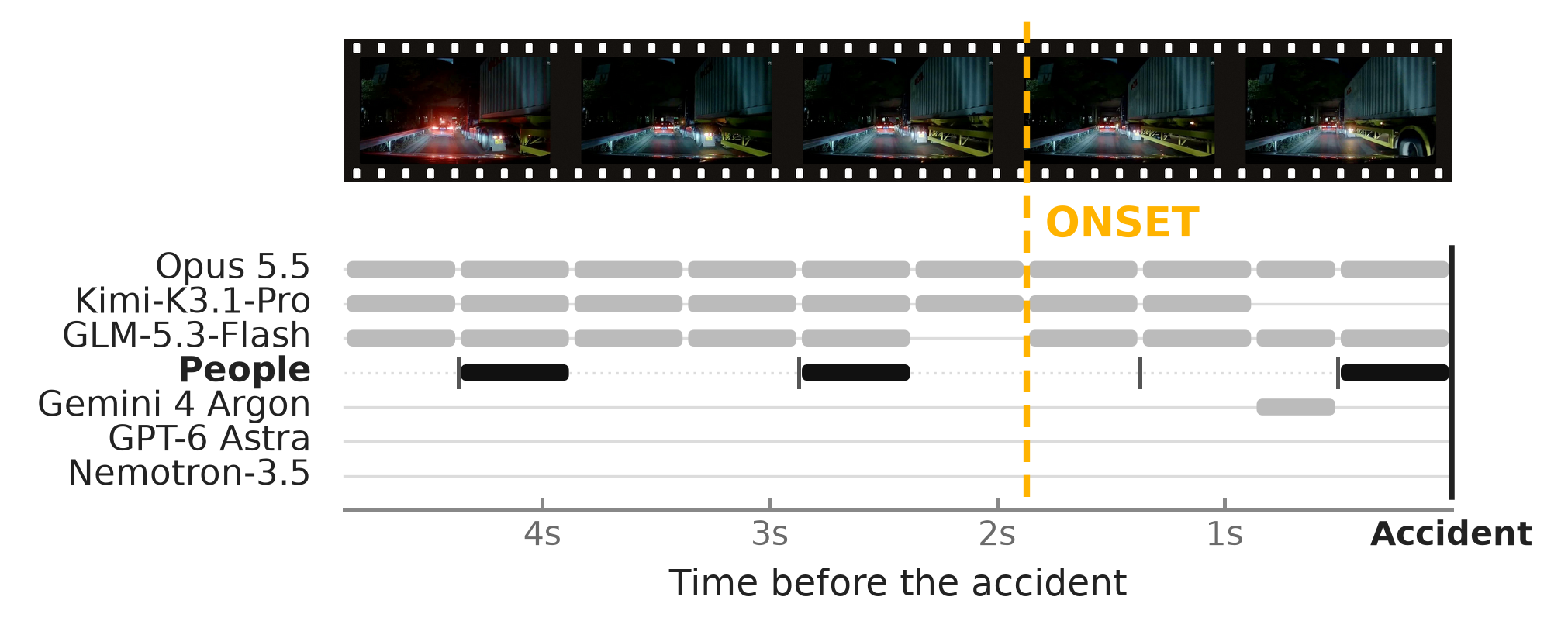}
\end{minipage}

\vspace{5pt}
\begin{minipage}[t]{0.5\textwidth}
\centering
\includegraphics[width=\linewidth]{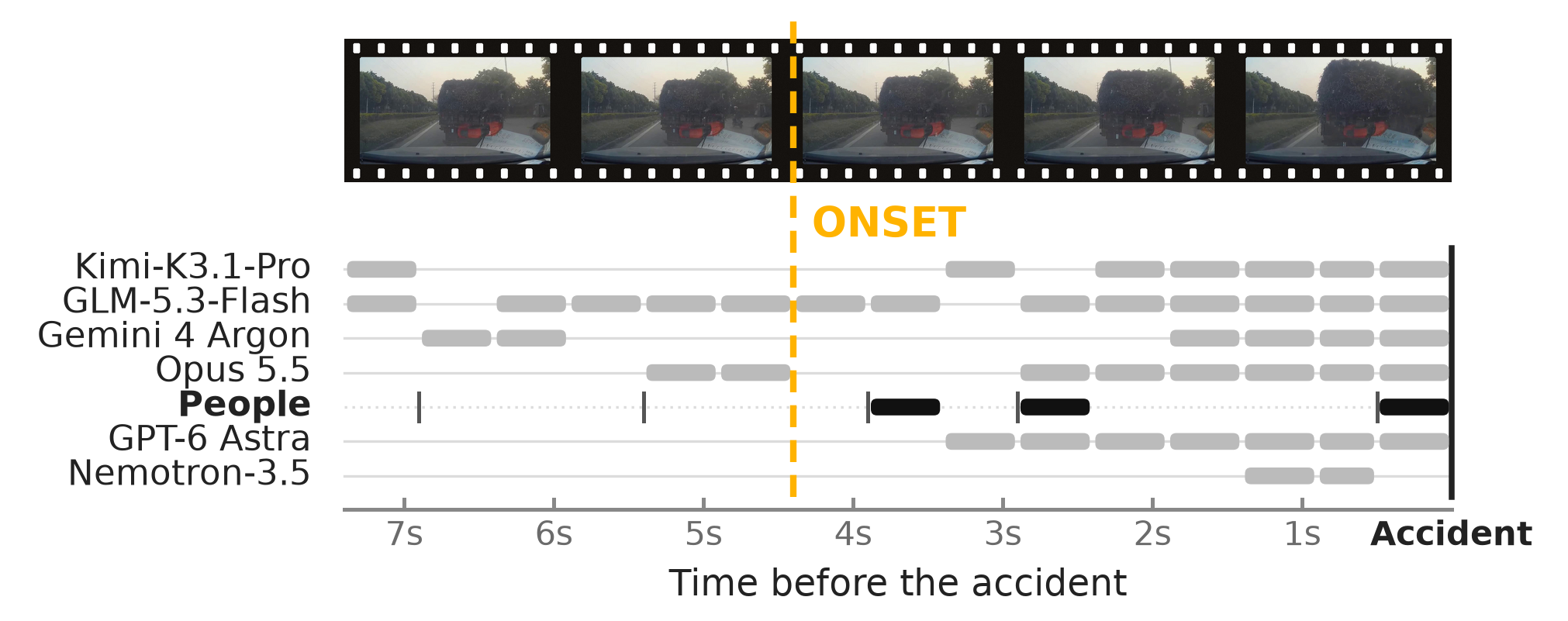}
\end{minipage}\hfill
\begin{minipage}[t]{0.5\textwidth}
\centering
\includegraphics[width=\linewidth]{figures/clip_timelines/21_mmau_dada_11_111.png}
\end{minipage}

\vspace{5pt}
\begin{minipage}[t]{0.5\textwidth}
\centering
\includegraphics[width=\linewidth]{figures/clip_timelines/15_mmau_dada_39_010.png}
\end{minipage}\hfill
\begin{minipage}[t]{0.5\textwidth}
\centering
\includegraphics[width=\linewidth]{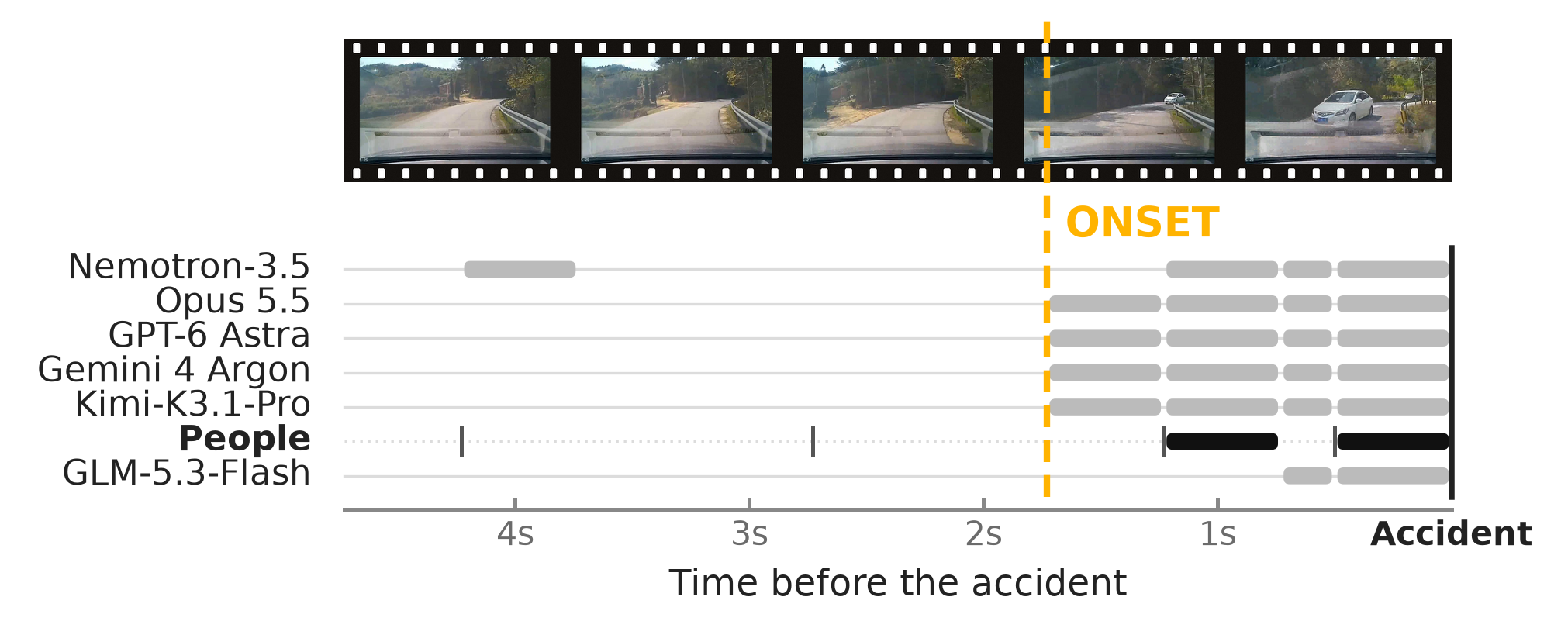}
\end{minipage}

\vspace{5pt}
\begin{minipage}[t]{0.5\textwidth}
\centering
\includegraphics[width=\linewidth]{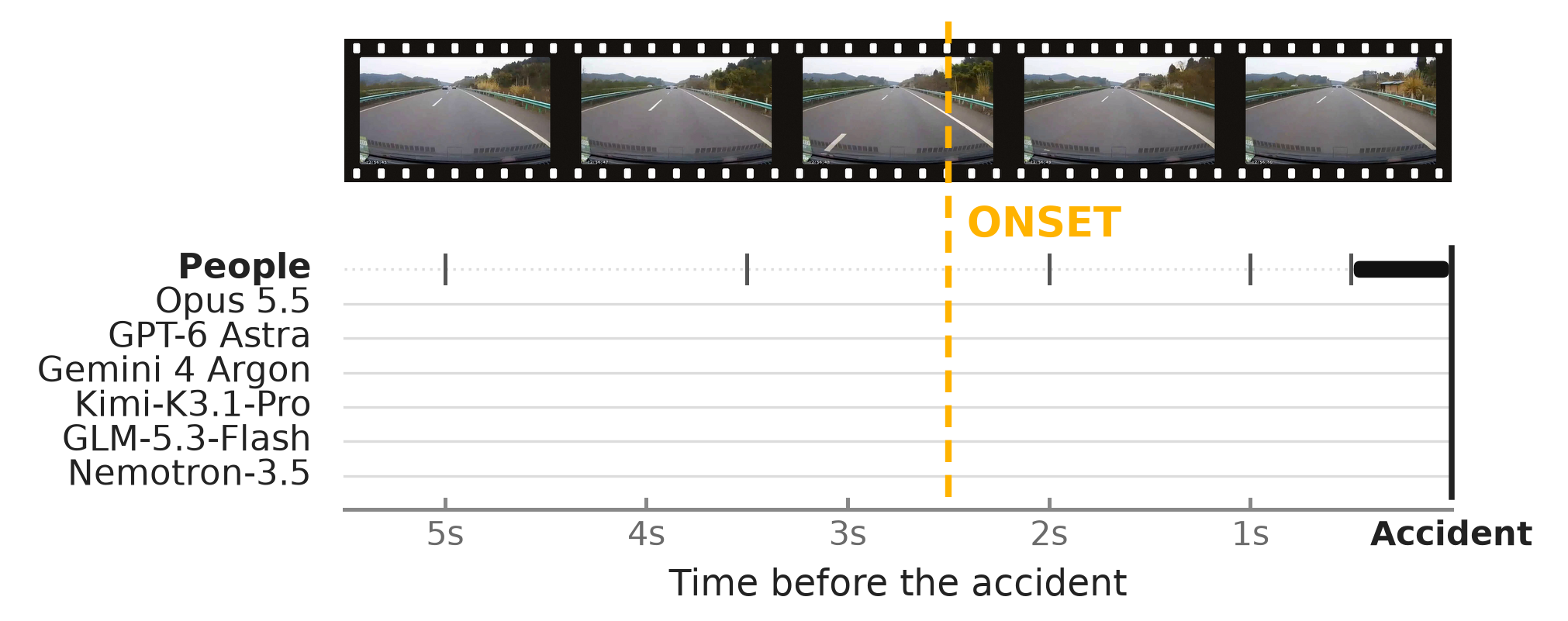}
\end{minipage}\hfill
\begin{minipage}[t]{0.5\textwidth}
\centering
\includegraphics[width=\linewidth]{figures/clip_timelines/27_pasvbench_000375.png}
\end{minipage}
\caption{\textbf{When people and models warn, clip by clip.} Each panel shows one hazard clip: five frames evenly spaced over the interval shown and, below them, one row per responder on an axis of time before the accident. Each grey block marks a model's warning at a half-second decision point, with rows ordered from earliest to latest first warning. People were queried at the moments marked by ticks on the dotted row (four or five per clip), and a black block marks a warning, meaning that at least two of the three raters warn. No block means no warning. The dashed yellow line marks the annotated hazard onset, and the solid black line on the right marks the accident. In five of the clips, people first warn before the annotated onset, in three of the clips they warn at the first moment after the onset, and in two of the clips, later.}
\label{fig:clip-timelines}
\end{figure*}

Table~\ref{tab:app-model-timing} gives the timing summaries defined in Section~\ref{sec:proto-obs} across all 666 hazard clips (6,778 windows, including L and P), together with the number of clips behind each value. Every model's largest single-step increase falls between 0.5\,s before and 1.5\,s after the annotated onset. Claude Opus~5.5 and GLM-5.3-Flash warn earliest, with median first warnings 3.4 and 3.5\,s before the accident, while Nemotron-3.5-Super-VL gives no warning before the accident on 52\% of hazard clips. Figure~\ref{fig:clip-timelines} shows the models' warnings at every 0.5\,s decision point, and people's at the four or five moments they rated, on ten hazard clips (extending the four examples in Figure~\ref{fig:clip-timelines-main}).

\begin{table*}[t]
\centering
\footnotesize
\begin{tabular}{@{}lcccc@{}}
\toprule
Model & Peak warning rate (\%) & Largest increase (pp) & Median first warning (s before accident) & No warning before the accident (\%) \\
\midrule
Claude Opus 5.5       & 84.6 (+1.5\,s; 292) & 11.4 ($-0.5\to0$\,s; 666) & 3.4 (645) & 3.2 \\
GPT-6 Astra           & 66.4 (+1.5\,s; 292) & 15.3 ($0\to{+0.5}$\,s; 666) & 2.1 (565) & 15.2 \\
Gemini 4 Argon        & 60.3 (+1.5\,s; 292) & 14.6 ($0\to{+0.5}$\,s; 666) & 2.0 (565) & 15.2 \\
Kimi-K3.1-Pro         & 48.8 (+1.5\,s; 291) & 9.4 ($0\to{+0.5}$\,s; 662) & 2.3 (528) & 20.7 \\
GLM-5.3-Flash         & 61.5 (+1.0\,s; 473) & 8.3 ($+0.5\to{+1.0}$\,s; 471) & 3.5 (617) & 7.4 \\
Nemotron-3.5-Super-VL & 18.8 (+1.5\,s; 292) & 5.5 ($+1.0\to{+1.5}$\,s; 292) & 2.3 (318) & 52.3 \\
\bottomrule
\end{tabular}
\caption{\textbf{Every model's largest increase in warnings comes within 1.5\,s of the annotated onset, but how early and how often the models warn differs widely.} Timing summaries on the 666 hazard clips. Peak warning rate: the highest warning rate over the onset-aligned 0.5\,s grid, among decision times with at least 100 clips (time after the onset; clips evaluated). Largest increase: the largest rise in warning rate between consecutive decision times, on the clips evaluated at both, among steps with at least 100 such clips (step relative to the onset; clips). Because the grid stops 0.5\,s before the accident, later grid times include only clips with a longer interval between onset and accident (all 666 clips through $+0.5$\,s, 474 at $+1.0$\,s, and 292 at $+1.5$\,s, minus any excluded invalid replies). Median first warning: the median time before the accident of each model's first warning, on the clips with any warning before the accident (clips in parentheses). No warning before the accident: the share of the 666 clips on which the model never warns before the accident.}
\label{tab:app-model-timing}
\end{table*}

\section{Human Reference and Annotation Study}
\label{app:human-reference}

\paragraph{Sample and annotation protocol.}
We collect the human reference on 820 observation windows drawn from 180 real clips: 120 MM-AU DADA accident clips, 30 PaSBench-Video hazard clips, and 30 PaSBench-Video hazard-free clips. On the 150 hazard clips, raters evaluate the named decision points A2 ($n=149$), A1 ($n=150$), B ($n=150$), B2 (on the $n=75$ clips where B2 precedes L), and L ($n=150$), yielding 674 hazard windows. On the 30 hazard-free clips, raters evaluate up to five matched decision times per clip (146 hazard-free windows), for 820 windows in total. Three raters completed all 820 windows independently. On binary warning decisions across all 820 windows, any two raters agree 81.9\% of the time (the mean pairwise exact agreement reported in Table~\ref{tab:human-reference}), and all three agree on 72.8\% of windows (Fleiss' $\kappa=0.63$). On the five-level ordinal risk ratings (0 to 4), inter-rater agreement is Krippendorff's $\alpha=0.75$.

\paragraph{Variation across individual raters and consensus thresholds.}
Across the 150 rated hazard clips, the three raters achieve individual within-clip onset AUCs of 0.76 (Rater~1), 0.68 (Rater~2), and 0.77 (Rater~3), with a mean of 0.73 (95\% CI 0.70 to 0.77; model comparisons against each individual rater and leave-one-out reference appear in Appendix~\ref{app:model-results}). Table~\ref{tab:app-rater-thresholds} shows how pre-onset warnings (at A2 or A1), clips with no warning through L, and early warnings on the 30 hazard-free clips change as the consensus threshold increases from at least one rater to all three (relaxing the requirement that raters warn at the same pre-onset window changes the MM-AU DADA pre-onset shares by at most three percentage points). Relative to the hazard-free baseline in Table~\ref{tab:app-rater-thresholds}, pre-onset warnings on MM-AU DADA occur 32 percentage points more often when requiring at least one rater (95\% CI 14 to 48) and 19 points more often for a majority (95\% CI 5 to 32), whereas at the unanimous threshold the 8-point difference (95\% CI $-4$ to 18) is not distinguishable from zero (within the 30 PaSBench hazard clips, where pre-onset rates are 37\%, 27\%, and 20\%, the 13- to 17-point differences from the 30 PaSBench hazard-free clips all have 95\% confidence intervals that include zero).

\begin{table*}[t]
\centering
\small
\setlength{\tabcolsep}{5pt}
\begin{tabular}{@{}lccccc@{}}
\toprule
& \multicolumn{2}{c}{MM-AU DADA (120 clips)} & \multicolumn{2}{c}{All hazard clips (150)} & Hazard-free clips (30) \\
\cmidrule(lr){2-3}\cmidrule(lr){4-5}\cmidrule(l){6-6}
Raters warning at same moment & Before onset & Never through L & Before onset & Never through L & Two earliest windows \\
\midrule
At least 1 of 3            & 52\% (62) [42, 60] & 3\% (4) [1, 7]    & 49\% (73) [41, 57] & 4\% (6) [1, 7]    & 20\% (6) [7, 37] \\
At least 2 of 3 (majority) & 29\% (35) [21, 38] & 9\% (11) [4, 15]  & 29\% (43) [21, 36] & 10\% (15) [5, 15] & 10\% (3) [0, 20] \\
All 3 of 3 (unanimous)     & 14\% (17) [8, 21]  & 29\% (35) [21, 38] & 15\% (23) [10, 21] & 28\% (42) [21, 35] & 7\% (2) [0, 17] \\
\bottomrule
\end{tabular}
\caption{\textbf{Pre-onset, absent, and hazard-free warnings across human consensus thresholds.} Entries report the percentage of clips, clip count in parentheses, and 95\% bootstrap confidence interval in brackets (in percentages, from 5,000 clip resamples) on which at least $k \in \{1, 2, 3\}$ raters warn at the same pre-onset decision time (A2 or A1), at no evaluated decision time through L, or at one of the two earliest windows on the 30 rated hazard-free clips.}
\label{tab:app-rater-thresholds}
\end{table*}

\subsection{Human Rating Interface}
\label{app:rating-page}

\begin{figure*}[tp]
\centering
\begin{minipage}[t]{0.49\textwidth}
\centering
\includegraphics[width=\linewidth]{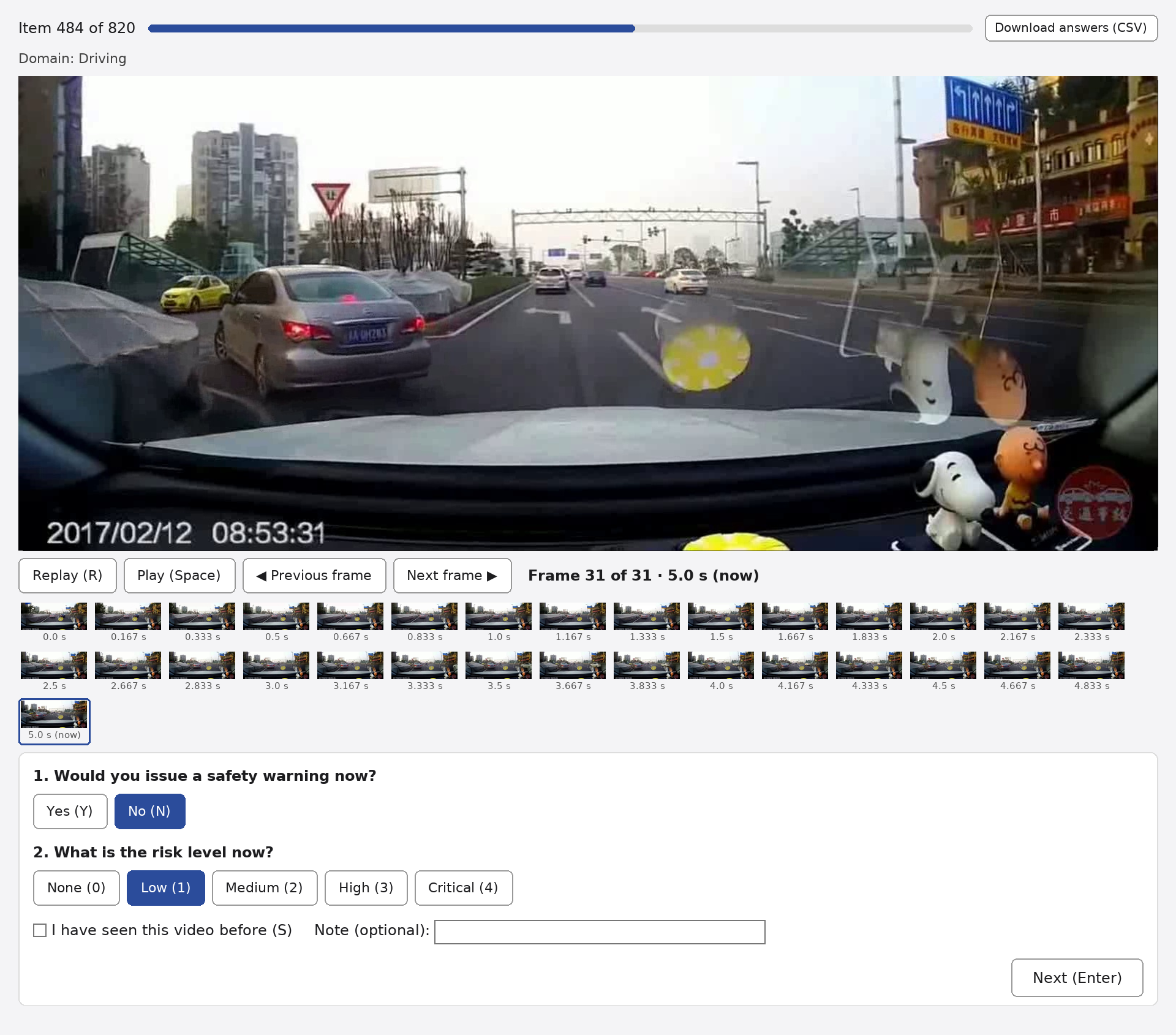}
\end{minipage}\hfill
\begin{minipage}[t]{0.49\textwidth}
\centering
\includegraphics[width=\linewidth]{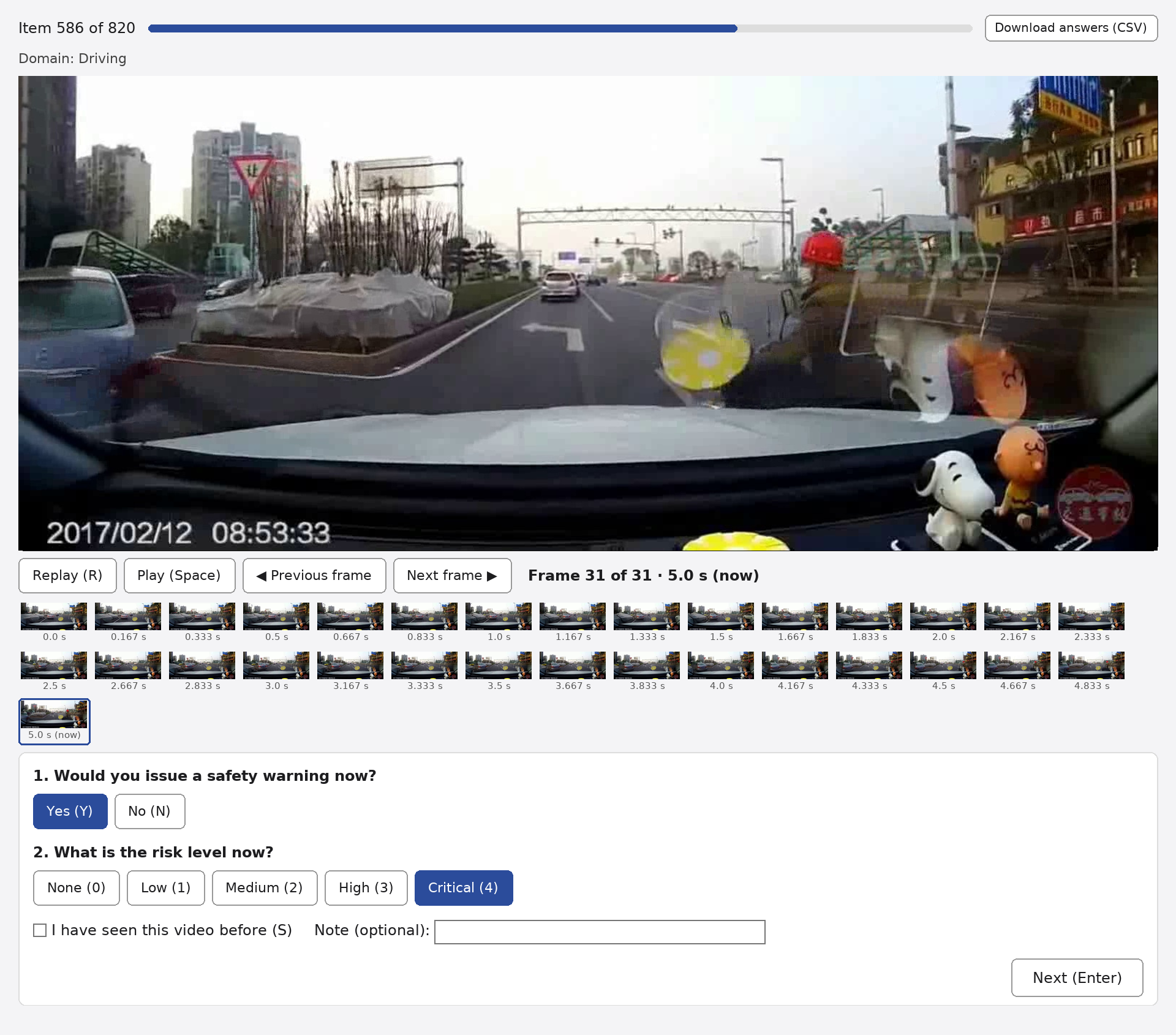}
\end{minipage}
\caption{\textbf{The rating page at two moments of one clip.} Both screens show an MM-AU DADA clip (accident category~5, video~9; MM-AU's description: ``a motorbike crosses the road''). Left: 1\,s before the annotated onset (A1), answered No and Low. Right: 0.5\,s before the accident (L), answered Yes and Critical; the motorcyclist is directly in front of the camera car.}
\label{fig:rating-screens}
\end{figure*}

Raters completed the evaluation in an offline browser interface (Figure~\ref{fig:rating-screens}) that displayed the exact 31-frame windows (5.0\,s at 6\,fps) and domain label (Driving or Daily Life) given to the models, while blinding clip identities, source datasets, and decision points (A2, A1, B, B2, or L). After three unscored practice items, each rater evaluated the 820 windows in an independent random order under two constraints: windows from the same clip always appeared in chronological order, separated by at least six other items (median 130), and submitted answers locked immediately so that later windows could not inform earlier judgments. Each window played once at real-time speed before raters could pause, replay, or step through frames. Following the decision rules in the models' prompt, raters recorded whether to issue a safety warning at the final frame and rated current risk on the five-level scale (\emph{none} to \emph{critical}). No clip was flagged as previously seen.

\section{Monitoring Protocol and Temporal Sampling}
\label{app:protocol}

\subsection{Monitoring Prompt and Model Queries}
\label{app:prompt}

\paragraph{Prompt and queries.}
Each query is a single stateless request containing the system prompt, the 31 chronological JPEG frames, and the user prompt in Figure~\ref{fig:prompt}, which adapts the neutral monitor prompt of PaSBench-Video~\cite{zhao2026pasbench} to independent 5.0\,s windows (omitting prior state and using window-relative timestamps). Only the \texttt{Domain} line varies by source (``Driving'' versus ``Daily Life''). Frames are sent at native resolution, except for five PaSBench clips scaled by 0.80--0.97 to stay below 15\,MiB. Human raters see the same frames and domain label (Appendix~\ref{app:rating-page}), and the counterfactual twins (Section~\ref{sec:proto-twins}) use the same prompt and settings.

\paragraph{Outputs.}
From the requested JSON response, we score the two fields matching the human judgments: \texttt{warn\_now} (\texttt{yes} or \texttt{no}) and \texttt{risk\_level} (\texttt{none}, \texttt{low}, \texttt{medium}, \texttt{high}, or \texttt{critical}; scored 0--4). Replies that lack a parsable JSON object or valid values for both fields are marked invalid.

\paragraph{Model settings and invalid answers.}
Table~\ref{tab:app-settings} lists the model identifiers, pinned API routes, and decoding settings. We set temperature to 0 where supported and leave reasoning settings at each provider's default (adaptive thinking for Claude Opus~5.5). Failed requests were retried; remaining invalid responses are excluded (Table~\ref{tab:app-settings}), reducing per-moment sample sizes in Table~\ref{tab:all-clips} by at most five, while all counterfactual twin windows returned valid responses. A fixed random 10\% subset (801 windows, shared across models) was queried a second time solely to compute repeat agreement ($\kappa$) in Table~\ref{tab:all-clips}.

\begin{table*}[t]
\centering
\small
\resizebox{\textwidth}{!}{%
\begin{tabular}{@{}lllcccc@{}}
\toprule
Model & Model identifier & Route and host & Temperature & Output limit (tokens) & Valid answers (of 8,011) & Valid repeat pairs (of 801) \\
\midrule
Claude Opus 5.5       & \texttt{claude-opus-5-5}              & Anthropic API                       & default & 16,384  & 8,011 & 801 \\
GPT-6 Astra           & \texttt{gpt-6-astra}                  & OpenAI API                          & default & 16,384 & 8,011 & 801 \\
Gemini 4 Argon        & \texttt{gemini-4-argon}               & Google AI Studio                    & 0       & 16,384 & 8,009 & 800 \\
Kimi-K3.1-Pro         & \texttt{moonshotai/kimi-k3.1-pro}     & OpenRouter, Moonshot AI (MXFP4)     & 0       & 16,384 & 7,978 & 795 \\
GLM-5.3-Flash         & \texttt{z-ai/glm-5.3-flash}           & OpenRouter, Z.AI (FP8)              & 0       & 16,384 & 7,987 & 795 \\
Nemotron-3.5-Super-VL & \texttt{nvidia/nemotron-3.5-super-vl} & OpenRouter, PrimeIntellect (NVFP4)  & 0       & 16,384 & 8,011 & 801 \\
\bottomrule
\end{tabular}%
}
\caption{\textbf{Model identifiers, routes, and settings.} Valid answers are windows with a valid parsed response. Valid repeat pairs are repeated windows on which both calls returned a valid response.}
\label{tab:app-settings}
\end{table*}

\begin{figure*}[t]
\centering
\begin{minipage}{0.96\linewidth}
\small\textbf{System prompt}
\begin{promptbox}
\begin{Verbatim}[fontsize=\scriptsize,breaklines=true,breakanywhere=true]
You are evaluating a real-time visual safety warning system. You only see frames up to the current timestamp. Do not infer future events from filenames, dataset knowledge, or hindsight. Warn only when the visible evidence already contains a concrete, actionable risk. Generic context such as normal traffic, a normal sidewalk, or ordinary people nearby is not enough. Return only valid compact JSON.
\end{Verbatim}
\end{promptbox}
\small\textbf{User message} (after the 31 frames, attached as separate images, oldest first)
\begin{promptbox}
\begin{Verbatim}[fontsize=\scriptsize,breaklines=true,breakanywhere=true]
Task: decide whether to issue a safety warning now.
Domain: Driving
Current timestamp: 5.0s
Current visual window: 0.0s to 5.0s
Frame timestamps, oldest to newest: 0.0s, 0.167s, 0.333s, 0.5s, 0.667s, 0.833s, 1.0s, 1.167s, 1.333s, 1.5s, 1.667s, 1.833s, 2.0s, 2.167s, 2.333s, 2.5s, 2.667s, 2.833s, 3.0s, 3.167s, 3.333s, 3.5s, 3.667s, 3.833s, 4.0s, 4.167s, 4.333s, 4.5s, 4.667s, 4.833s, 5.0s
Prior state information before this window:
none
Rules:
 - Base the decision only on the supplied frames and prior state information.
 - If a contact sheet is supplied, read it left-to-right and top-to-bottom as oldest-to-newest frames.
 - warn_now=yes only if a concrete visible cue shows an actionable risk now.
 - warn_now=no if the situation is merely abnormal, generic, ambiguous, or not yet actionable.
 - warning_text must be concise and user-facing, at most 12 words.
 - state_summary is the complete compact state at current_time_sec, about persistent scene context, objects, and motion trends already observed. No future predictions.
Return this JSON schema exactly:
 {
    "warn_now": "yes|no",
    "risk_level": "none|low|medium|high|critical",
    "reason": "",
    "warning_text": "",
    "recommended_action": "",
    "confidence": 0.0,
    "state_summary": []
 }
\end{Verbatim}
\end{promptbox}
\end{minipage}
\caption{\textbf{The monitoring prompt, verbatim.} Each query sends the system prompt, then a user message with the 31 frames as separate images (oldest first) and the text shown. Only the \texttt{Domain} line varies. The contact-sheet rule is retained from PaSBench-Video's prompt~\cite{zhao2026pasbench} and does not apply, because frames are sent individually.}
\label{fig:prompt}
\end{figure*}

\subsection{Frame Sampling and Observation Times}
\label{app:frame-sampling}

\paragraph{Observation window.}
A decision at time $t_{\text{now}}$ uses the preceding 5.0\,s as 31 frames spaced $1/6$\,s apart, at $t_k = t_{\text{now}} - (30 - k)/6$\,s for $k = 0, \dots, 30$, taking the nearest frame of the source video.

\paragraph{Why 31 frames spaced 1/6\,s apart.}
The hazards we study unfold quickly: in the 490 evaluated MM-AU DADA clips, the median interval from annotated onset to accident is 1.7\,s, and 92\% of accidents occur within 3.0\,s of onset. Video-safety benchmarks usually sample at 1--2\,fps~\cite{qiu2026sports,jindal2025ai}, which would leave about two to three frames in that interval. Adding frames requires denser sampling, because a longer look-back would exclude most clips: at 0.5\,s spacing, 32 frames need 15.5\,s of preceding footage, which only 6 of the 432 MM-AU DADA clips and 75 of the 169 PaSBench hazard clips with an A1 window provide. A $1/6$\,s interval falls on every fifth frame of 30\,fps footage and every fourth frame of the video editor's 24\,fps output, and it divides the 0.5\,s step between decision points, so adjacent windows share 28 of their 31 frames. It also exceeds the largest timing drift we observed in pilot edits (133\,ms). Thirty-one frames over exactly 5.0\,s matches PaSBench-Video's maximum of 31 frames per window~\cite{zhao2026pasbench}.

\paragraph{Decision times.}
Let $r$ denote the annotated hazard onset and $a$ the accident time. On each hazard clip, decision points lie every 0.5\,s ($t_{\text{now}} = r + 0.5j$\,s for integer $j$), starting up to 3.0\,s before the onset once 5.0\,s of footage has elapsed ($\max(5.0\,\text{s},\, r - 3.0\,\text{s})$) and continuing up to 0.5\,s before the accident or 5.0\,s after the onset ($\min(a - 0.5\,\text{s},\, r + 5.0\,\text{s})$). We add L (0.5\,s before the accident) when it does not fall on this grid, and P (1.5\,s after the accident) where trailing footage permits, yielding 6,778 windows on the 666 hazard clips. The six named decision points are:
\begin{itemize}
    \item \textbf{A2} (early lead-up; 2.5\,s before the onset, $r - 2.5$\,s; available on 415 clips),
    \item \textbf{A1} (pre-onset baseline; 1.0\,s before the onset, $r - 1.0$\,s; 601 clips),
    \item \textbf{B} (immediate post-onset; 0.5\,s after the onset, $r + 0.5$\,s; all 666 clips),
    \item \textbf{B2} (developing hazard; 1.5\,s after the onset, $r + 1.5$\,s; the 292 clips where the accident occurs at least 2.0\,s after the onset, $a - r \ge 2.0$\,s),
    \item \textbf{L} (final pre-accident moment; 0.5\,s before the accident, $a - 0.5$\,s; all 666 clips), and
    \item \textbf{P} (post-accident check; 1.5\,s after the accident, $a + 1.5$\,s; 618 clips).
\end{itemize}
Because the accident always occurs at least 1.0\,s after the onset ($a - r \ge 1.0$\,s; Appendix~\ref{app:datasets}), B never follows L; on the 26 MM-AU DADA clips where the accident occurs exactly 1.0\,s after the onset ($a - r = 1.0$\,s), B and L are the same window. Human raters assess B2 only where it precedes L (75 of the 150 rated hazard clips) and do not assess P.

\paragraph{Sensitivity to frame density.}
For Claude Opus~5.5, GPT-6 Astra, and Gemini~4 Argon, we compare our 31-frame ($1/6$\,s) protocol against an 11-frame (0.5\,s) run over the same 5.0\,s windows on 84 windows from 26 clips (25 hazard clips). Across all three models, warning decisions match on 86--95\% of windows and within-clip onset AUC changes by at most 0.04 (Table~\ref{tab:app-frame-density}).

\begin{table}[t]
\centering
\small
\setlength{\tabcolsep}{6pt}
\begin{tabular}{@{}lcc@{}}
\toprule
Model &
\begin{tabular}[b]{@{}c@{}}Warning\\agreement\end{tabular} &
\begin{tabular}[b]{@{}c@{}}$|\Delta\text{Onset}|$\\AUC\end{tabular} \\
\midrule
Claude Opus 5.5 & 86\% & 0.00 \\
GPT-6 Astra     & 95\% & 0.02 \\
Gemini 4 Argon  & 92\% & 0.04 \\
\bottomrule
\end{tabular}
\caption{\textbf{Decisions change little between 11 and 31 frames over the same 5.0\,s window.} Warning agreement is the share of 84 windows (from 26 clips) on which the binary warning decision is identical at 11 frames (0.5\,s spacing) and 31 frames ($1/6$\,s spacing). $|\Delta\text{Onset}|$ AUC is the absolute difference in within-clip onset AUC (B vs.\ A1) on the 25 hazard clips; every 95\% confidence interval includes zero.}
\label{tab:app-frame-density}
\end{table}

\paragraph{Hazard-free clips.}
Because hazard-free clips have no onset or accident, we pair each of the 259 clips with a randomly selected hazard clip from the same source (using a fixed seed) and evaluate it at the same elapsed timestamps. We sample up to five decision points per clip: 2.0\,s and 1.0\,s before that hazard clip's onset ($r - 2.0$\,s and $r - 1.0$\,s), 0.5\,s and 1.5\,s after its onset ($r + 0.5$\,s and $r + 1.5$\,s), and the last point on its 0.5\,s grid before the accident. This yields 1,233 hazard-free windows, for 8,011 windows in total.

\section{Dataset Survey and Selection}
\label{app:datasets}

We survey 37 video datasets that annotate unsafe, hazardous, or anomalous events to identify footage suitable for evaluating how a monitor's decision should evolve as danger develops (Table~\ref{tab:dataset-audit}). For each dataset, we examine its source footage, annotation granularity, quality-control protocol, and licensing terms against the five requirements introduced in Section~\ref{sec:intro}: (i)~real-world footage; (ii)~sub-second temporal annotations of both hazard onset and accident occurrence; (iii)~sufficient lead-up footage preceding the onset; (iv)~comparable hazard-free control footage; and (v)~spatial localization of the hazard together with licence terms permitting controlled counterfactual edits. No single existing dataset satisfies all five criteria on its own.

\paragraph{Accidents and hazards.}
The nine accident and physical-hazard datasets come closest to our setting. All nine use real footage, and six annotate both hazard onset and accident timestamps at sub-second precision, yet each lacks at least one required component. PaSBench-Video~\cite{zhao2026pasbench} includes matched hazard-free clips, but does not provide spatial bounding boxes and draws part of its footage from Nexar~\cite{moura2025nexar} and SmartHome-Bench~\cite{zhao2025smarthome}, whose licences restrict derivative edits. Conversely, MM-AU~\cite{fang2024abductive} provides hand-corrected road-user bounding boxes under an academic licence that permits counterfactual editing, but contains only accident clips without hazard-free controls. SPRINT~\cite{qiu2026sports} and three of RiskCueBench's~\cite{luo2026riskcuebench} four splits are distributed only as external YouTube links. CCD~\cite{bao2020uncertainty} places every collision within the final 2\,s of a fixed 5\,s clip, creating a predictable temporal structure. OOPS!~\cite{epstein2020oops} annotates the onset of unintentional action failures rather than physical safety hazards, and contains no failure-free clips.

\paragraph{Violence, crime and anomalies.}
The eleven surveillance, violence, and anomaly datasets mark intervals in which abnormal activity takes place. These intervals record the duration of an ongoing fight, theft, or disturbance rather than separating an emerging hazard onset from a subsequent harmful outcome. Four of these benchmarks provide temporal intervals only for their test splits (UCF-Crime~\cite{sultani2018real}, XD-Violence~\cite{wu2020not}, MSAD~\cite{zhu2024advancing}, and NWPU Campus~\cite{cao2023new}), and none provides spatial bounding boxes isolating a causal hazard object. Three include fictional or staged scenes: 47\% of XD-Violence clips are drawn from movies, VSD consists primarily of commercial feature films, and NWPU Campus uses staged anomalies. In addition, four restrict derivative redistribution or editing (MSAD~\cite{zhu2024advancing}, NTU CCTV-Fights~\cite{perez2019detection}, NWPU Campus~\cite{cao2023new}, and SmartHome-Bench~\cite{zhao2025smarthome}).

\paragraph{Hate and harmful content.}
In the seven hate-speech and harmful-content datasets, harm is defined by spoken language, symbols, or offensive acts rather than a developing physical hazard requiring timely intervention. Five annotate temporal spans (though DeHate~\cite{dalal2024dehate} marks them for only part of its videos and ImpliHateVid~\cite{rehman2025implihatevid} publicly releases clip-level annotations), while KuaiMod~\cite{lu2025vlm} and TikGuard~\cite{balat2024tikguard} label whole videos. HateClipSeg~\cite{wang2025hateclipseg} and MultiHateClip~\cite{wang2024multihateclip} distribute platform video IDs rather than raw footage.

\paragraph{Generated video and model-written labels.}
Eight of the ten remaining benchmarks rely on synthetic or simulated video. In Video-SafetyBench~\cite{liu2025video}, clips are generated to depict harmful content from the initial frame, while MCV SafetyBench~\cite{kang2026jailbreaking} and DIH-Bench~\cite{wang2026distributed} construct synthetic multi-clip or compositional harms. ASIMOV-2.0-Video~\cite{jindal2025ai}, HomeSafe-Bench~\cite{pu2026homesafe}, and EgoSafetyBench~\cite{panpatil2026egosafetybench} mark transitions to unsafe states in generated or simulated environments, where pre-onset visual cues are determined by the generator rather than real-world physics. VESPER~\cite{hachiuma2026vlms} pairs synthetic (Veo~3.0) safe and dangerous robot-task clips trimmed before injury with clip-level stop or continue labels. SafeWatch-Bench~\cite{chen2025safewatch} (which combines real and synthetic clips) and the two real-video benchmarks, SafeVid-350K~\cite{wang2025safevid} and VideoSafetyEval~\cite{sun2026evaluation}, assign clip-level labels with descriptions generated or filtered by language models.

\paragraph{Onset annotation protocols and human reference.}
Published accident benchmarks use different protocols to define hazard onset. Nexar uses the median timestamp across ten annotators~\cite{moura2025nexar}; SPRINT re-annotates clips when three annotators diverge by more than 0.5\,s~\cite{qiu2026sports}; OOPS! reports a median inter-annotator spread of approximately 0.5\,s across three workers~\cite{epstein2020oops}; and RiskCueBench reports boundary discrepancies below four frames~\cite{luo2026riskcuebench}. In the two datasets we draw from, MM-AU averages onset timestamps across five annotators~\cite{fang2024abductive} and PaSBench-Video adjudicates onset labels from two annotators with a third~\cite{zhao2026pasbench}, without publishing per-clip inter-annotator variance. More fundamentally, an onset annotated in hindsight with knowledge of the eventual accident does not establish when a causal observer viewing only prior frames should issue a warning. We therefore use dataset onset timestamps solely to align observation windows, and collect an independent human reference with quantified inter-rater agreement ($\kappa=0.63$, $\alpha=0.75$) on the exact 31-frame inputs shown to the models (Section~\ref{sec:res-human}; Appendix~\ref{app:human-reference}).

\paragraph{Selection rules.}
We enforce four inclusion rules: (1)~we evaluate real-world footage only, excluding PaSBench-Video's 79 synthetic clips; (2)~all reference labels and onset anchors are human-annotated; (3)~we use only video files distributed directly through official dataset repositories rather than scraping third-party platforms, which excludes SPRINT and RiskCueBench's YouTube-hosted splits (while RiskCueBench's CCD-derived car-crash split lacks sufficient pre-onset duration for 5.0\,s windows); and (4)~we perform generative counterfactual editing strictly on footage whose licence permits derivative modification, evaluating Nexar and SmartHome-Bench clips only in their unedited form.

\paragraph{Datasets used.}
Combining MM-AU~\cite{fang2024abductive} and PaSBench-Video~\cite{zhao2026pasbench} satisfies all five requirements. MM-AU provides sub-second onset and accident timestamps for DADA-2000 dashcam footage~\cite{fang2019dada}, along with road-user bounding boxes initialized by an object detector and manually corrected by ten annotators every fifth frame. PaSBench-Video contributes real hazard clips across five domains (including three non-driving settings) alongside matched hazard-free clips from three sources. Of PaSBench-Video's 105 DADA-derived clips, 102 overlap with MM-AU; we retain these in the PaSBench-Video split and remove them from our MM-AU pool so that no underlying video is evaluated twice. Filtering for clips with at least 5.5\,s of footage before the annotated onset and at least 1.0\,s between onset and accident yields 490 MM-AU DADA accident clips and 176 real PaSBench-Video hazard clips, which we pair with all 259 PaSBench-Video hazard-free clips (925 clips total). For counterfactual temporal twins (Section~\ref{sec:proto-twins}; Appendix~\ref{app:twins}), we draw exclusively from DADA-2000 clips covered by MM-AU's hand-corrected bounding boxes and academic licence, yielding 207 verified twins (193 from the MM-AU split and 14 from PaSBench-Video's DADA subset).

\paragraph{Comparison with CUVA.}
CUVA~\cite{du2024uncovering} provides approximately 1,000 real-world anomaly videos with human-annotated temporal segments and causal narratives, but its design targets long-horizon post-hoc causation rather than sub-second anticipatory warning. First, CUVA records segment boundaries at 1\,s resolution, which is coarser than our 0.5\,s decision step. Second, roughly half of its clips begin with the anomaly already underway (median onset at 0\,s), leaving insufficient lead-up for 5.0\,s pre-onset windows. Third, many clips are multi-shot compilations and every clip contains an anomaly, providing no hazard-free controls. Fourth, no spatial bounding boxes are provided for counterfactual object removal.

\begin{table*}[t]
\centering
\scriptsize
\setlength{\tabcolsep}{4pt}
\begin{tabular}{@{}lccccc p{0.50\linewidth}@{}}
\toprule
Dataset & (i) & (ii) & (iii) & (iv) & (v) & Main limitation for this study \\
\midrule
\multicolumn{7}{@{}l}{\textit{Accidents and hazards}} \\
PaSBench-Video$^\dagger$~\cite{zhao2026pasbench} & $\bullet$ & $\bullet$ & $\circ$ & $\bullet$ & $\circ$ & No native boxes (MM-AU boxes apply to DADA subset only); Nexar/SmartHome splits forbid edits \\
MM-AU$^\dagger$~\cite{fang2024abductive} & $\bullet$ & $\bullet$ & $\circ$ & $\times$ & $\bullet$ & No hazard-free control clips \\
DADA-2000$^\ddagger$~\cite{fang2019dada} & $\bullet$ & $\bullet$ & $\circ$ & $\times$ & $\times$ & Superseded by MM-AU refined timestamps and road-user boxes; no hazard-free clips \\
Nexar$^\ddagger$~\cite{moura2025nexar} & $\bullet$ & $\bullet$ & $\circ$ & $\bullet$ & $\times$ & Licence prohibits manipulated media; no hazard boxes \\
OOPS!$^\ddagger$~\cite{epstein2020oops} & $\bullet$ & $\circ$ & $\circ$ & $\times$ & $\times$ & Marks action-failure onset only; mishaps need not pose physical danger \\
SPRINT~\cite{qiu2026sports} & $\bullet$ & $\bullet$ & --- & $\bullet$ & $\times$ & Distributed as YouTube URLs only (subject to video removal) \\
RiskCueBench~\cite{luo2026riskcuebench} & $\bullet$ & $\bullet$ & $\times$ & $\bullet$ & $\times$ & Crash split too short for 5.0\,s pre-onset windows; other splits via YouTube URLs \\
CCD~\cite{bao2020uncertainty} & $\bullet$ & $\circ$ & $\times$ & $\circ$ & --- & Collisions always occur in the final 2\,s of fixed 5\,s clips \\
P$^3$IBench~\cite{cui2026perception} & $\bullet$ & $\circ$ & --- & --- & $\times$ & Evaluates response windows with mixed human and LLM annotations \\
\midrule
\multicolumn{7}{@{}l}{\textit{Violence, crime and anomalies}} \\
UCF-Crime$^\ddagger$~\cite{sultani2018real} & $\bullet$ & $\circ$ & --- & $\bullet$ & $\times$ & Temporal anomaly intervals provided for test split only; no spatial boxes \\
UCA~\cite{yuan2024towards} & $\bullet$ & $\circ$ & --- & $\bullet$ & $\times$ & Event-level caption intervals rather than hazard onset and impact \\
HIVAU-70k~\cite{zhang2025holmes} & $\bullet$ & $\circ$ & --- & $\bullet$ & $\times$ & Coarse event boundaries with LLM-generated descriptions \\
XD-Violence~\cite{wu2020not} & $\circ$ & $\circ$ & --- & $\bullet$ & $\times$ & 47\% movie footage; temporal intervals in test split only \\
MSAD~\cite{zhu2024advancing} & $\bullet$ & $\circ$ & --- & $\bullet$ & $\times$ & Gated access; licence prohibits derivative edits \\
CUVA~\cite{du2024uncovering} & $\bullet$ & $\times$ & $\circ$ & $\times$ & $\times$ & 1\,s timestamp resolution; half of clips begin inside the anomaly \\
UBI-Fights~\cite{degardin2020human} & $\bullet$ & $\circ$ & --- & $\bullet$ & $\times$ & Fight intervals only; no hazard-onset separation or spatial boxes \\
NTU CCTV-Fights~\cite{perez2019detection} & $\bullet$ & $\circ$ & --- & --- & $\times$ & Fight intervals only; licence prohibits derivative datasets \\
VSD~\cite{demarty2015vsd} & $\circ$ & $\circ$ & --- & $\bullet$ & $\times$ & Primarily commercial feature-film footage \\
NWPU Campus~\cite{cao2023new} & $\circ$ & $\circ$ & --- & $\bullet$ & $\times$ & Primarily staged anomalies; licence prohibits derivative datasets \\
SmartHome-Bench$^\ddagger$~\cite{zhao2025smarthome} & $\bullet$ & $\times$ & $\times$ & $\bullet$ & $\times$ & Clip-level labels only; licence prohibits derivative edits \\
\midrule
\multicolumn{7}{@{}l}{\textit{Hate and harmful content}} \\
HateMM~\cite{das2023hatemm} & $\bullet$ & $\times$ & --- & $\bullet$ & $\times$ & Harm is primarily verbal/symbolic; 1\,s span granularity \\
HateClipSeg~\cite{wang2025hateclipseg} & $\bullet$ & $\times$ & --- & $\bullet$ & $\times$ & Speech-driven harm; distributed as platform video IDs only \\
MultiHateClip~\cite{wang2024multihateclip} & $\bullet$ & $\times$ & --- & $\bullet$ & $\times$ & Speech-driven harm; distributed as platform video IDs only \\
DeHate~\cite{dalal2024dehate} & $\bullet$ & $\times$ & --- & --- & $\times$ & Temporal spans annotated for only a subset of videos \\
ImpliHateVid~\cite{rehman2025implihatevid} & $\bullet$ & $\times$ & --- & --- & $\times$ & Implicit hate speech; only clip-level metadata publicly released \\
KuaiMod~\cite{lu2025vlm} & $\bullet$ & $\times$ & $\times$ & --- & $\times$ & Video-level moderation labels; released as sparse frames only \\
TikGuard~\cite{balat2024tikguard} & $\bullet$ & $\times$ & $\times$ & --- & $\times$ & Video-level moderation labels without temporal onset \\
\midrule
\multicolumn{7}{@{}l}{\textit{Generated video and model-written labels}} \\
Video-SafetyBench~\cite{liu2025video} & $\times$ & $\times$ & $\times$ & $\times$ & $\times$ & Synthetic video; harmful content present from the first frame \\
SafeWatch-Bench~\cite{chen2025safewatch} & $\circ$ & $\times$ & $\times$ & --- & $\times$ & Video-level labels generated by multi-agent GPT-4o pipeline \\
MCV SafetyBench~\cite{kang2026jailbreaking} & $\times$ & $\circ$ & --- & --- & $\times$ & Synthetic multi-clip compositions; boundaries defined by concatenation \\
DIH-Bench~\cite{wang2026distributed} & $\times$ & $\circ$ & --- & --- & $\times$ & Primarily synthetic compositional harm; dataset not publicly released \\
ASIMOV-2.0-Video~\cite{jindal2025ai} & $\times$ & $\circ$ & --- & $\bullet$ & $\times$ & Synthetic Veo~3.0-generated footage; no editing licence specified \\
HomeSafe-Bench~\cite{pu2026homesafe} & $\times$ & $\circ$ & --- & --- & --- & Simulated/synthetic household scenes with keyframe labels \\
EgoSafetyBench~\cite{panpatil2026egosafetybench} & $\times$ & $\circ$ & --- & --- & $\times$ & Synthetic egocentric clips; LLM-annotated with partial human verification \\
SafeVid-350K~\cite{wang2025safevid} & $\bullet$ & $\times$ & $\times$ & --- & $\times$ & Clip-level safety labels and GPT-4-generated descriptions \\
VideoSafetyEval (VSE)~\cite{sun2026evaluation} & $\bullet$ & $\times$ & $\times$ & --- & $\times$ & Clip-level labels; dataset not publicly released \\
VESPER~\cite{hachiuma2026vlms} & $\times$ & $\times$ & $\times$ & $\bullet$ & $\times$ & Synthetic Veo~3.0-generated safe/dangerous clips; clip-level stop/continue labels \\
\bottomrule
\end{tabular}
\caption{\textbf{Survey of 37 video safety, anomaly, and accident datasets against our five evaluation requirements; combining MM-AU and PaSBench-Video satisfies all five.} Requirements (Section~\ref{sec:intro}): (i)~real-world footage; (ii)~temporally precise annotations of hazard onset and accident; (iii)~sufficient lead-up footage preceding the onset; (iv)~comparable hazard-free control footage; (v)~spatial hazard localization and licence permission for counterfactual editing. $\bullet$~satisfies, $\circ$~partially satisfies, $\times$~does not satisfy, ---~not applicable or not specified in release. $^\dagger$Primary datasets used in our evaluation. $^\ddagger$Source footage included via MM-AU or PaSBench-Video and evaluated unedited (except DADA-2000 clips covered by MM-AU annotations and licensing).}
\label{tab:dataset-audit}
\end{table*}

\section{Ethical Considerations}
\label{app:ethics}
\label{app:human-ethics}

Automated video safety monitoring has potential benefits for accident prevention and early hazard detection, but also raises risks associated with surveillance, privacy, and inappropriate intervention. False warnings can undermine trust, while missed or delayed warnings may create a false sense of security. Our experiments evaluate model behavior under controlled observation conditions, and do not establish readiness for deployment in safety-critical systems.

The study uses existing footage of real accidents and everyday hazards, some of which may contain identifiable individuals or distressing events. Counterfactual video editing introduces an additional concern: generated footage can misrepresent events if presented without its experimental context. We explicitly label all edited videos as synthetic counterfactuals to distinguish them from original footage, and use and release them solely for academic evaluation of model responses.

\paragraph{Human subjects and consent.}
The human reference study (Section~\ref{sec:res-human}; Appendix~\ref{app:human-reference}) and the blinded twin-observer study (Section~\ref{sec:exp-audit}; Appendix~\ref{app:twin-controls}) collected judgments about the need for safety warnings and the severity of perceived risk in 5.0\,s (31-frame) observation windows using the browser-based rating page of Appendix~\ref{app:rating-page}. These judgments were used to characterize human warning behavior and establish a reference for model evaluation. Separately, the edit-quality audit (Section~\ref{sec:exp-audit}; Appendix~\ref{app:twin-controls}) used a side-by-side comparison page on which two coders compared the original and removal windows at L to record whether the target road user was removed and whether the rest of the scene was preserved. All annotations were provided by adult participants (three raters for the primary reference, one blinded observer on the counterfactual twins, and two coders for the edit audit) who read the written task instructions and the description of the dashcam and everyday footage before beginning, and gave informed consent to participate. Participants were unpaid volunteers. Participation was self-paced, and participants could pause or stop at any time. All windows shown to people end at or before L, 0.5\,s before the annotated accident, so no one sees footage after the annotated accident; the post-accident window P is evaluated on models only (Appendix~\ref{app:frame-sampling}). No personally identifiable information, demographic data, or biometric attributes were collected from participants. Response files record only non-identifying rater codes, item judgments, and response durations, and any participant may have their annotations removed upon request.

\paragraph{Source footage and licensing compliance.}
All video frames shown to participants and models are drawn from active, publicly distributed research benchmarks, MM-AU~\cite{fang2024abductive} (building on DADA-2000~\cite{fang2019dada}) and PaSBench-Video~\cite{zhao2026pasbench}, compiled from public road-safety, dashcam, and everyday video sources; as documented in our 37-dataset survey (Appendix~\ref{app:datasets}), no single surveyed dataset meets all five requirements of our evaluation. None of the datasets we use has been withdrawn or discredited by its creators. We downloaded all videos from their official repositories and restricted counterfactual editing to DADA-2000 clips from MM-AU and from PaSBench-Video's DADA subset, which both datasets distribute under CC BY-NC, a licence that permits derivatives for non-commercial use. Nexar's dataset licence permits modification but prohibits using the dataset to create manipulated media~\cite{moura2025nexar}, and SmartHome-Bench's licence (CC BY-NC-ND~4.0) prohibits derivative works, so we kept all Nexar and SmartHome-Bench clips unedited and evaluated them only as distributed by PaSBench-Video. Participants were instructed to delete local copies of the study folders after returning their CSV files.

\section{Reproducibility}
\label{app:reproducibility}

Our evaluation uses fixed observation windows, a common 31-frame sampling protocol, and identical decision times across models. Each model receives independent queries with no accumulated state. We record the model identifier, provider configuration, prompt, sampled frame times, raw responses, and parsing outcomes (Appendix~\ref{app:protocol}). All statistical comparisons are paired on the same video clips, and 95\% confidence intervals are computed via clip-level bootstrap resampling.

The appendices document counterfactual twin construction and frame alignment (Appendix~\ref{app:twins}), control, observer, and edit-audit details (Appendix~\ref{app:twin-controls}), extended model timing and rater-sensitivity results (Appendix~\ref{app:model-results}), the human annotation interface and consensus-threshold analyses (Appendix~\ref{app:human-reference}), the temporal sampling procedure and verbatim monitoring prompt (Appendix~\ref{app:protocol}), and the dataset selection rules (Appendix~\ref{app:datasets}). The accompanying supplementary archive includes: (1)~all five video versions (\texttt{original}, \texttt{reencoded}, \texttt{identity}, \texttt{recolour}, and \texttt{removal}) for the 11.07\,s edited segment of the motorcyclist incident in Figure~\ref{fig:motivation} (\texttt{mmau\_dada\_06\_076}); (2)~video triplets (\texttt{original}, \texttt{recolour}, and \texttt{removal}) covering the 5.0\,s window at L (plus 0.25\,s) for eight counterfactual twins from Figure~\ref{fig:twin-examples}; and (3)~full-resolution frame-strip PNGs and JSON metadata for the 19 removal windows on which the observer warned (Table~\ref{tab:app-observer-descriptions} and Figures~\ref{fig:app-observer-strips-a}--\ref{fig:app-observer-strips-c}). To support full reproduction while respecting upstream dataset licences, we will release our evaluation code, prompts, window manifests, counterfactual editing specifications, and anonymised human and model response records for academic research use, together with the edited counterfactual twin videos (constructed solely from DADA-2000 clips whose CC BY-NC licence permits non-commercial derivatives). For all unedited footage, including sources whose licences forbid derivatives or manipulated media, we provide scripts that fetch the underlying videos directly from the official MM-AU and PaSBench-Video repositories.

%% file: limitations.tex
\section{Limitations}
\label{app:limitations}

The primary practical limitation of our methodology is the human calibration and auditing effort required to keep temporal and counterfactual video evaluation rigorous. Because retrospective dataset onset labels do not capture when danger first becomes apparent to a causal observer, and because current generative video editors can leave target road users partially visible or alter surrounding traffic, reliable evaluation cannot rely on automated video generation or raw dataset timestamps alone. Ensuring that these factors remain within measured bounds required collecting a human reference from multiple raters with quantified agreement and sensitivity checks leaving out one rater at a time (Appendices~\ref{app:human-reference} and~\ref{app:model-results}), constructing five versions of each incident and aligning their frames by matching the background (Appendix~\ref{app:twins}), and conducting both a blinded observer study and an edit audit by two independent coders to separate clean removals from imperfect edits (Appendix~\ref{app:twin-controls}).

Finally, we evaluate models on independent 5.0\,s observation windows with no accumulated conversation history. This is a deliberate diagnostic choice that prevents early false alarms from autoregressively biasing later decisions, and isolates how monitors respond to the visual evidence available at each moment (Appendix~\ref{app:protocol}). Extending human reference curves and counterfactual video twins to stateful streaming monitors, adaptive warning thresholds, and interactive intervention policies is a natural direction for future work.